\documentclass[journal]{IEEEtran}

\ifCLASSINFOpdf  
\else
\fi
\usepackage{float} 
\usepackage{subfigure}
\usepackage{booktabs} 
\usepackage{amssymb}
\usepackage{cite}
\usepackage{amsmath}
\usepackage{threeparttable}
\usepackage{multirow}
\usepackage{graphicx}
\usepackage{bm}
\usepackage{multirow}
\usepackage{algorithm}
\usepackage{algpseudocode}
\usepackage{amsmath}

\usepackage{color, soul}
\setulcolor{black} 

\begin{document}
\title{Fast ORB-SLAM without Keypoint Descriptors}


\author{Qiang Fu, Hongshan Yu, Xiaolong Wang, Zhengeng Yang, Yong He, Hong Zhang,~\IEEEmembership{Fellow,~IEEE}, Ajmal Mian



\thanks{This work was supported in part by the National Natural Science Foundation of China under Grant U2013203, 61973106, U1913202, and by the China Scholarship Council under Grant 201906130082. in part by the Key Research and Development Project of Science and Technology Plan of Hunan Province under Grant 2022GK2014, in part by the Project of Talent innovation and Sharing Alliance of Quanzhou City under Grant 2021C062L. 
\textit{(Corresponding author: Hongshan Yu, Zhengeng Yang)}}. 

\thanks{Qiang Fu, Hongshan Yu,Zhengeng Yang and Yong He are with the National Engineering Laboratory for Robot Visual Perception and Control Technology, College of Electrical and Information Engineering, Hunan University, China (email: yuhongshancn@hotmail.com). Qiang Fu is also with the Department of Computing Science, University of Alberta, Canada (email: cn.fq@qq.com). Hongshan Yu is also with Quanzhou Institute of industrial design and machine intelligence innovation, Hunan University }

\thanks{Hong Zhang is with the Southern University of Science and Technology, Shenzhen, China.}

\thanks{ Xiaolong Wang is  with the School of Mathematics and Information Science, Shaanxi Normal University, China.}

\thanks{Ajmal Mian is with the Department of Computer Science, The University of Western Australia, WA 6009. }

\thanks{Qiang Fu and Hongshan Yu contributed equally to this work. }


}

\markboth{Journal of \LaTeX\ Class Files, 2021}%
{Shell \MakeLowercase{\textit{et al.}}: Bare Demo of IEEEtran.cls for IEEE Transactions on Image Process Journals}

\maketitle

\vspace{-5mm}

\begin{abstract}
Indirect methods for visual SLAM are gaining popularity due to their robustness to environmental variations. 
ORB-SLAM2 \cite{orbslam2} is a benchmark method in this domain, however, it consumes significant time for computing descriptors that never get reused unless a frame is selected as a keyframe. 
To overcome these problems, we present FastORB-SLAM which is light-weight and efficient as it tracks keypoints between adjacent frames without computing descriptors. To achieve this, a two stage descriptor-independent keypoint matching method is proposed based on sparse optical flow. In the first stage, we predict initial keypoint correspondences via a simple but effective motion model and then robustly establish the correspondences via pyramid-based sparse optical flow tracking. In the second stage, we leverage the constraints of the motion smoothness and epipolar geometry to refine the correspondences. In particular, our method computes descriptors only for keyframes. We test FastORB-SLAM on \textit{TUM} and \textit{ICL-NUIM} RGB-D datasets and compare its accuracy and efficiency to nine existing RGB-D SLAM methods. Qualitative and quantitative results show that our method achieves state-of-the-art accuracy and is about twice as fast as the ORB-SLAM2.
\end{abstract}

\vspace{-1mm}


\IEEEpeerreviewmaketitle

\vspace{-3.5mm}

\section{Introduction}
\vspace{-1mm}

\IEEEPARstart{V}{isual} simultaneous localization and mapping (SLAM) has been an active field of research in recent years \cite{orbslam2, orbslam1, svo2, svo1, dso, cnn-slam, cnn-svo, openvslam}. SLAM provides a powerful solution for mobile robots to estimate six degrees-of-freedom (DoF) pose (position and orientation) and recover the 3D structure of the surroundings from a camera's video stream. Visual SLAM is gaining importance in many application areas \cite{tip1}, such as virtual reality (VR), augmented reality (AR), unmanned aerial vehicle (UAV) or unmanned ground vehicle (UGV) navigation, and autonomous mobile robots. \par

\begin{figure}[t]
\centering  
	
	\subfigure[Raw Image]{
	\includegraphics[width=0.23\textwidth]{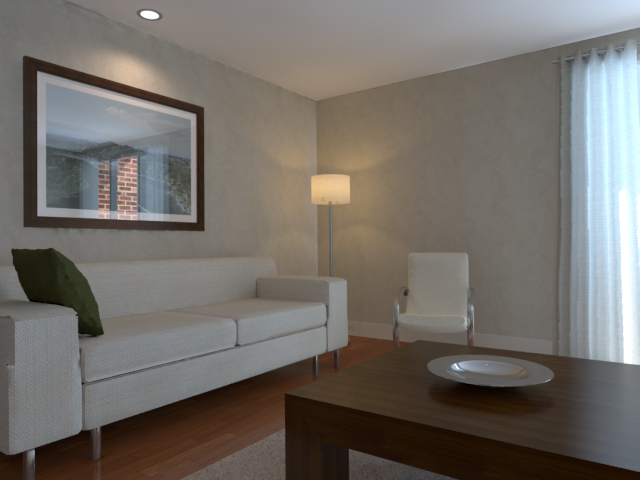}}
	\subfigure[Keypoints in ORB-SLAM]{
	\includegraphics[width=0.23\textwidth]{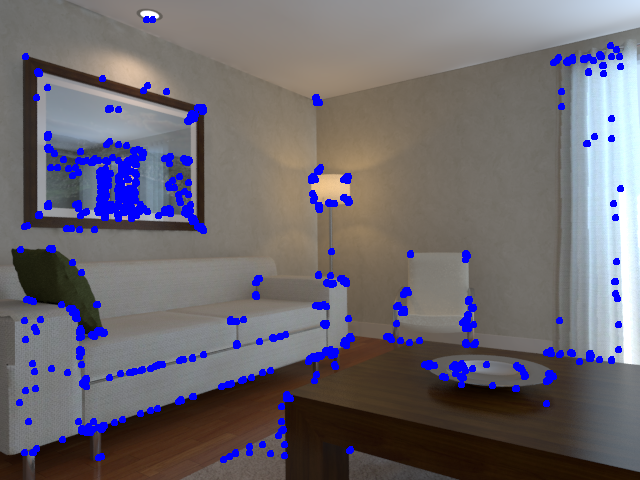}}\hspace{-1mm}
	\subfigure[Keypoint Matching without Descriptor (Ours)]{
	\includegraphics[width=0.46\textwidth]{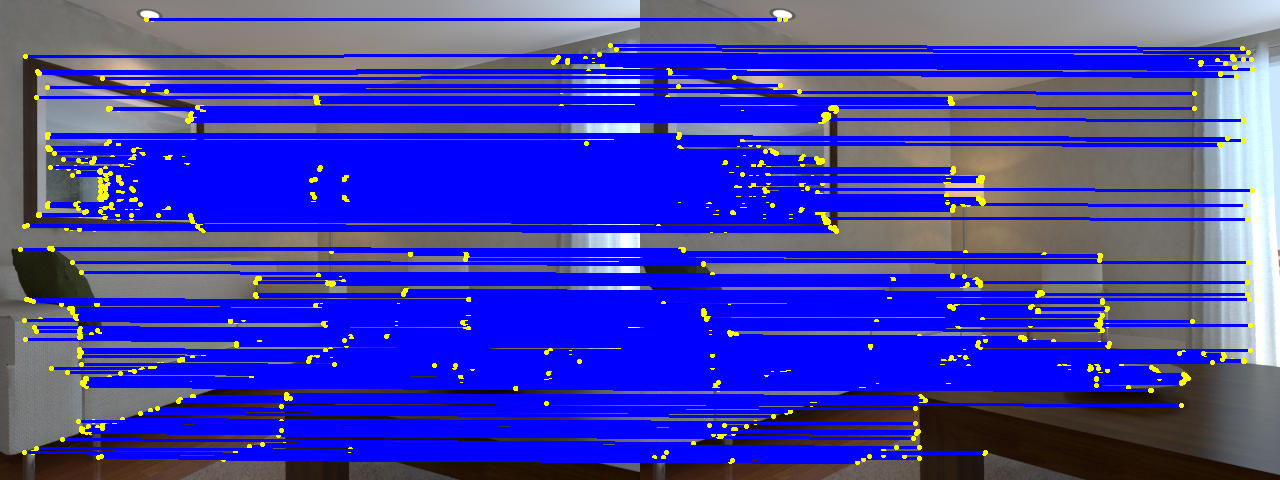}}	
\vspace{-1mm}
\caption{Illustration of our keypoint matching method between two adjacent frames from the \textit{ICL-NUIM} dataset \cite{ICL}. ORB-SLAM takes $\backsim$16 ms to extract keypoints ($\backsim$8 ms for detection + $\backsim$8 ms for description) under default parameters (max 1000 keypoints), whereas our method takes only $\backsim$12 ms to establish reliable keypoint correspondences without extracting descriptors.}\label{figure1}
\vspace{-6mm}
\end{figure}


High-accuracy and low-computational cost are the two core requirements of visual SLAM \cite{tip2, posetip, fu1, dvo, rgbdslam, lei, bundle, badslam}. Current methods are divided into photometric-based direct methods, e.g., DSO \cite{dso} and SVO \cite{svo2}, and feature-based indirect methods \cite{plslam, direct-imu, fu2}. 
Direct methods recover pose by minimizing the pixels' photometric errors whereas indirect methods leverage discriminative image features to recover camera pose by minimizing the reprojection errors between the feature correspondences, and implement loop closure (relocation) to eliminate the global drift based on the feature descriptors. Point-based methods track discriminative keypoints along successive frames and then recover the camera motion trajectory. These methods are robust because the discriminative keypoints are relatively invariant to changes in viewpoint and illumination. Recently, many indirect SLAM methods were proposed for real-time applications \cite{openvslam, plslam}. Among these, ORB-SLAM2 is considered to be the current state-of-the-art SLAM method \cite{dynaslam, dxslam, zhao}. It is developed based on many excellent works, e.g., the first real-time visual SLAM system, PTAM \cite{ptam}, a fast place recognition method, BoW2 \cite{bags}, and an efficient graph-based bundle adjustment (BA) algorithm, \textit{covisibility graph} \cite{covision}. Therefore, ORB-SLAM2 achieves better accuracy and robustness than other existing solutions. 
\par

Mainstream indirect methods such as ORB-SLAM2 implement three threads: \textit{Tracking}, \textit{Local Map} and \textit{Loop Closure}. The \textit{Tracking} thread establishes keypoint correspondences in adjacent frames based on descriptor matching, and then estimates and outputs camera pose in real time. Once a current frame is selected as a keyframe, the last two threads are activated to refine camera motion but not in real time. The \textit{Tracking} part is considered as the foundation of any SLAM system, since it not only has an immediate impact on accuracy and robustness but also provides association information for the other two threads. Naturally, it takes up most of the computational resources. \par

We observe that the computation of keypoint descriptors in indirect methods is time-consuming and the descriptors are not reused except in the case of keyframes. 
This wastes significant computational sources. If we can establish reliable keypoint correspondences without extracting descriptors between adjacent frames (or equivalently in \textit{Tracking}), it will greatly reduce the computational cost without loss of precision.
\par

Based on the above, we present FastORB-SLAM, a fast and lightweight visual SLAM system that outputs high-accuracy 6-DoF pose estimates in real time. Unlike indirect methods such as ORB-SLAM2, our method matches keypoints between adjacent frames in \textit{Tracking} without extracting descriptors (see Fig.~\ref{figure1}). The keypoint matching method is designed into two stages: The first stage is for robust keypoint matching where we predict the initial keypoint correspondences by a uniform acceleration motion (UAM) model and then use a pyramid-based sparse optical flow algorithm to establish coarse keypoint correspondences. The second stage is for inlier refinement where we exploit grid-based motion statistics \cite{gms2} to filter out outliers and then utilize the epipolar constraint to further refine the correspondences. 
%
%
%
Our main contributions are summarized as follows:
\begin{itemize}
\item We present a light-weight SLAM system, coined FastORB-SLAM, which is developed based on ORB-SLAM2 and sparse optical flow. FastORB-SLAM exploits a new structure where matching keypoints between adjacent frames is based on minimizing the grayscale errors and matching keypoints between non-adjacent frames (keyframes) is based on keypoint descriptors. This design balances the competing needs between the  localization accuracy and computational complexity.

\item We propose a two-stage keypoint matching algorithm to establish reliable keypoint correspondences between adjacent frames without descriptors. Our algorithm exploits the UAM model to predict initial keypoint correspondences which not only improves the accuracy of keypoint matching but also reduces the computational complexity of correspondence search.


\item FastORB-SLAM was compared with nine (almost all) existing representative open-source RGB-D SLAM systems in terms of localization accuracy (RMSE) and computation time on well-known RGB-D datasets,s \textit{TUM} \cite{TUM} and \textit{ICL-NUIM} \cite{ICL}. Qualitative and quantitative results show our method achieves state-of-the-art performance.

\item FastORB-SLAM is about twice as fast as the benchmark ORB-SLAM2 with highly competitive localization accuracy. See our demo at https://b23.tv/dJqNkU4. 
\end{itemize}
%
%

\section{Related Work}
\label{related work}
High accuracy and low-computational cost are the two core requirements of visual SLAM \cite{svo1, orbslam1}. 
Current visual SLAM methods are divided into photometric-based direct methods and feature-based indirect methods:

\noindent \textbf{Photometric-based Direct SLAM:}
These methods solve pose estimation by minimizing pixel-level intensity errors of images \cite{svo2}. Recent representative works include semi-direct methods \cite{svo2} and sparse direct methods  \cite{dso}. 
Forster \textit{et al.} proposed semi-direct visual odometry (SVO) \cite{svo2}, a two-thread framework that consists of \textit{Tracking} and \textit{Local Mapping}. It tracks sparse pixels at the FAST corners \cite{orb} to recover motion in \textit{Tracking}, and refines the pose in \textit{Local Mapping}. SVO uses a \textit{depth filter} model to estimate pixel depth values and filters outliers. It models the triangulated depth observations with a Gaussian Uniform distribution. If the triangulated depths of the same feature point are within a small range, then the mean and variance of the depth values can be obtained using a Gaussian distribution. However, if the depth values are spread out, then they follow a uniform distribution. If a feature contains many outliers, it is filtered out as it does not converge to a Gaussian distribution with a small variance. To solve this problem, Loo \textit{et al.} proposed CNN-SVO \cite{cnn-svo} which uses a mono depth prediction network to predict depth values at corners to improve upon the robustness of SVO. \par

Engel \textit{et al.} proposed direct sparse odometry (DSO) \cite{dso}, a probabilistic model that directly minimizes photometric error without computing keypoints or descriptors. Similar to CNN-SVO, Yang \textit{et al.} leveraged deep learning based depth prediction to improve the performance of DSO \cite{ddso, d3vo}. Wang \textit{et al.} proposed Stereo DSO, where the depth value of a pixel is estimated by multi-view geometry \cite{sdso}. Schubert \textit{et al.} adopted a rolling shutter model to improve robustness \cite{rolldso}. Von \textit{et al.} integrated an Inertial Measurement Unit (IMU) sensor to improve the robustness in case of quick movements \cite{imudso}. Similar works with an IMU include \cite{vins1, vins2, vio}. Gao \textit{et al.} added a \textit{Loop Closure} thread to eliminate global drift errors \cite{ldso}. Extensive experiments show that direct methods have an obvious advantage in computing speed \cite{dso}, however, they have poor robustness and accuracy. \par

Direct methods generally use a coarse-to-fine strategy to improve robustness where the camera pose is successively solved for in a multi-level image pyramid, such as in \cite{svo2, svo1, dso, nid, fusion}. In other works, the term ``coarse-to-fine'' is used for a two-stage process that estimates a coarse value in the first stage, and then refines the coarse value to improve accuracy in the second stage. For example, Ding \textit{et al.} proposed a two-stage (coarse-to-fine) retrieval framework called CamNet for camera re-localization \cite{camnet}. CamNet first obtains a coarse relative pose estimate using an image-based retrieval module and then refines the estimate using a pose-based retrieval module. Mori \textit{et al.} adopted a two-stage (coarse-to-fine) convolutional neural network (CNN)-based framework that first estimates a coarse keypoint orientation and then refines it \cite{coarse}. The proposed FastORB-SLAM not only adopts a two stage process to establish accurate keypoint correspondences, but also uses an image pyramid to iteratively solve for the movement vector in the first stage. As a result, FastORB-SLAM can establish reliable keypoint correspondences
%

\noindent \textbf{Feature-based Indirect SLAM:}
This group of methods leverage salient image features, such as point or line features, to recover and refine camera motion by minimizing reprojection errors of the feature correspondences \cite{ptam, plslam, fu2}. \par

Georg \textit{et al.} \cite{ptam} proposed PTAM, the first real-time feature indirect SLAM method, which includes two parallel threads. PTAM estimates pose in real-time in the \textit{Tracking} thread, and refines the camera motion in the \textit{Local Mapping} thread. Many follow up works have been proposed based on PTAM for real-time applications \cite{rgbdslam, openvslam}. Among these, ORB-SLAM2 is known as the current state-of-the-art as it achieves unprecedented performance \cite{orbslam1}. In addition to the above mentioned two parallel threads, ORB-SLAM2 adds a \textit{Loop Closure} thread as a global constraint. This thread builds on the bag of words (BoWs) model \cite{bags} and the \textit{covisibility graph} \cite{covision}. The former is used to measure similarity of two frames and the latter is used for high-efficiency large-scale Bundle Adjustment. \par 

Subsequently, many methods have been proposed based on the ORB-SLAM2. Point-based methods give poor localization accuracy and even fail in scenes with low-texture where they cannot track enough keypoints. For this problem, Gomez-Ojeda \textit{et al.} \cite{plslam} and Fu \textit{et al.} \cite{fu2} integrated line features into the ORB-SLAM2 system to improve robustness in low-texture scenes. To meet the requirement of pose estimation in dynamic scenes, Bescos \textit{et al.} proposed Dyna-SLAM \cite{dynaslam} which adds a preprocessing step based on ORB-SLAM2 to recognize and then cull dynamic objects using the Mask-RCNN network \cite{mask}. ORB-SLAM2 has also been extended for other applications, such as robot navigation \cite{locationorb, Buyval}, semantic SLAM \cite{Zhao, Webb}, etc. Recently (in Aug 2020), ORB-SLAM3 was released on arXiv \cite{orbslam3} which integrates IMU with ORB-SLAM2. \par

Apart from point detection, binary feature description and efficient matching play important roles in point-based SLAM \cite{fan3}. For the description problem, Fan \textit{et al.} \cite{fan1} leveraged an unsupervised framework to learn binary descriptors, which produce higher accuracy than previous unsupervised and even most supervised methods. Similar unsupervised learning-based works include DBD-MQ \cite{duan} and Deepbit \cite{lin}. Binary descriptors are generally based on intensity only, however, \cite{fan2} proposed intensity and gradient-based features which obtain SOTA performance. For the matching problem, \cite{fan4} proposed a Euclidean space-based descriptor matching method, instead of the conventional Hamming space-based, to achieve improved search accuracy.

Indirect methods extract sparse features/descriptors, match descriptors of successive frames, recover camera motion and refine pose and map structure through minimizing reprojection errors between feature correspondences. Hence, compared with direct methods, indirect methods take more computational resources to extract salient indirect features. Robust features make the system more reliable at the additional computational cost of feature extraction. \par

\noindent \textbf{Summary:} A complete SLAM system (direct or indirect) must include three threads: \textit{Tracking}, \textit{Local Mapping}, and \textit{Loop Closure}. \textit{Tracking} runs at front-end and outputs current camera pose in real-time. \textit{Local Mapping} and \textit{Loop Closure} run at back-end but not in real-time. They are designed to refine (optimize) camera motion and map structure with local or global constraints. \textit{Loop Closure} is an essential thread to improve robustness in a life-time operation because it provides a powerful constraint to correct globally accumulated errors. Moreover, it can also be used for re-localization when the system fails to track the efficient features \cite{xl, zhBoCNF, zhkp}. \par

Whether minimizing photometric errors in direct methods or reprojection errors in indirect methods, it boils down to a non-linear least-squares optimization problem, which can be efficiently solved by the BA algorithm \cite{plslam}. Once correspondences are established, pose estimation or refinement problem can be solved through the BA optimization. Hence, it is extremely important for visual SLAM to establish accurate feature correspondences. \par

\begin{figure*}[htp]
\centering  
	\subfigure{
	\includegraphics[width=0.97\textwidth]{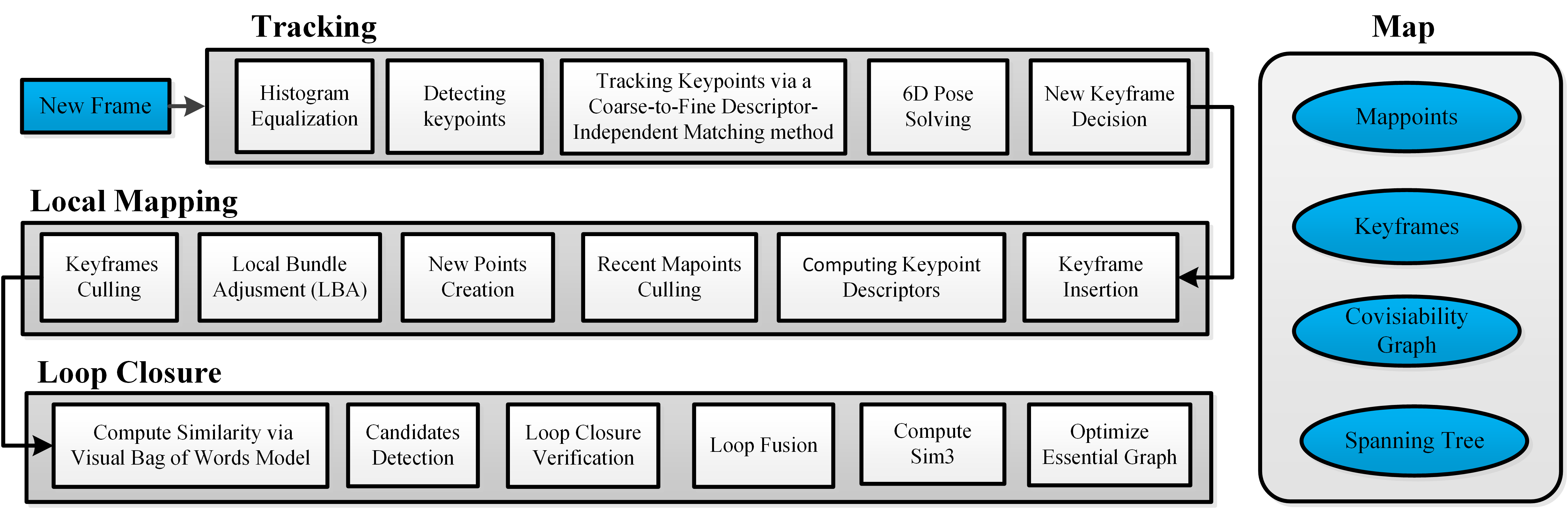}}
	\vspace{-3mm}
\caption{System overview. FastORB-SLAM consists of three threads: \textit{Tracking}, \textit{Local Mapping}, and \textit{Loop Closure}. \textit{Tracking} estimates 6-DoF camera pose in real-time. \textit{Local Mapping} adds a new keyframe and optimizes local keyframes by BA optimization. \textit{Loop Closure} constantly checks a loop and corrects the global drift with BA optimization. The Map structure contains information of keyframe, map points, \textit{covisible graph}, and spanning tree. The structure is compact and designed for efficient computation \cite{orbslam1}, it maintains useful observations 
and timely culls useless information to avoid redundant computation.}\label{system}
\vspace{-3mm}
\end{figure*}

\vspace{-2mm}
\section{System Overview} 
\vspace{-1mm}

We propose FastORB-SLAM, a fast and light-weight visual SLAM system as shown in Fig.~\ref{system}. In a nutshell, the structure of FastORB-SLAM is developed based on the ORB-SLAM2 and implements three threads: \textit{Tracking}, \textit{Local Mapping}, and \textit{Loop Closure}. 
\textit{Tracking} runs at front-end and outputs real-time camera pose estimates as well as provides observation information between frames for the other two threads. \textit{Local Mapping} and \textit{Loop Closure} run at back-end but not in real-time. The are activated only when a frame is selected as a keyframe to eliminate local or global drift errors for highly accurate pose estimation. Camera pose estimation and optimization are implemented based on a \textit{Map} structure.   
\par

Compared with ORB-SLAM2, a conventional indirect method, our method establishes keypoint correspondences between adjacent frames based on grayscale matching instead of descriptor matching. Hence, our method does not need to compute keypoint descriptors during \textit{Tracking}. In particular, the descriptors are computed only when the current frame is selected as a keyframe, whereas ORB-SLAM2 computes the descriptors for every frame.
\par

Compared with DSO \cite{dso}, a conventional direct method, our method leverages salient point features to improve robustness instead of directly using pixel grayscale values, and is able to eliminate global errors with the extracted keypoint descriptors at back-end.
\par

Compared with SVO \cite{svo2}, a conventional semi-direct method, there are three main differences. Firstly, our method adopts different keypoint detector. Secondly, SVO cannot implement a loop closure as it does not extract descriptors. Finally, SVO tracks keypoints to recover motion by directly minimizing photometric errors which has a limitation that if keypoint correspondences contain many outliers, these outliers will lead to a very poor localization accuracy (details in Section \ref{related work}). FastORB-SLAM first establishes keypoint correspondences and then deals with the outlier problem with an explicit two-stage descriptor-independent keypoint matching method. After that, the camera motion is recovered by minimizing the reprojection errors between the keypoint correspondences.
\par 

FastORB-SLAM exploits a new structure where matching keypoints between adjacent frames is based on the grayscale invariance and matching keypoints between non-adjacent frames (keyframes) is based on descriptor invariance. To achieve this, a two-stage descriptor-independent keypoint matching method is proposed. Hence, our method only extracts descriptors when the current frame is selected as a keyframe. This design balances the competing needs between the localization accuracy and computational complexity. More details of FastORB-SLAM are given below: \par

\noindent \textbf{Map:} The map structure is compact and designed for high-efficiency computation when the system performs BA \cite{orbslam1}, it retains useful observations and timely culls useless information to avoid unnecessary computation. The structure consists of:
\begin{itemize}
	\item \textit{Keyframes:} Each keyframe contains camera pose parameters, observed keypoints, and descriptors.
	\item \textit{Mappoints:} Each mappoint consists of a 3D landmark that is observed by the corresponding keypoint and its 3D position in the world coordinate system. 
	\item \textit{Covisibility graph:} This graph contains covisibility information of keyframes \cite{covision}, where each node represents a keyframe, and the edges between keyframes are created only when sharing a minimum number of landmarks (set to 20 in our experiments). Implementing local BA means to optimize the current keyframe and its neighbor keyframes (nodes) in the graph for fast pose refinement.
	\item \textit{Spanning tree:} Spanning tree stands for the minimum connected representation of a graph that includes all keyframes. Once a spanning tree is established, a corresponding \textit{essential graph} is created. Unlike \textit{covisibility graph}, an edge in \textit{essential graph} is created only when two keyframes share over 100 landmarks, hence, it is more sparse. The spanning tree and \textit{essential graph} proposed by ORB-SLAM \cite{orbslam1} allow for a fast global BA.
\end{itemize}\par
Optimizing the graph is equivalent to optimizing keyframe poses (nodes) based on observation constraints (edges). Controlling graph scale (nodes and edges) controls computational requirements. \par

\begin{figure}[tp]
\centering  
	\subfigure{
	\includegraphics[width=0.073\textwidth]{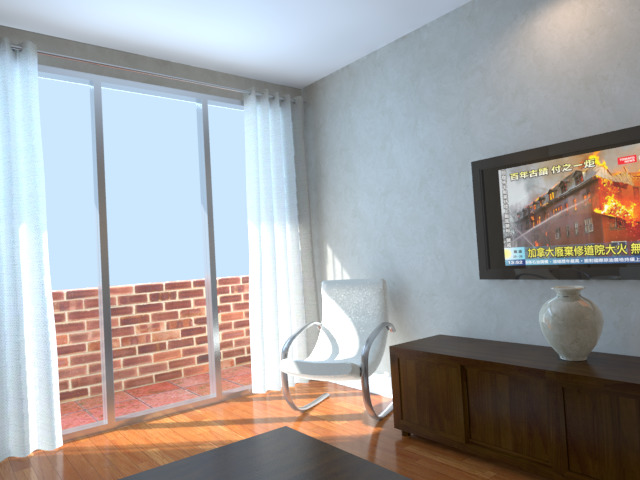}}\hspace{-1mm}
	\subfigure{\
	\includegraphics[width=0.073\textwidth]{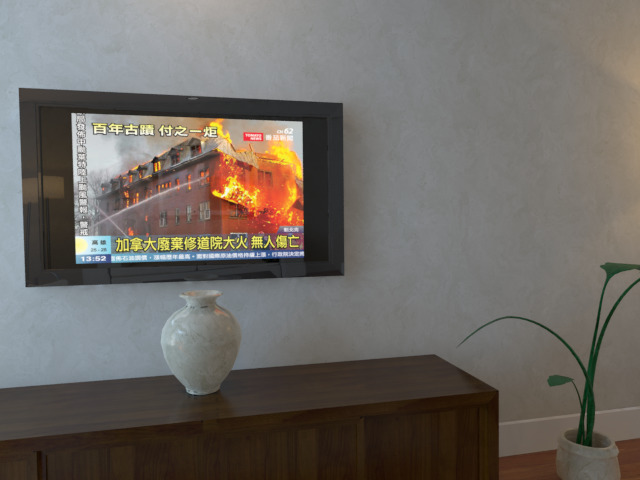}}\hspace{-1mm}	
	\subfigure{
	\includegraphics[width=0.073\textwidth]{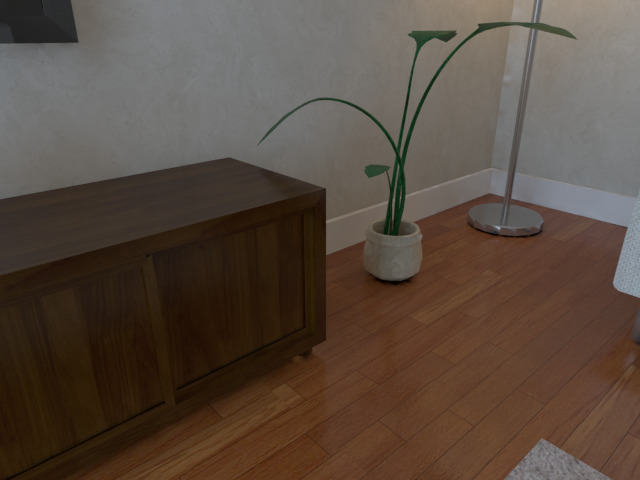}}\hspace{-1mm}
	\subfigure{
	\includegraphics[width=0.073\textwidth]{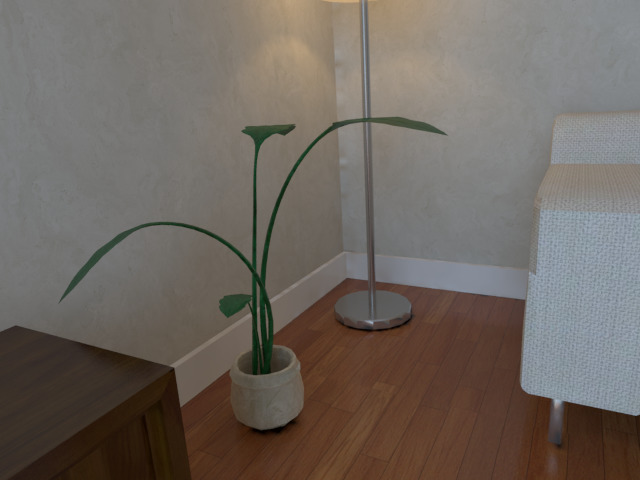}}\hspace{-1mm}
	\subfigure{
	\includegraphics[width=0.073\textwidth]{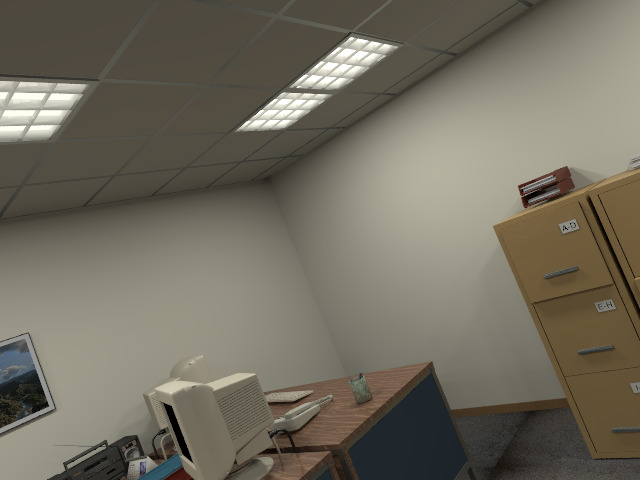}}\hspace{-1mm}
	\subfigure{
	\includegraphics[width=0.073\textwidth]{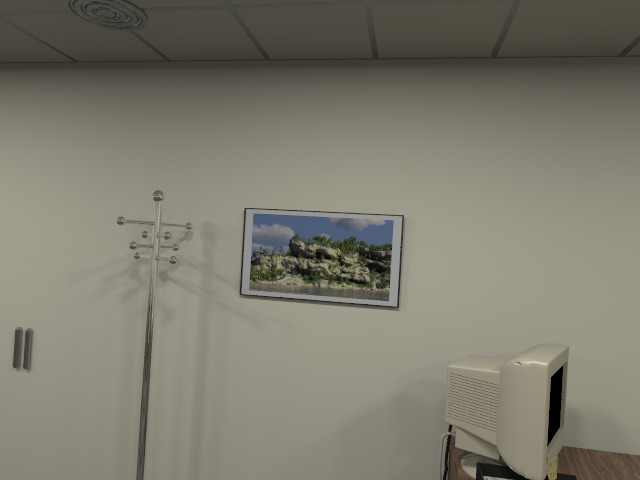}}
	\vspace{-2mm}
	\quad

	\subfigure{
	\includegraphics[width=0.073\textwidth]{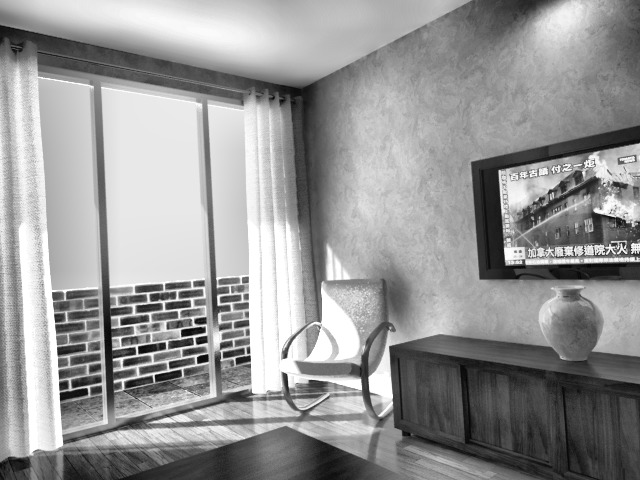}}\hspace{-1mm}
	\subfigure{\
	\includegraphics[width=0.073\textwidth]{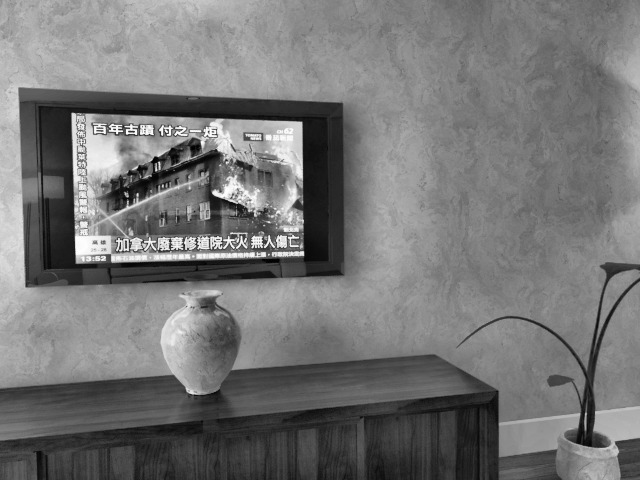}}\hspace{-1mm}	
	\subfigure{
	\includegraphics[width=0.073\textwidth]{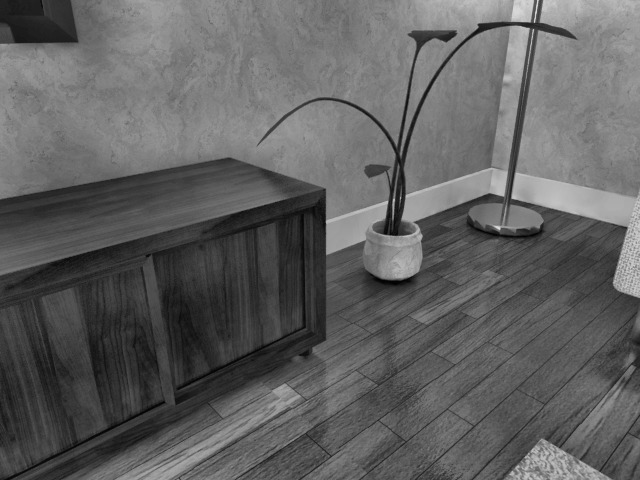}}\hspace{-1mm}	
	\subfigure{
	\includegraphics[width=0.073\textwidth]{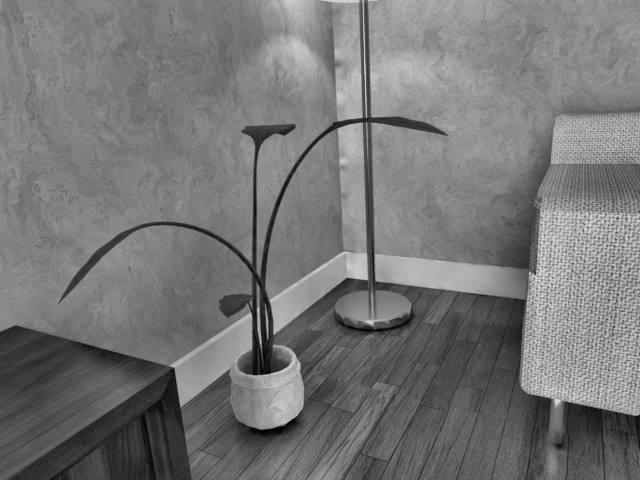}}\hspace{-1mm}
	\subfigure{
	\includegraphics[width=0.073\textwidth]{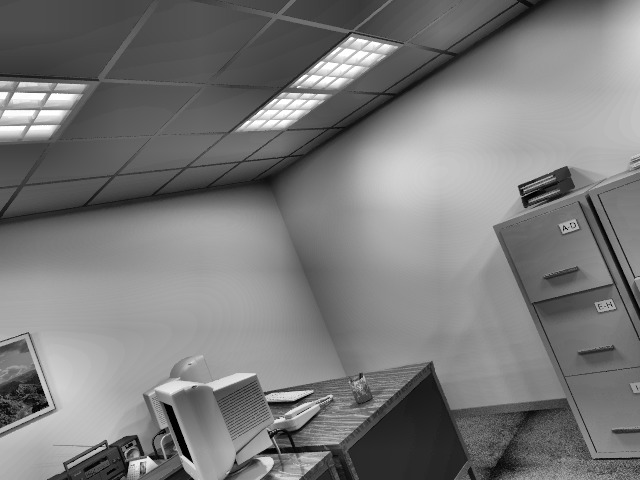}}\hspace{-1mm}
	\subfigure{
	\includegraphics[width=0.073\textwidth]{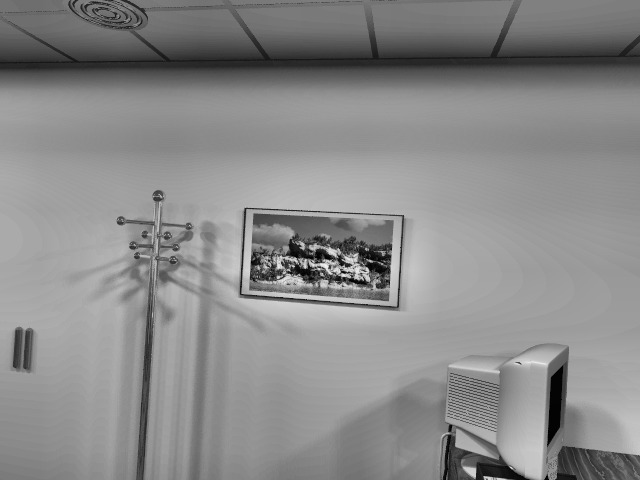}}
	\quad
	
	
\caption{Examples of preprocessing an image with an adaptive histogram equalization algorithm to reduce illumination effects. Top row represents original images from the \textit{ICL-NUIM} dataset and bottom row represents images after equalization. It takes $\backsim$1.5 ms per frame.}\label{equalize}
\vspace{-4mm}
\end{figure}

\begin{figure*}[tp]
\centering  

	\subfigure{
	\includegraphics[width=0.102\textwidth]{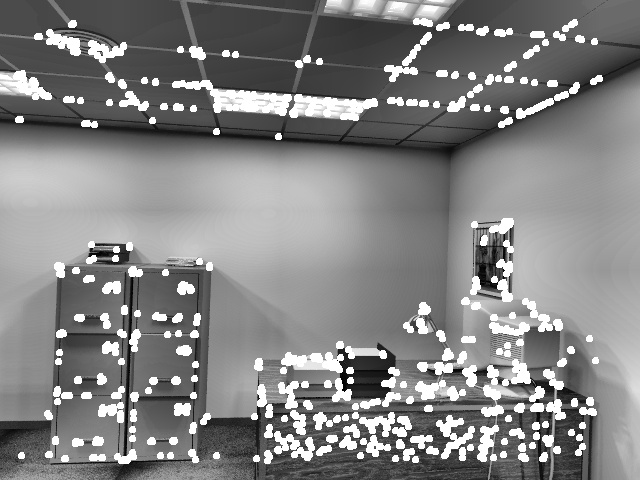}}\hspace{-1mm}
	\subfigure{\
	\includegraphics[width=0.102\textwidth]{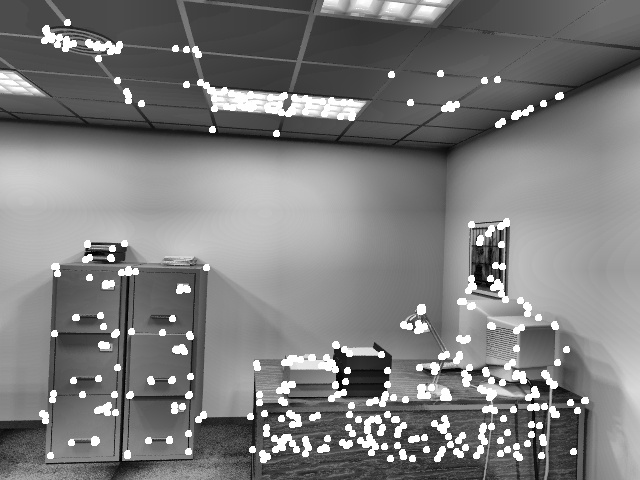}}\hspace{-1mm}	
	\subfigure{
	\includegraphics[width=0.102\textwidth]{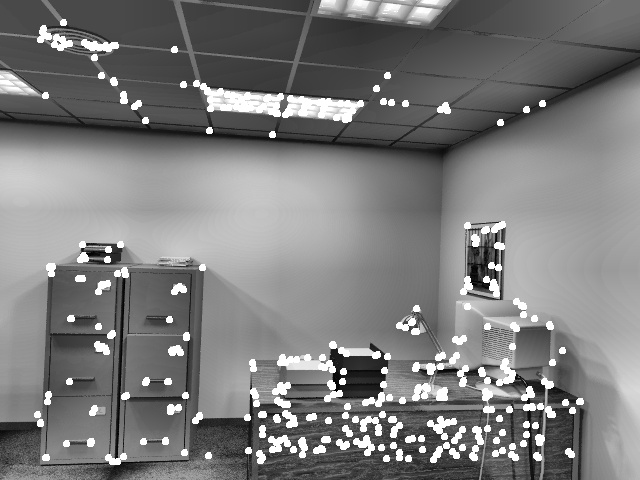}}\hspace{-1mm}
	\subfigure{
	\includegraphics[width=0.102\textwidth]{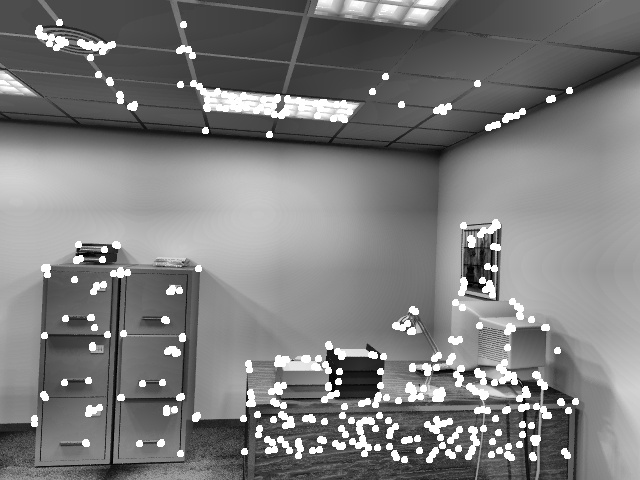}}\hspace{-1mm}
	\subfigure{
	\includegraphics[width=0.102\textwidth]{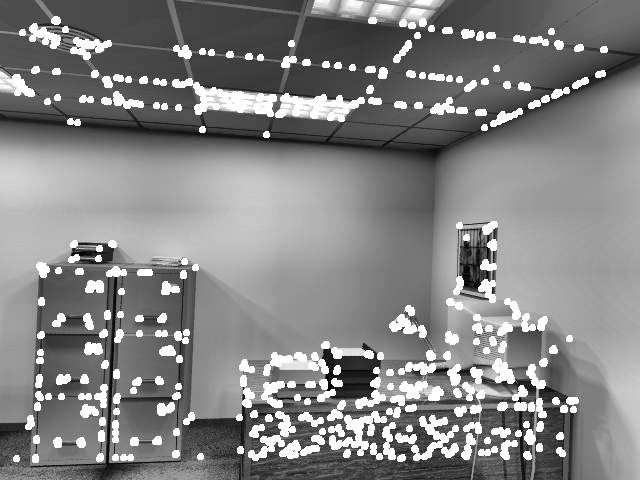}}\hspace{-1mm}
	\subfigure{
	\includegraphics[width=0.102\textwidth]{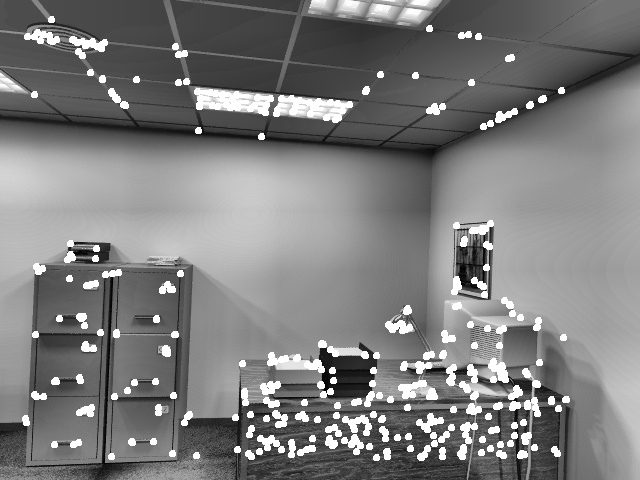}}\hspace{-1mm}
	\subfigure{
	\includegraphics[width=0.102\textwidth]{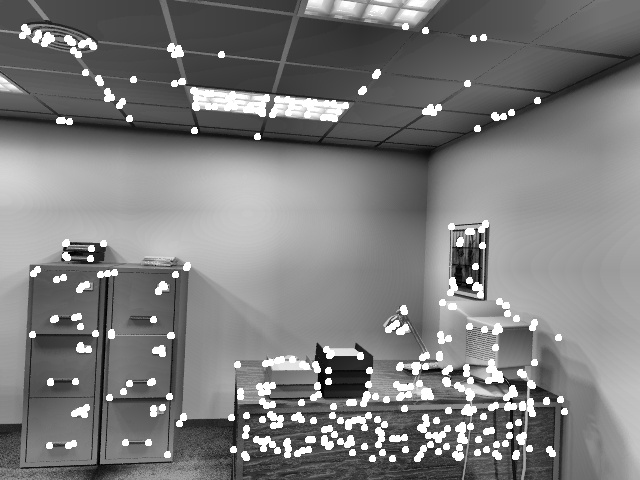}}\hspace{-1mm}
	\subfigure{
	\includegraphics[width=0.102\textwidth]{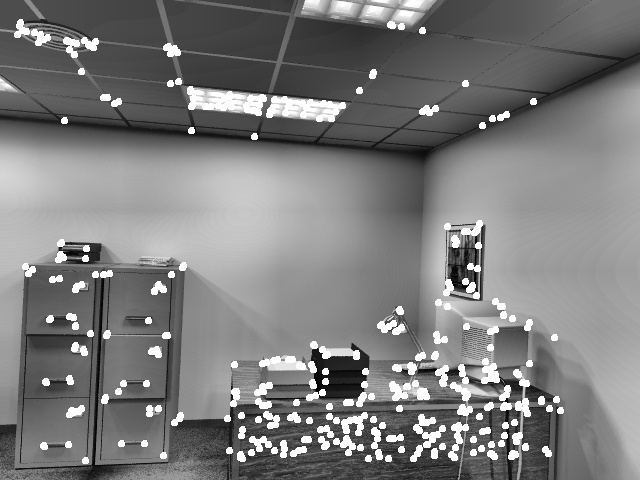}}\hspace{-1mm}
	\subfigure{
	\includegraphics[width=0.102\textwidth]{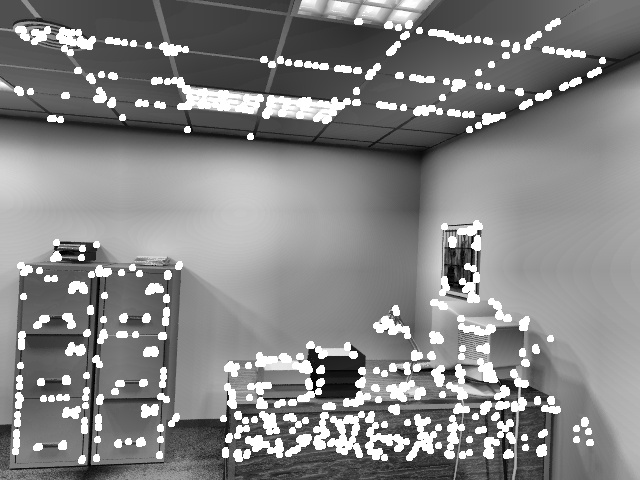}}
	\vspace{-2mm}	
	\quad
	
	\subfigure{
	\includegraphics[width=0.102\textwidth]{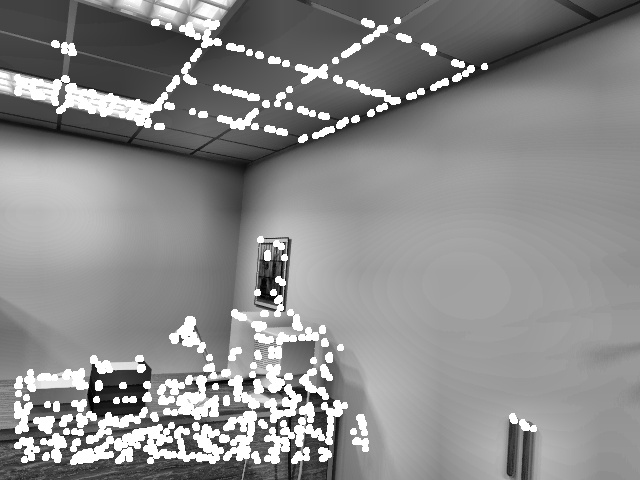}}\hspace{-1mm}
	\subfigure{\
	\includegraphics[width=0.102\textwidth]{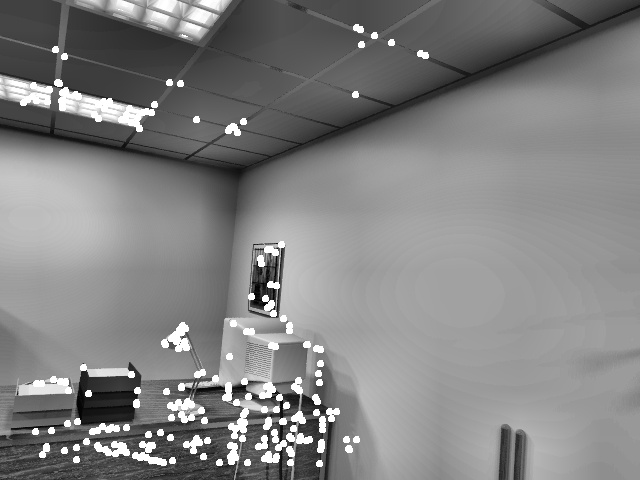}}\hspace{-1mm}	
	\subfigure{
	\includegraphics[width=0.102\textwidth]{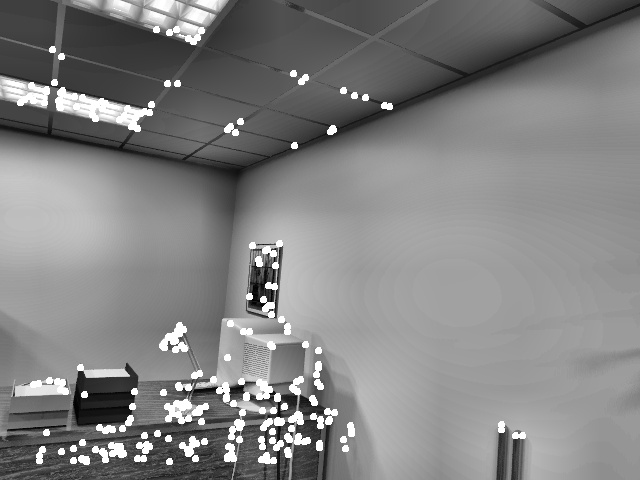}}\hspace{-1mm}
	\subfigure{
	\includegraphics[width=0.102\textwidth]{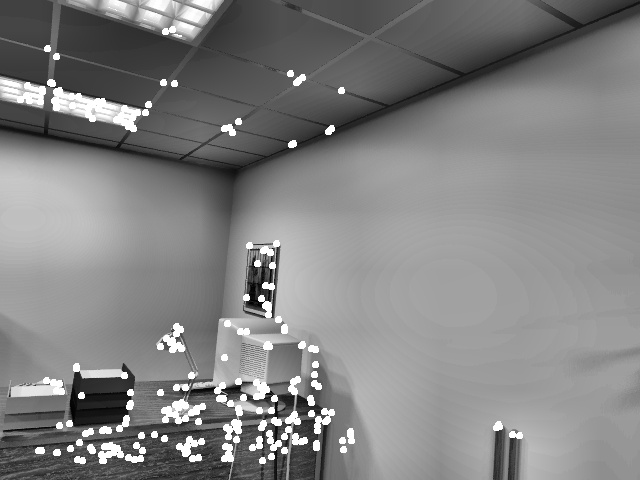}}\hspace{-1mm}
	\subfigure{
	\includegraphics[width=0.102\textwidth]{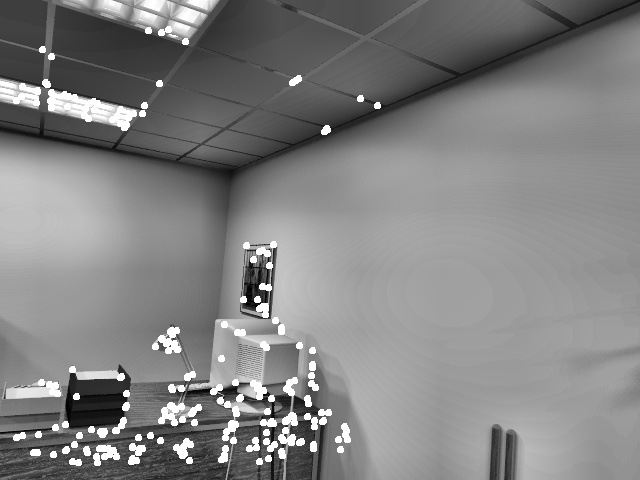}}\hspace{-1mm}
	\subfigure{
	\includegraphics[width=0.102\textwidth]{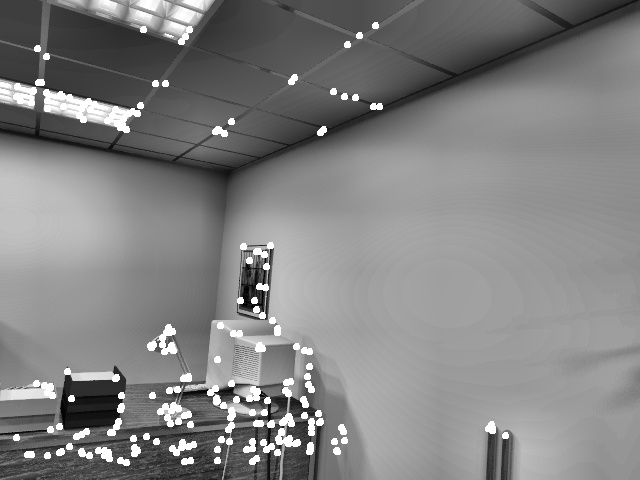}}\hspace{-1mm}
	\subfigure{
	\includegraphics[width=0.102\textwidth]{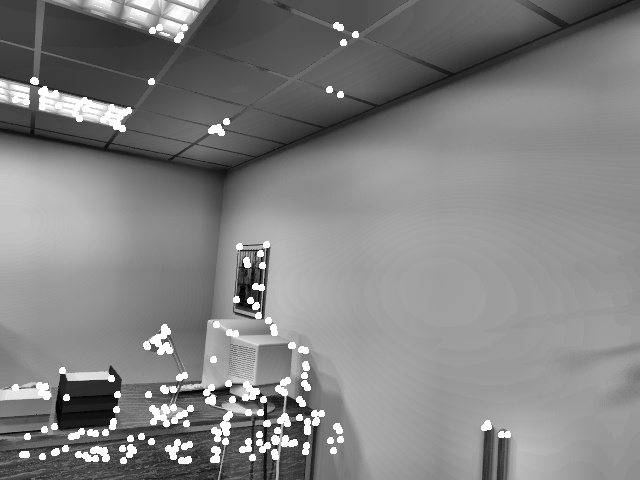}}\hspace{-1mm}
	\subfigure{
	\includegraphics[width=0.102\textwidth]{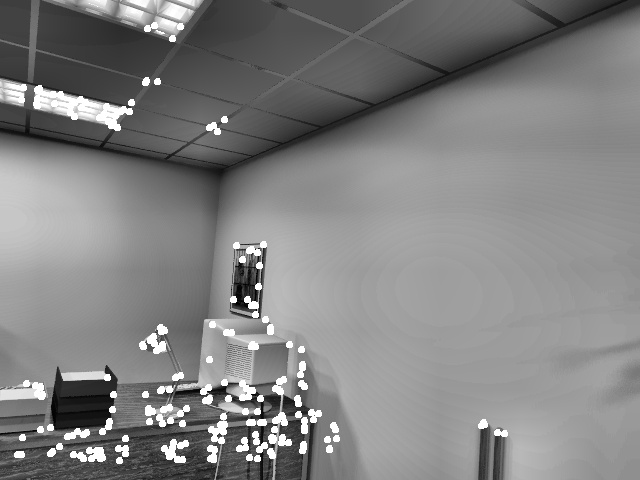}}\hspace{-1mm}
	\subfigure{
	\includegraphics[width=0.102\textwidth]{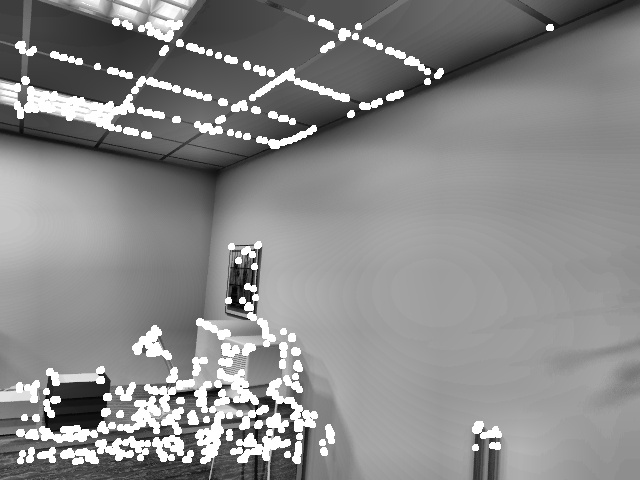}}
	\vspace{-2mm}	
	\quad
	
	\subfigure{
	\includegraphics[width=0.102\textwidth]{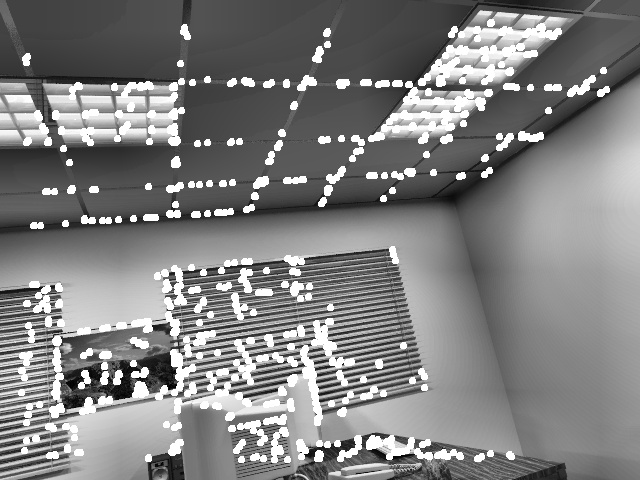}}\hspace{-1mm}
	\subfigure{\
	\includegraphics[width=0.102\textwidth]{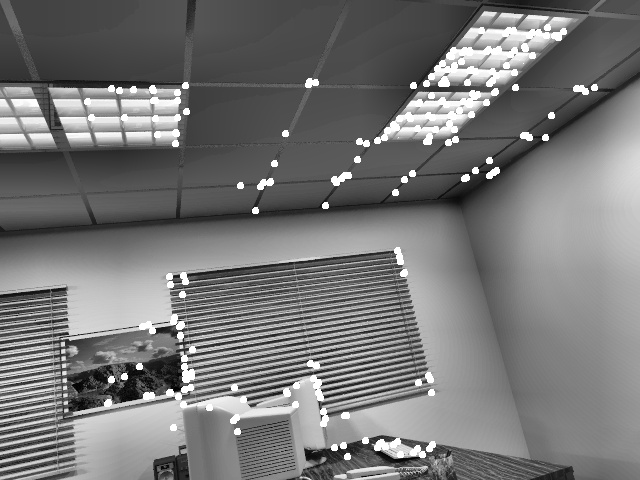}}\hspace{-1mm}	
	\subfigure{
	\includegraphics[width=0.102\textwidth]{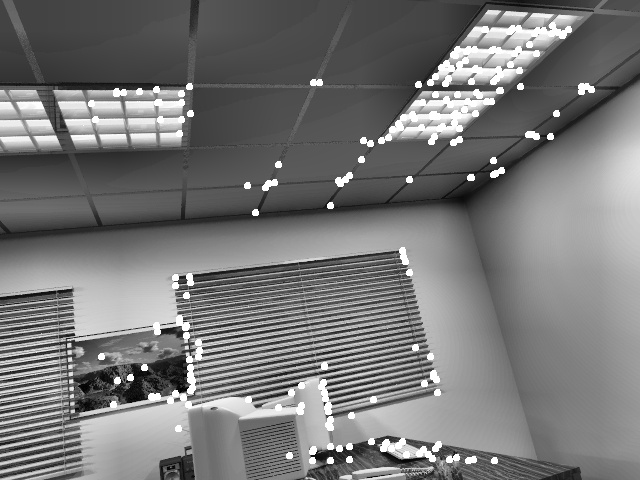}}\hspace{-1mm}
	\subfigure{
	\includegraphics[width=0.102\textwidth]{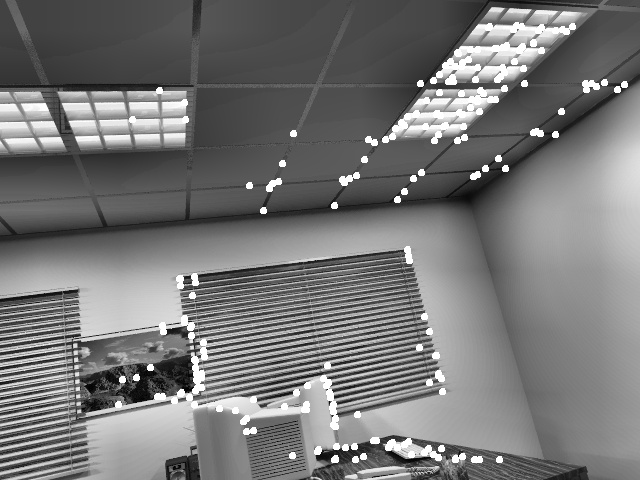}}\hspace{-1mm}
	\subfigure{
	\includegraphics[width=0.102\textwidth]{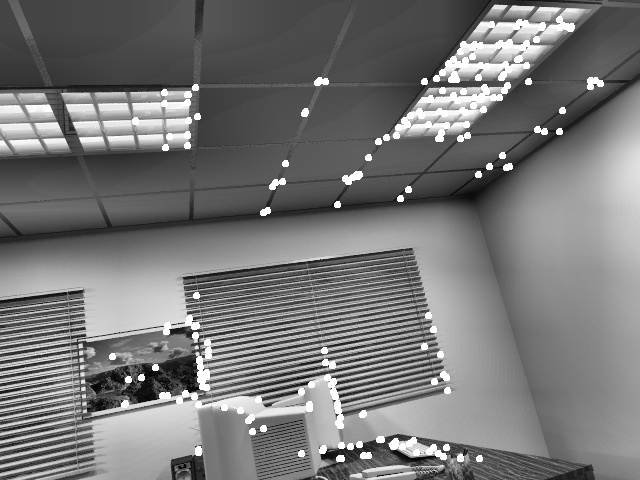}}\hspace{-1mm}
	\subfigure{
	\includegraphics[width=0.102\textwidth]{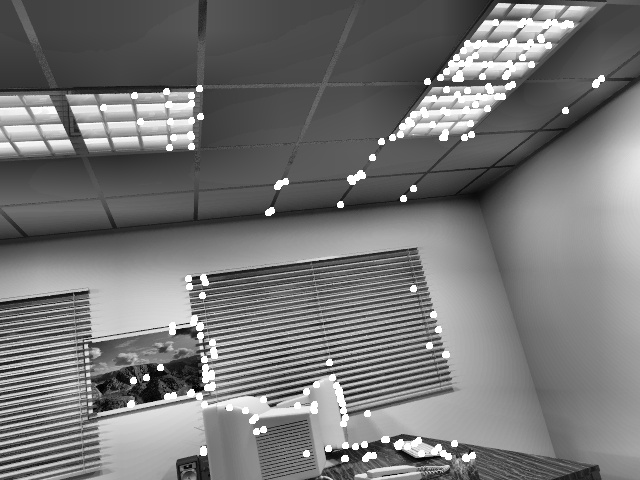}}\hspace{-1mm}
	\subfigure{
	\includegraphics[width=0.102\textwidth]{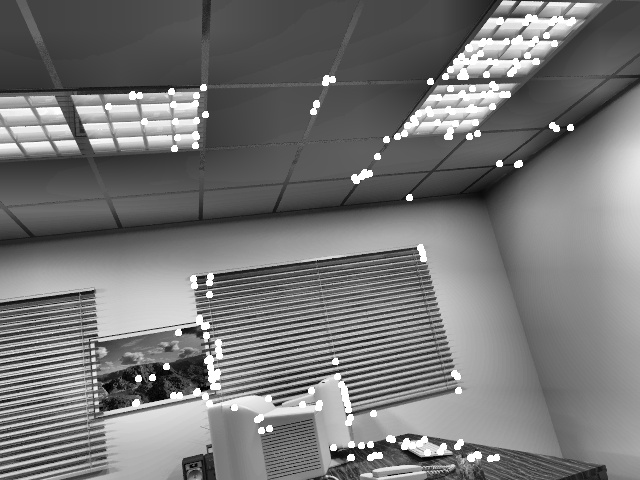}}\hspace{-1mm}
	\subfigure{
	\includegraphics[width=0.102\textwidth]{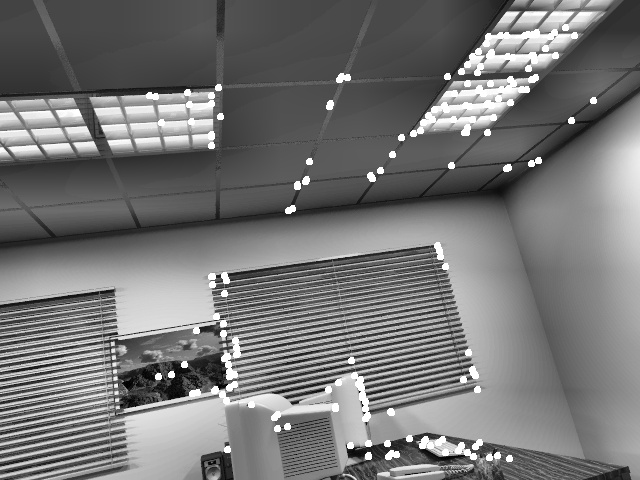}}\hspace{-1mm}
	\subfigure{
	\includegraphics[width=0.102\textwidth]{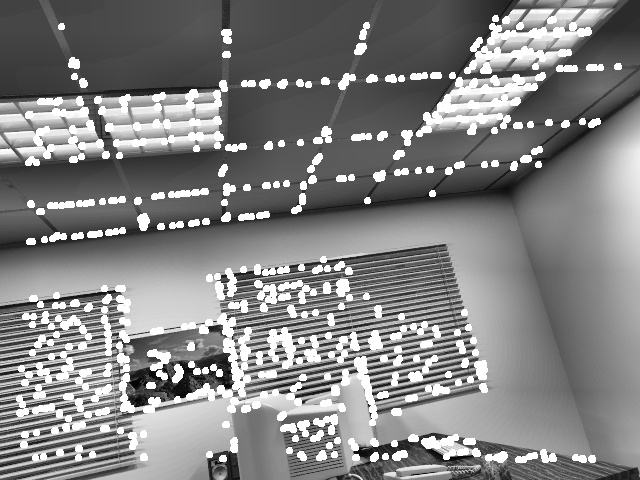}}
	\vspace{-2mm}
	\quad
	
	\subfigure{
	\includegraphics[width=0.102\textwidth]{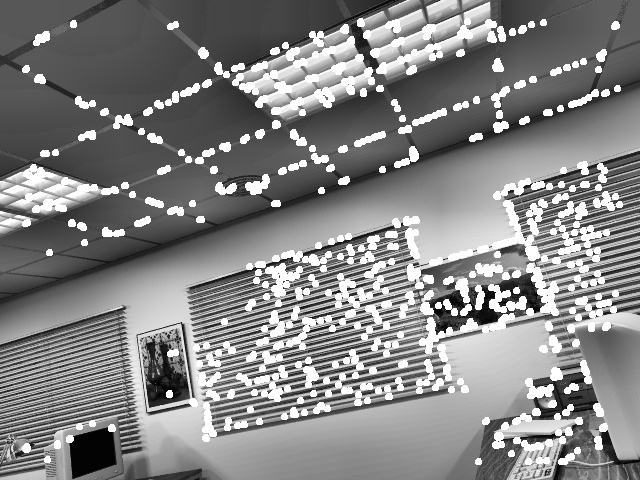}}\hspace{-1mm}
	\subfigure{\
	\includegraphics[width=0.102\textwidth]{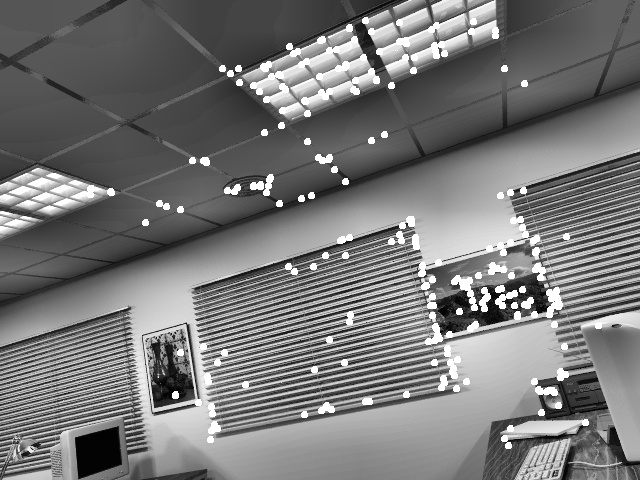}}\hspace{-1mm}	
	\subfigure{
	\includegraphics[width=0.102\textwidth]{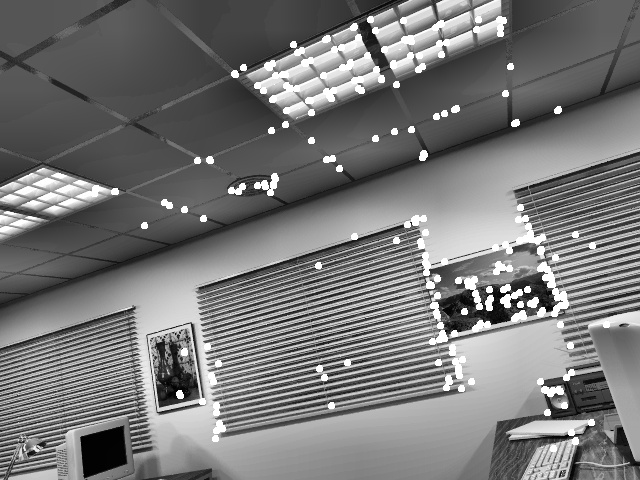}}\hspace{-1mm}
	\subfigure{
	\includegraphics[width=0.102\textwidth]{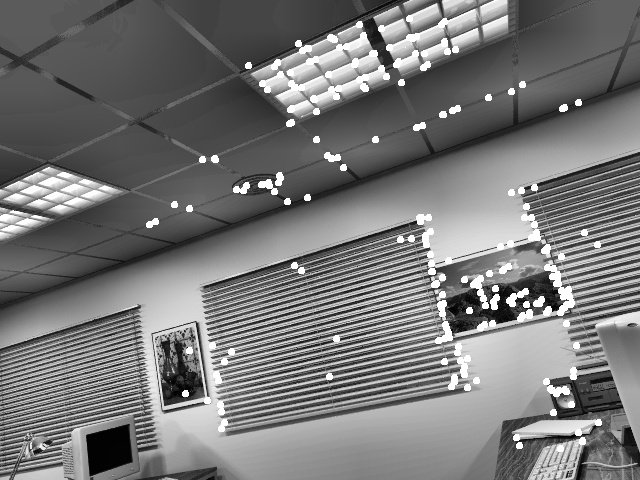}}\hspace{-1mm}
	\subfigure{
	\includegraphics[width=0.102\textwidth]{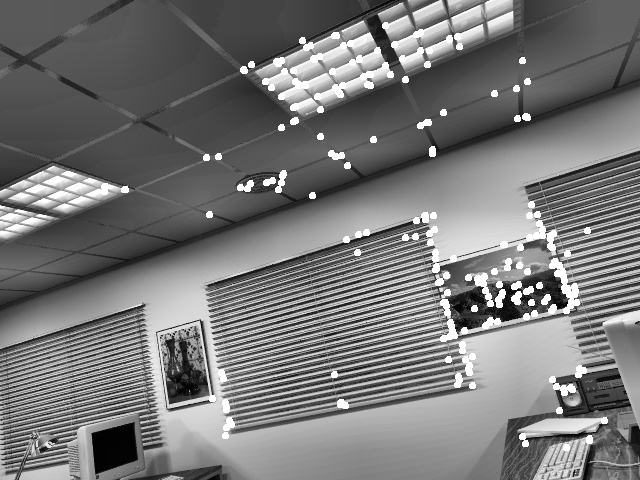}}\hspace{-1mm}
	\subfigure{
	\includegraphics[width=0.102\textwidth]{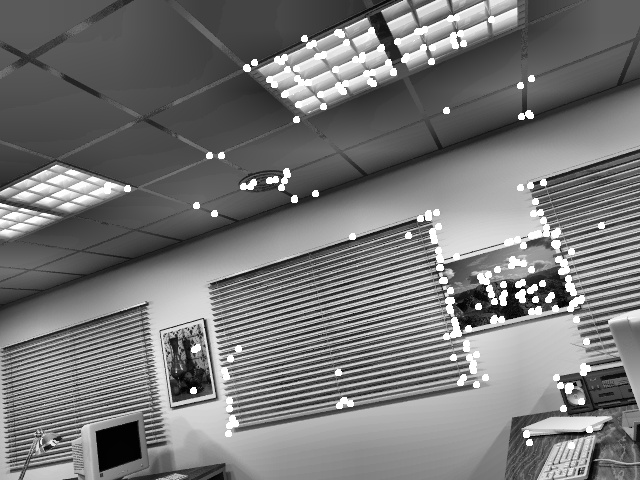}}\hspace{-1mm}
	\subfigure{
	\includegraphics[width=0.102\textwidth]{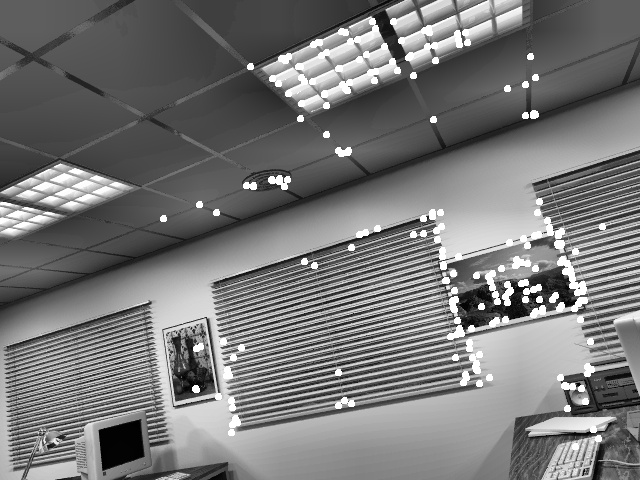}}\hspace{-1mm}
	\subfigure{
	\includegraphics[width=0.102\textwidth]{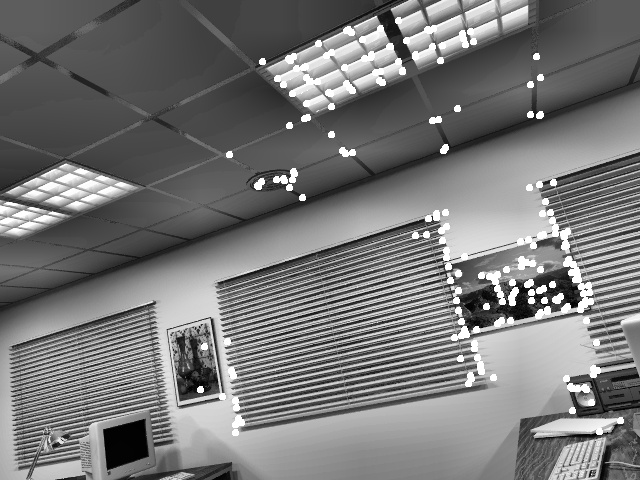}}\hspace{-1mm}
	\subfigure{
	\includegraphics[width=0.102\textwidth]{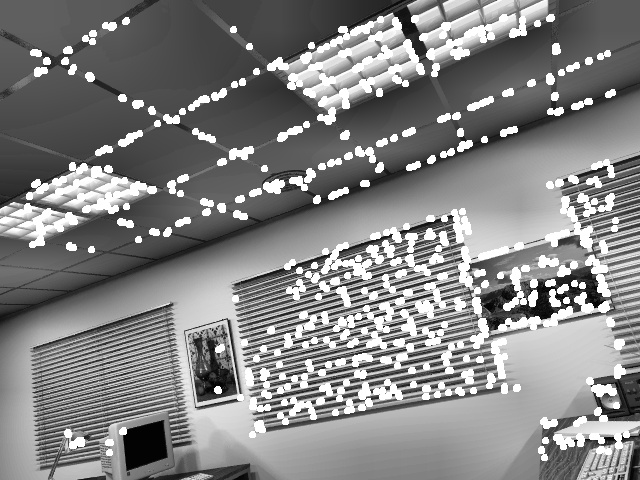}}
	\vspace{-2mm}
	\quad	
	
	\subfigure{
	\includegraphics[width=0.102\textwidth]{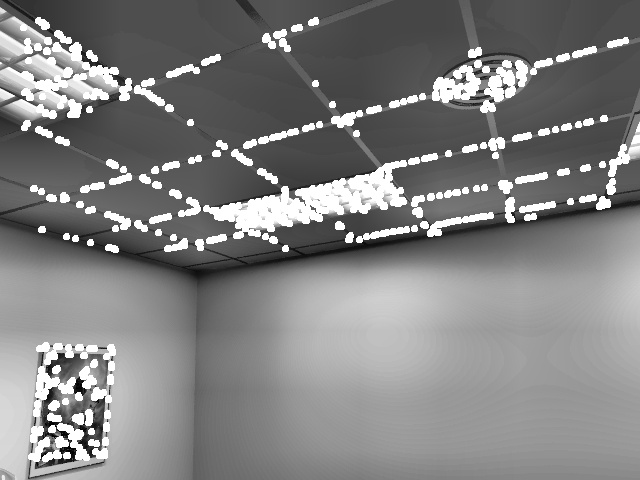}}\hspace{-1mm}
	\subfigure{\
	\includegraphics[width=0.102\textwidth]{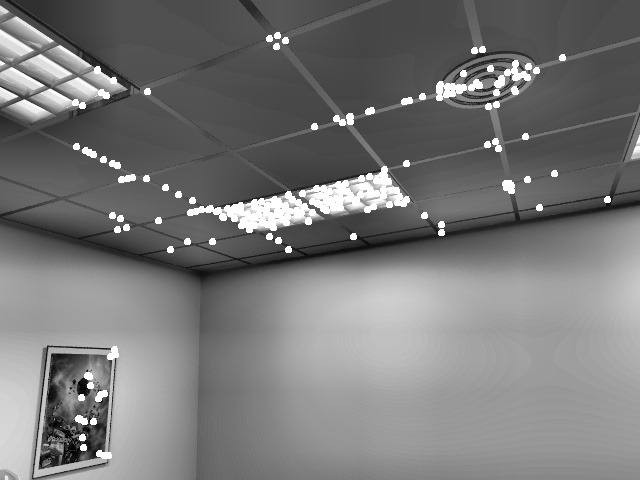}}\hspace{-1mm}	
	\subfigure{
	\includegraphics[width=0.102\textwidth]{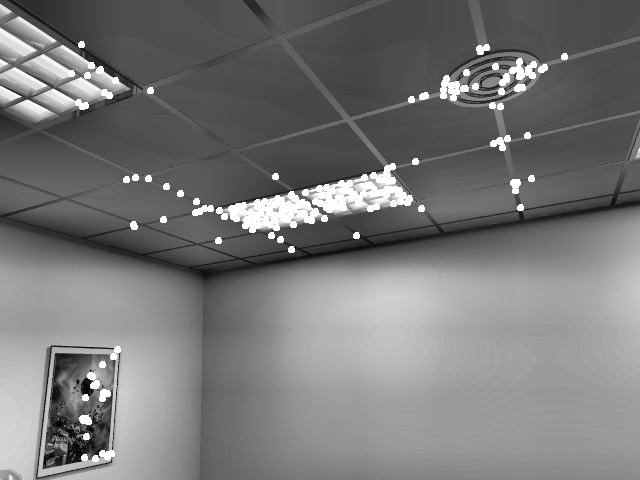}}\hspace{-1mm}
	\subfigure{
	\includegraphics[width=0.102\textwidth]{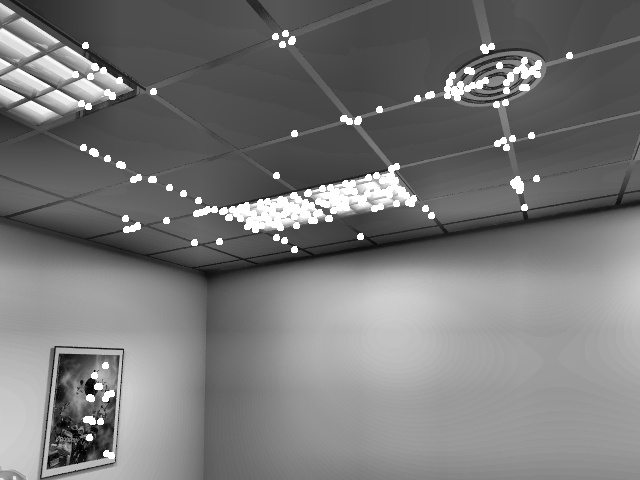}}\hspace{-1mm}
	\subfigure{
	\includegraphics[width=0.102\textwidth]{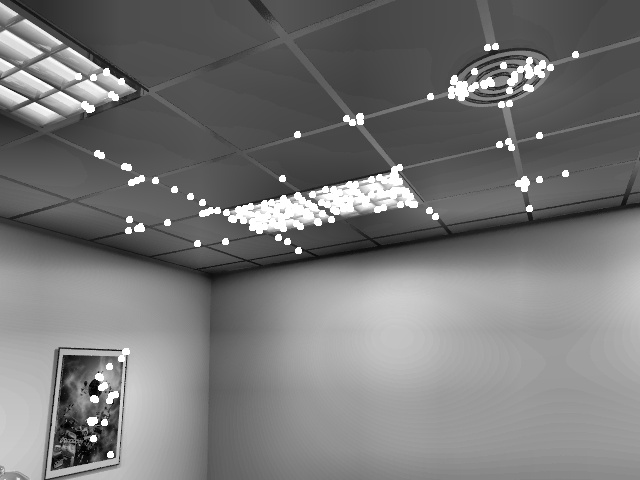}}\hspace{-1mm}
	\subfigure{
	\includegraphics[width=0.102\textwidth]{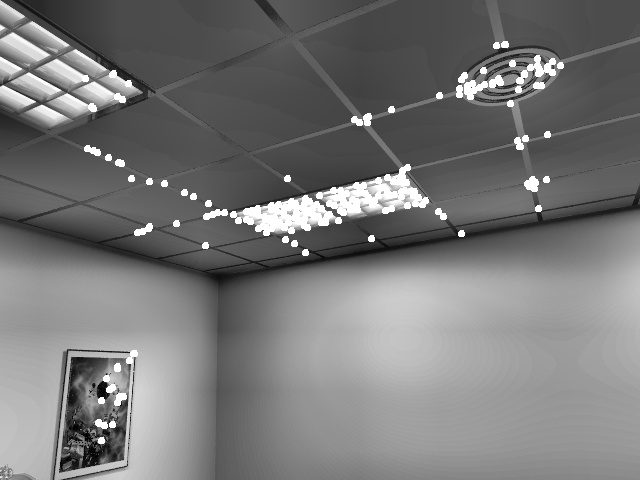}}\hspace{-1mm}
	\subfigure{
	\includegraphics[width=0.102\textwidth]{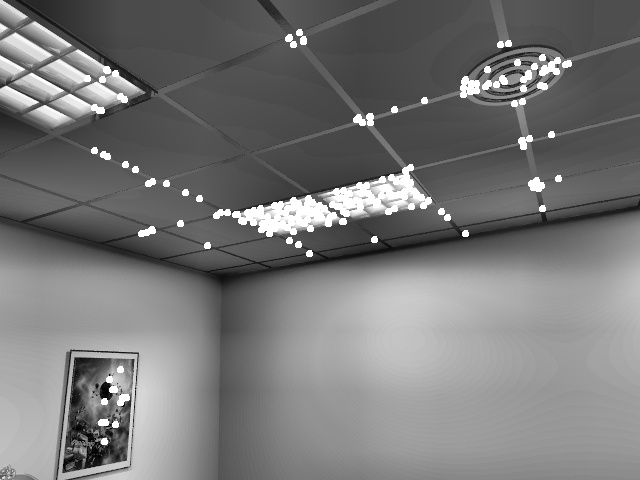}}\hspace{-1mm}
	\subfigure{
	\includegraphics[width=0.102\textwidth]{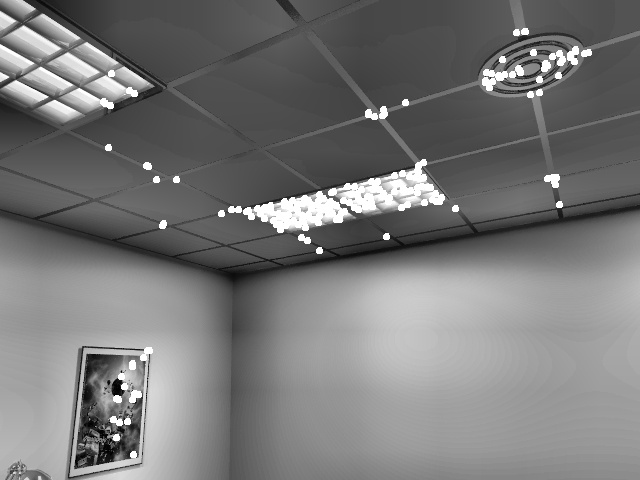}}\hspace{-1mm}
	\subfigure{
	\includegraphics[width=0.102\textwidth]{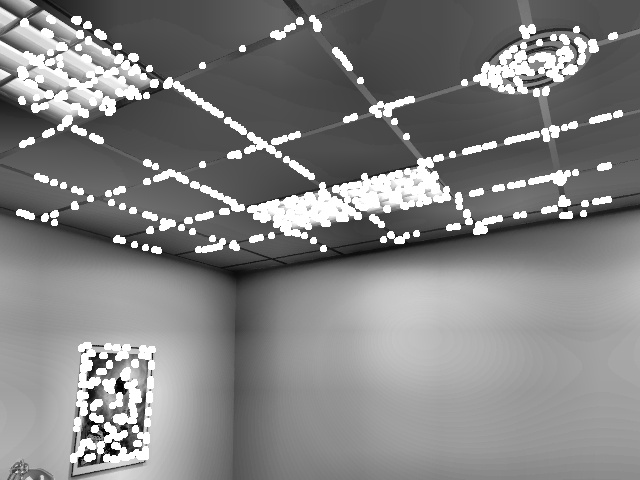}}
	\quad

\caption{Examples for keypoints (inliers) in continuous frames. Before detecting keypoints, incoming images are preprocessed using adaptive histogram equalization to reduce illumination variations \cite{histogram}. Our system can track enough keypoints at any time. Notably, it does not need to extract keypoints if inliers are enough. In this case, it only detects keypoints in the first and the last column. In this sequence (\textit{ICL-NUIM-Office 3}), FastORB-SLAM yields better localization accuracy than ORB-SLAM2 with less computation time, please see Table \ref{ticl} and Fig. \ref{iclfigure}. }\label{keypointshow}
\vspace{-3mm}
\end{figure*}

\noindent \textbf{Tracking:} This thread outputs real-time pose results and provides the observation information between frames for the remaining two threads. Each frame is pre-processed by the adaptive histogram equalization algorithm \cite{histogram} to reduce the effects of illumination variation (see Fig. \ref{equalize}). Next, keypoints are detected with the improved ORB algorithm \cite{orbslam2} and an initial camera pose is predicted with the proposed UAM model. Keypoint correspondences are then established via two stage (coarse-to-fine) descriptor-independent matching (details in Section \ref{sparsematching}). Similar to other SOTA methods such as ORB-SLAM2 and SVO, the keypoint correspondences are searched from the last frame, the nearest keyframe, and local Map. 
Once the correspondences are established, the pose estimate is refined with the BA optimization. Compared with ORB-SLAM2, efficiency of this thread comes from two aspects: there is no need to compute keypoint descriptors or detect keypoints when the inlier number is high (300), as shown in Fig. \ref{keypointshow}. \par

After obtaining the camera pose of the current frame, the system judges whether the current frame is a keyframe i.e. it is over 20 frames apart from the rest, has at least 50 keypoints, or the \textit{Local mapping} thread is idle. \par 

\noindent \textbf{Local Mapping:} This thread is activated when a frame is selected as a keyframe. The system computes keypoint descriptors for the keyframe and the thread looks for previous keyframes that connect to the current one in the \textit{Covisibility graph} based on descriptor matching. All mappoints that are seen by those keyframes are also extracted. Finally, the corresponding graph structure is created in \textit{Covisibility graph}. Next, the current keyframe and connected keyframe poses are optimized to smooth local drift errors. Note that this thread only optimizes the keyframe poses that are observed by the current frame, which is a local BA process. After that, redundant keyframes are discarded to control the graph size. \par

\noindent \textbf{Loop Closure:} After \textit{Local Mapping}, this thread is activated to look for a possible loop closure to eliminate global drift errors. In this work, we follow ORB-SLAM2 and adopt the DBoW2 model \cite{bags} to search and measure the similarity between the newest keyframe and other remaining keyframes. If loop closure is established, a spanning tree that contains all nodes (keyframes) is created. The spanning tree is a minimum connected representation of the graph where each node is connected to a sole parent node and a sole child node. Finally, an \textit{essential graph} is created according to the spanning tree and the global BA is applied to optimize the \textit{essential graph}. Whereas ORB-SLAM2 detects a loop by loading a text-format dictionary \cite{bags} ($\backsim$ 3000 ms) in advance, the proposed FastORB-SLAM converts the dictionary to a binary format for quick upload on startup ($\backsim$30 ms).

\noindent \textbf{Camera Motion Model:} 
Camera motion in visual SLAM is considered as a three-dimension rigid motion including 3-DoF in translation and 3-DoF in rotation. 
We represent the motion group $\texttt{SE}(3)$ as:
\begin{equation}
\texttt{SE}(3) = \left\{ \textbf{T}=\begin{bmatrix}\textbf{R} & \textbf{t} \\0 & 1 \end{bmatrix}  \in \mathbb{R}^{4*4} \right\},
\label{SE}
\end{equation}
where $\textbf{T}$ denotes the camera pose, $\textbf{R} \in SO(3)$ and $\textbf{t} \in \mathbb{R}^{3}$ denote rotation and translation respectively. The Lie algebra \texttt{se}(3) and the vector tangents are formed by elements of the type: 
\begin{equation}
 \bm{\xi}=\begin{bmatrix} \bm{\phi}  \\ \bm{\rho}  \end{bmatrix} \in \mathbb{R}^6, \bm{\xi}^{\wedge} = \begin{bmatrix}[ \bm{\phi} ]_{\times} & \bm{\rho} \\0 & 0 \end{bmatrix} \in \texttt{se}(3),
\label{lie1}
\end{equation}
where $\bm{\phi}$ and $\bm{\rho}$ denotes the rotation vector and the translation vector, respectively. $[\bm{\phi}]_{\times}$ denotes the skew-symmetric matrix of $\bm{\phi}$. In particular, for every $\textbf{T} \in \texttt{SE}(3)$ there exist twist coordinates $\bm{\xi}=[\bm{\phi}, \bm{\rho}]^{T} \in \mathbb{R}^6$ such that:
\begin{equation}
 \textbf{T}=\texttt{exp}(\bm{\xi}^{\wedge}). 
 \label{e:exp}
\end{equation}\par

Let $\textbf{T}_{cw}$ and $\textbf{T}_{rw}$ be the current (frame) pose and reference (previous) pose in the world coordinate system. Let $\textbf{T}_{rw}^{cw}$ be a relative motion transformation between the two poses. $\textbf{T}_{cw},\textbf{T}_{rw}$, and $\textbf{T}_{rw}^{cw}$ belong to the Lie group $\texttt{SE}(3)$ which is a smooth manifold such that the multiplication: $\texttt{SE}(3) \cdot \texttt{SE}(3)\rightarrow \texttt{SE}(3)$. We have:
\begin{equation}
\textbf{T}_{cw} = \textbf{T}_{rw}^{cw}\textbf{T}_{rw}  , \textbf{T}_{rw}^{cw} = \textbf{T}_{cw}\textbf{T}_{rw}^{-1}.
\label{trans}
\end{equation}
The camera motion is coded using the Sophus library.

\section{Descriptor-independent keypoint matching method} \label{sparsematching}

\begin{algorithm}[tp]
  \caption{Descriptor-Independent Keypoint Matching with the UAM model.}
  \label{algorithm}
  \begin{algorithmic}[1]
    \Require    
    Reference frame $I_{r}$, A keypoint on the reference frame $I_{r}(x,y)$; current frame $I_{c}$; Last three frame poses $\textbf{T}_{cw-1}$, $\textbf{T}_{cw-2}$, and $\textbf{T}_{cw-3}$;           
    \Ensure
    Movement vector $\textbf{m}$ and keypoint correspondence $I_{r}(x,y)\leftrightarrow I_{c}(x+dx,y+dy)$ ;
    \State Model camera motion with the UAM model, then predict the relative motion translation matrix $\textbf{T}_{cw}^{cw-1}$ by Eq.~(\ref{velocitypr}) based on $\textbf{T}_{cw-1}$, $\textbf{T}_{cw-2}$, and $\textbf{T}_{cw-3}$.  
    \State According to the predicted $\textbf{T}_{cw}^{cw-1}$, predict current frame pose $\textbf{T}_{cw}^{\star}$ by Eq.~(\ref{poseprediction}), and then cast $\textbf{T}_{cw}^{\star} \in \texttt{SE}(3) \rightarrow \textbf{T}^{\star} \in \mathbb{\bm{R}}^{3\times4}$;
    \State According to $\textbf{T}^{\star}$, predict an initial keypoint correspondence $I_{c}(x^{\star}, y^{\star})$ by Eq.~(\ref{pro});
    \State Solve the movement vector $\textbf{m}$ with \textbf{Algorithm \ref{pyramidalgorithm}}; \\
    \Return $\textbf{m}$ and $I_{r}(x,y)$ $\rightarrow I_{c}(x+dx,y+dy)$;   
  \end{algorithmic}  
\end{algorithm}

Two adjacent frames in a time-varying sequence have two characteristics: small parallax angle and brightness invariance. Based on these assumptions, we propose a two-stage descriptor-independent keypoint matching method to establish reliable keypoint correspondences. This proposed method is divided into two stages: 
\par

	\textbf{Stage one } is implemented by \textbf{Algorithm \ref{algorithm}} for all keypoints to robustly establish coarse keypoint correspondences:
	\begin{enumerate}
	\item Predict keypoint correspondences with the UAM model giving the algorithm a good initial guess and potentially reducing the search space and computations; 
	\item Establish keypoint correspondences in an eight-level pyramid structure based on the sparse optical flow algorithm. 
	\end{enumerate}\par
	
	\par
	\textbf{Stage two} is for inlier refinement:
	\begin{enumerate}	
	\item Leverage camera motion smoothness constraint to filter outliers;
	\item Adopt the RANSAC-based strategy to further refine the keypoint correspondences.
	\end{enumerate}\par 
	
Compared with the classical sparse optical flow algorithm, KLT\cite{KLT}, our descriptor-independent method is designed for the indirect SLAM problem, and focuses on keypoint matching instead of direct velocity vector estimation. Specifically, we use 
the UAM model to predict the initial correspondences and solve for the movement vector in an eight-level image pyramid for improved robustness. Whereas KLT does not include an outlier filtering process, we add outlier removal in our proposed solution for SLAM. Specifically, we exploit grid-based motion statistics \cite{gms2} to filter outliers that do not conform to the motion smoothness constraint. This makes our method suitable for the SLAM problem. \par

Recent advances improve performance of optical flow by leveraging deep networks, however, they do not give real time performance on machines with limited computing power, e.g., Selflow \cite{selflow}, Flownet \cite{flownet}, and Liteflownet \cite{liteflownet}. Moreover, some SLAM methods use optical flow simply to recognize moving objects and improve the localization accuracy in dynamic environments \cite{zhang20, cheng19, yu18}. In contrast, our method uses optical flow to address the keypoint matching problem instead of the velocity estimation or the moving object recognition problem, and gives real time performance on a 1.10 GHz CPU.
\par
Our method is largely descriptor-independent and extracts descriptors only when the current frame is selected as a keyframe. This makes our SLAM system efficient as shown in Fig. \ref{system}. In the rest of this section, we formulate the descriptor-independent keypoint matching problem, and then describe the specific steps in detail.

\begin{figure}[tb]
\centering  
	\includegraphics[width=0.45\textwidth]{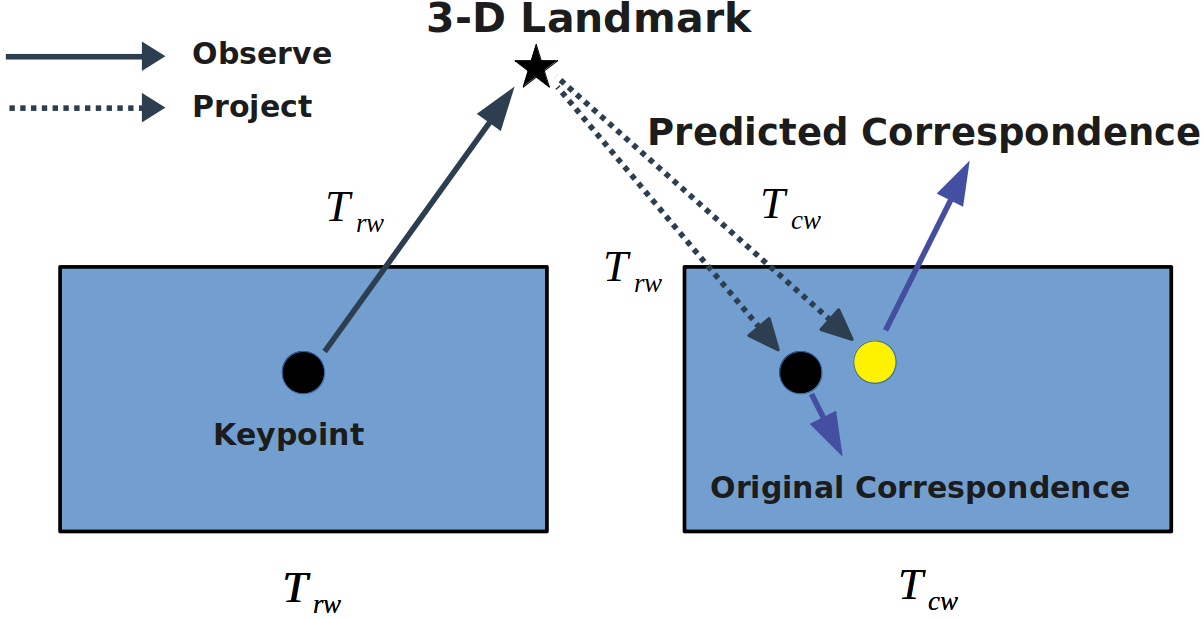}
	\vspace{-3mm}
\caption{Illustration of predicting a keypoint correspondence by projecting a 3D landmark to the current frame. Given a reference frame pose $\textbf{T}_{rw}$, a predicted current frame pose $\textbf{T}_{cw}$, a keypoint on the the reference frame and its 3D landmark, the predicted (initial) correspondence on the current frame can be obtained with the projection model. Note that in this case (see Section \ref{motionmodel1}), $\textbf{T}_{cw} =\textbf{T}_{cw}^{\star}$, $\textbf{T}_{rw}= \textbf{T}_{cw-1}$.}\label{predict}
\vspace{-3mm}
\end{figure}

\vspace{-3mm}
\subsection{Problem Model} \label{foundation}
In this work, the goal of the descriptor-independent keypoint matching method is defined as:
\par 
\noindent \textbf{Goal:} given a keypoint $(x,y)$ in a reference frame $I_{r}$, find its corresponding location $(x+dx, y+dy)$ in the current frame $I_{c}$, or equivalently, find the movement vector $\textbf{m} = (dx, dy)$. Thus a correspondence can be established by:
\begin{equation}
I_{r}(x,y)\leftrightarrow I_{c}(x+dx,y+dy).
\label{keycorrespondences}
\end{equation}

\noindent \textbf{Theoretical Foundation:} The proposed matching method works on two assumptions \cite{KLT}:
\begin{itemize} 
	\item Grayscale invariance:   Pairs of matched keypoints between adjacent frames should share the same brightness/grayscale value.
	\item Neighborhood motion consistency: All points in the neighborhood of a keypoint must have consistent motion.
\end{itemize}  \par
Let us define $I_{x,y,t}$ to be the grayscale value of a keypoint coordinate $(x,y)$ at $t$ in the first frame. After $dt$ time, the keypoint moves to $(x+dx,y+dy)$ in the next frame. We introduce the process of the problem formulation based on the above two assumptions. \par
\noindent \textbf{Assumption 1}: for grayscale invariance we have: 
\begin{equation}
\left[\begin{matrix}\textbf{I}_{x}&\textbf{I}_{y}\end{matrix}\right] \left[ \begin{matrix} \textbf{u} \\ \textbf{v} \end{matrix}\right] = -\textbf{I}_{t},
\label{op5}
\end{equation}
where $\textbf{u} = dx/dt$, $\textbf{v} = dy/dt$, $\textbf{I}_{x}=\partial I/ \partial x$, $\textbf{I}_{y}=\partial I/ \partial y$, $ \textbf{I}_{t}={\partial I}/{\partial t}$. $\textbf{I}_{x}$, $\textbf{I}_{y}$ and $\textbf{I}_{t}$ are the known image gradients along the $x$, $y$ and $time$ axes, respectively. Thus, the keypoint matching problem translates into solving keypoint movement over time. However, this equation cannot be solved as it has two unknown variables $\textbf{u}$ and $\textbf{v}$. Hence, we need the second assumption. \par

\noindent \textbf{Assumption 2}: The neighborhood motion consistency assumption means that all pixels in an $\omega * \omega$ patch around the keypoint must have consistent movement:
\begin{equation}
\left[ \begin{matrix} \textbf{I}_{x} & \textbf{I}_{y} \end{matrix} \right]_{k} \left[ \begin{matrix} \textbf{u} \\ \textbf{v} \end{matrix}\right] = -(\textbf{I}_{t})_{k}, k=1,...,\omega^{2}.
\label{aaa}
\end{equation}
\par Eq.~(\ref{aaa}) is over-determined. The problem of keypoint matching boils down to solving for two unknowns $(\textbf{u}, \textbf{v})$ given $k$ equations using least squares fitting. Note that considering $t$ is a fixed scalar between two adjacent frames e.g., Kinect 1 captures images at $30$ Hz, which means $t=1/30$ s. In this work, we consider
$ \textbf{m}=(\textbf{u}, \textbf{v}) = (dx, dy)$.

\vspace{-3mm}
\subsection{Correspondence Predication with UAM}
\label{motionmodel1}
Recall the \textbf{Goal} in last Section. Given a good initial guess to solve the movement vector $\textbf{m}$ (see Fig. \ref{predict}), it is not only able to improve the robustness of keypoint tracking, but also potentially reduce the iterative optimization computation. Based on this, we predict the keypoint correspondences in the current frame with the UAM model.  \par

Previous methods generally use a motion model to mitigate the delay or fuse information from multiple sensors. Specifically, a motion model like Kalman filter is used to predict the vehicle pose when no pose estimation is available, enabling the system to obtain a smoother trajectory. For example, Feng \textit{et al.} \cite{feng} used two motion models including an approximate uniform motion and approximate uniform acceleration to fuse IMU and ultra-wideband measurement information for smooth motion trajectory estimation and reduced jitter in positioning data. Martínez-Carranza \textit{et al.} \cite{combining} used a motion model-based stochastic filter to predict vehicle pose in those frames where no pose estimate is available due to differences in the sensor measurement frequencies. In contrast, we use a single sensor (RGB-D camera) instead of multi-sensor inputs, and predict the initial pose of the current frame with the UAM model for keypoint matching instead of direct pose estimation.

\begin{figure}[tbp]
\centering  
	\includegraphics[width=0.28\textwidth]{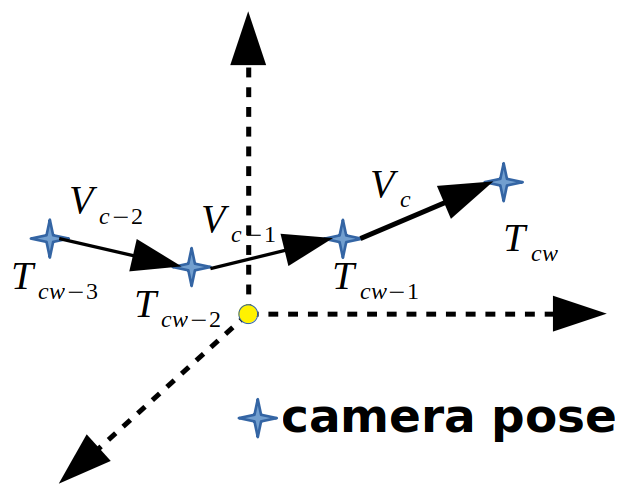}
	\vspace{-3mm}
\caption{Illustration of the UAM model used to predict the current pose $\textbf{T}_{cw}$ with the predicted velocity $\textbf{V}_{c}$.}\label{motionmodel}
\vspace{-5mm}
\end{figure}

\par 
As shown in Fig.~\ref{motionmodel}, we first cast the camera motion to the UAM model, and then estimate the velocity between the reference (last) frame and the current frame. Next, the current frame pose is predicted via the velocity. Finally, initial keypoint correspondences are obtained by projecting the 3D landmarks that are observed by the keypoints in the reference frame to the current frame using the predicted current camera pose (see Fig. \ref{predict}). The specific operations are given below: \par
\textbf{Notation:} Let $\textbf{T}_{cw-3}, \textbf{T}_{cw-2}, \textbf{T}_{cw-1}, \textbf{T}_{cw} \in \texttt{SE}(3)$ be four sequential adjacent frame poses in the world coordinate system, where $\textbf{T}_{cw}$ denotes the current frame pose. Let $\textbf{T}_{cw}^{cw-1} \in \texttt{SE}(3)$ be the relative motion transformation matrix between $\textbf{T}_{cw-1}$ and $\textbf{T}_{cw}$. Let $\textbf{V}_{c}, \textbf{V}_{c-1}$ and $\textbf{V}_{c-2}$ denote the average velocities from $\textbf{T}_{cw-1}$ to $\textbf{T}_{cw}$, from $\textbf{T}_{cw-2}$ to $\textbf{T}_{cw-1} $, and from $\textbf{T}_{cw-3}$ to $\textbf{T}_{cw-2}$, respectively, as shown in Fig.~\ref{motionmodel}.
\par
{\textbf{Motion Model:} We first define the camera velocity before describing the motion model. The velocity $\textbf{V}$ in time $t$ can be given in multiple view geometry by introducing the twist coordinates 
\begin{equation}
{\textbf{V}(t)}^{\wedge}= \begin{bmatrix}[ \bm{\phi}(t) ]_{\times} & \bm{\rho} (t) \\0 & 0 \end{bmatrix} \in \texttt{se}(3),
\label{velocityalgebra}
\end{equation}
or equally, we follow the definition in \cite{lie}:
\begin{equation}
\bm{\xi}^{\wedge} =(\textbf{V}t)^{\wedge}=\textbf{V}^{\wedge}t \in \texttt{se}(3).
\label{velocity11}
\end{equation}
Now, we assume that the camera motion conforms to} the UAM model as shown in Fig.~\ref{motionmodel}. This model assumes that the acceleration is constant i.e. the increment between two adjacent velocities is assumed to be equal. We have:
\begin{equation}
\textbf{V}_{c-1} \otimes \textbf{V}_{c}=\textbf{V}_{c-1} \otimes \textbf{V}_{c-2},
\label{e:acceleV}
\end{equation}
where $\otimes$ represents the computation of increment between two velocities. Further,
observe that  $\textbf{T}_{cw}^{cw-1}$ essentially denotes the displacement between $\textbf{T}_{cw-1}$ and $\textbf{T}_{cw}$. With Eq.~(\ref{e:exp}), mapping the velocity to the displacement by the exponential map:
\begin{equation}
\texttt{exp}(\textbf{V}_{c}^{\wedge} \delta t) = \textbf{T}_{cw}^{cw-1} =\textbf{T}_{cw}\textbf{T}_{cw-1}^{-1} \in \texttt{SE}(3). 
\label{velocitydefine}
\end{equation}
where $\delta t$ denotes the elapsed time. Observe that $\delta t$ is typically considered as a fixed scalar in visual SLAM. For instance, we use Kinect 1 as the RGB-D camera whose frame rate is $30$ Hz, or $\delta t=1/30 \approx 0.033$ s.  Since $\delta t$ is fixed and small in magnitude, within the context of robot navigation, Eq.~(\ref{e:acceleV}) can be transformed into the following:
\begin{equation}
\textbf{T}_{cw-1}^{cw-2} \otimes \textbf{T}_{cw}^{cw-1}=\textbf{T}_{cw-2}^{cw-3} \otimes \textbf{T}_{cw-1}^{cw-2}.
\end{equation}
Now, the problem of velocity prediction changes to predicting the relative transformation matrix between the current and reference frame. The relative transformation matrix belongs to $\texttt{SE}(3)$, with Eq.~(\ref{trans}), the increment operation can be computed by:
\begin{equation}
\begin{aligned}
&\textbf{T}_{cw-1}^{cw-2} \otimes \textbf{T}_{cw}^{cw-1} = \textbf{T}_{cw}^{cw-1} {\textbf{T}_{cw-1}^{cw-2}}^{-1}\\
&\textbf{T}_{cw-2}^{cw-3} \otimes \textbf{T}_{cw-1}^{cw-2} = \textbf{T}_{cw-1}^{cw-2} {\textbf{T}_{cw-2}^{cw-3}}^{-1}\\
\end{aligned}
\end{equation}
where $\textbf{T}_{cw-1}^{cw-2} = \textbf{T}_{cw-1}\textbf{T}_{cw-2}^{-1}$ and $\textbf{T}_{cw-2}^{cw-3} = \textbf{T}_{cw-2}\textbf{T}_{cw-3}^{-1}$. 
\noindent Combining the above equations, we have:
\begin{equation}
\begin{aligned}
\textbf{T}_{cw}^{cw-1} 
			&= \textbf{T}_{cw-1}\textbf{T}_{cw-2}^{-1} \textbf{T}_{cw-3}\textbf{T}_{cw-2}^{-1} \textbf{T}_{cw-1}\textbf{T}_{cw-2}^{-1} 	\\
			&=\textbf{T}_{cw-1}^{cw-2} (\textbf{T}_{cw-2}^{cw-3})^{-1}\textbf{T}_{cw-1}^{cw-2}\\	
\end{aligned}.
\label{velocitypr}
\end{equation}
Such a compact computation makes the implementation easy in terms of engineering and proves very effective in our experiments.
 

\noindent \textbf{Correspondences Prediction:} Once the predicted transformation matrix $\textbf{T}_{cw}^{cw-1}$ is estimated, an initial current frame pose $\textbf{T}_{cw}^{\star}$ can be predicted by:
\begin{equation}
\textbf{T}_{cw}^{\star}=\textbf{T}_{cw}^{cw-1}\textbf{T}_{cw-1}.
\label{poseprediction}
\end{equation}\par

For intuitive understanding, we use a $3\times4$ matrix to denote the transformation instead of $\texttt{SE}(3)$, which is $\textbf{T}_{cw}^{\star} \in \texttt{SE}(3) \rightarrow \textbf{T}^{\star} = \begin{bmatrix}\textbf{R}, \textbf{t} \end{bmatrix} \in \mathbb{R}^{3\times4}$, where $\textbf{R}$ is a $3\times 3$ rotation matrix, $\textbf{t}$ is a $3\times 1$ translation matrix. \par

As shown in Fig. \ref{predict}, given a keypoint $I_{r}(x, y)$ on the reference frame and a 3D landmark observed by $I_{r}(x, y)$ in the world coordinate system, by projecting the 3D landmark $(X,Y,Z)$ to the current frame with $\textbf{T}$, an initial guess of keypoint correspondence $I_{c}(x^{\star}, y^{\star})$ on the current frame can be obtained:
\begin{equation}
\textbf{p}^{\star} \propto \textbf{K}\textbf{T}^{\star}\textbf{P} = s\textbf{K}\textbf{T}^{\star}\textbf{P},
\label{pro}
\end{equation} 
where $\textbf{p}^{\star}=[x^{\star}, y^{\star}, 1]^{T}$, $\textbf{p}=[x,y,z,1]^{T}$, $s$ represents the scale factor and $\textbf{K}$ denotes the Camera Intrinsic Matrix which is computed in advance. For all keypoitns, the initial predicted correspondences can be obtained by Eq.~(\ref{pro}). 



\vspace{-3mm}
\subsection{Movement Vector based on an Eight-level Image Pyramid}
 \label{lkmp}

\begin{algorithm}[tb]
  \caption{ Solve for Movement Vector Based on an Eight-Level Image Pyramid.}
  \label{pyramidalgorithm}
  \begin{algorithmic}[1]
    \Require
    Reference frame $I_{r}$, current frame $I_{c}$, a keypoint $I_{r}(x,y)$ on the reference frame, 
    initial (predicted) keypoint correspondence $I_{c}(x^{\star}, y^{\star})$ on the current frame;      
    \Ensure
      The movement vector $\textbf{m}$ to establish Eq.~(\ref{keycorrespondences});
    \State Describe $I_{c}$ and $I_{r}$ in an eight-level Pyramid: $L_{1},...,L_{m}$, scale ratio = $1.2$, where $L_{m} = 8$ denotes the deepest layer;
    \State Compute movement vector $\textbf{m}^{L_{m}}$ at $L_{m}$ via an iterative KLT method for the iterative optimization process:
	\begin{itemize}
	\item Objective function: Eq.~(\ref{eee});
	\item Gradient: Eq.~(\ref{e:gradiant});
	\item Termination condition: Eq.~(\ref{e:termination}).
	\end{itemize}	
    \State Propagate the result $\textbf{m}^{L_{m}}$ to upper layer $L_{m}-1$ as an initial guess for keypoint movement $\textbf{m}^{L_{m}-1}$;
    \State Refine the movement vector $\textbf{m}^{L_{m}-1}$ at Level $L_{m}-1$ by Eq.~(\ref{eee});
    \State Propagate the result $\textbf{m}^{L_{m}-1}$ to the level $L_{m}-2$ and so on up to the $1$-st level to get $\textbf{m}^{1}$;   
    \\
    \Return $\textbf{m}= \textbf{m}^{1}$;    
  \end{algorithmic}  
\end{algorithm}

\noindent \textbf{Pyramid Model:} The movement vector is computed in an image pyramid for robustness. Similar to ORB-SLAM2, we detect the keypoints in an eight-level image pyramid with scale ratio $= 1.2$. Let the pyramid levels be $L = 1, ..., L_{m}$ (where $L_{m}=8$ is the deepest pyramid level), $\textbf{m}^{L}$ be the keypoint movement vector on the $I^{L}$-th image pyramid, hence, $\textbf{m}=\textbf{m}^{1}$ is the solution. The steps are summarized in \textbf{Algorithm \ref{pyramidalgorithm}}.

\begin{figure*}[htb]
\centering 
\subfigure[Image Pairs]{
\begin{minipage}{0.24\textwidth}
\includegraphics[width=1\textwidth]{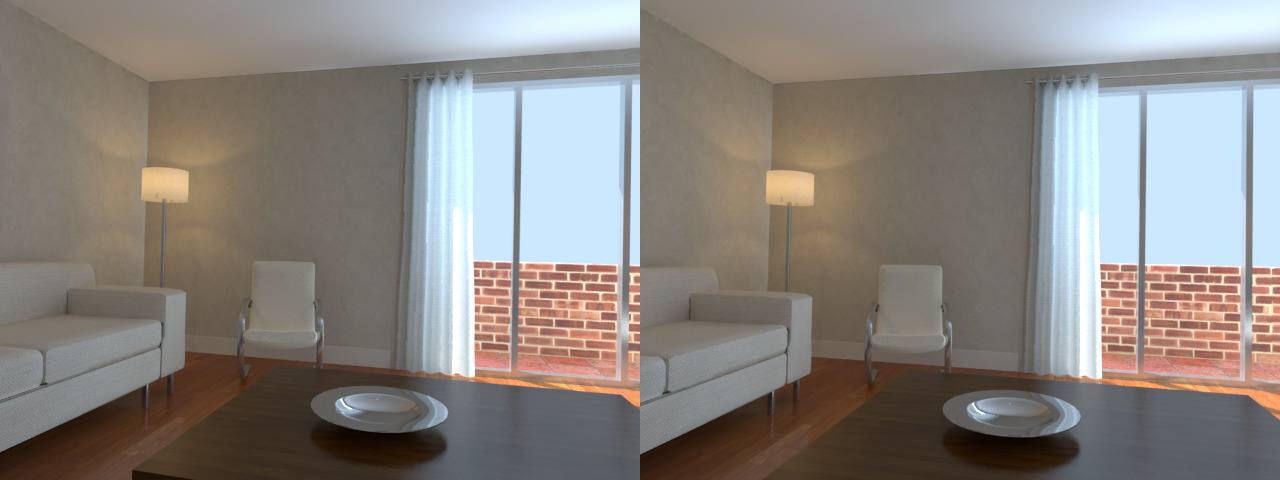} 
\includegraphics[width=1\textwidth]{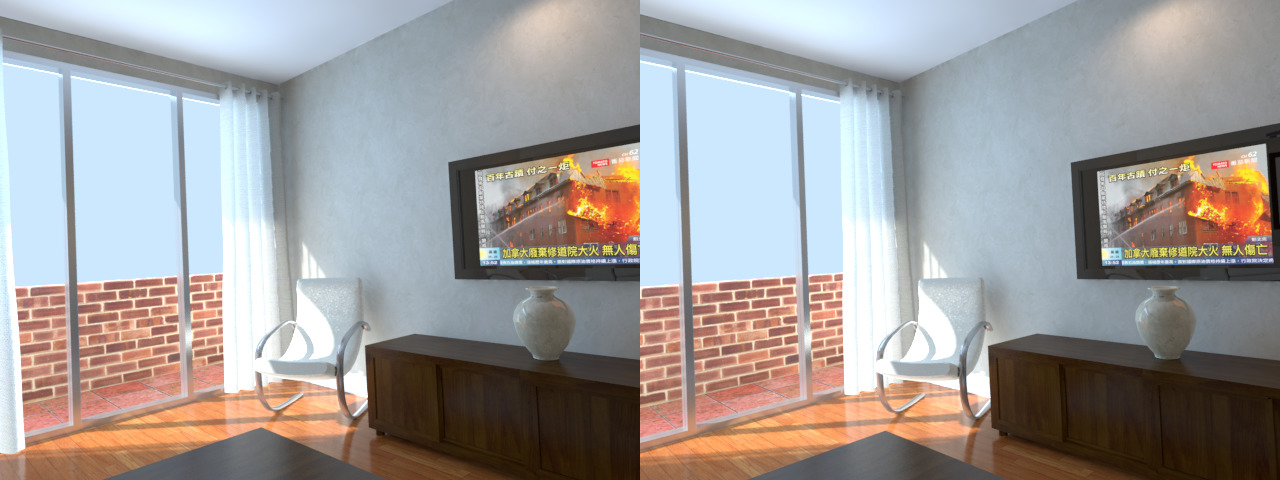}
\includegraphics[width=1\textwidth]{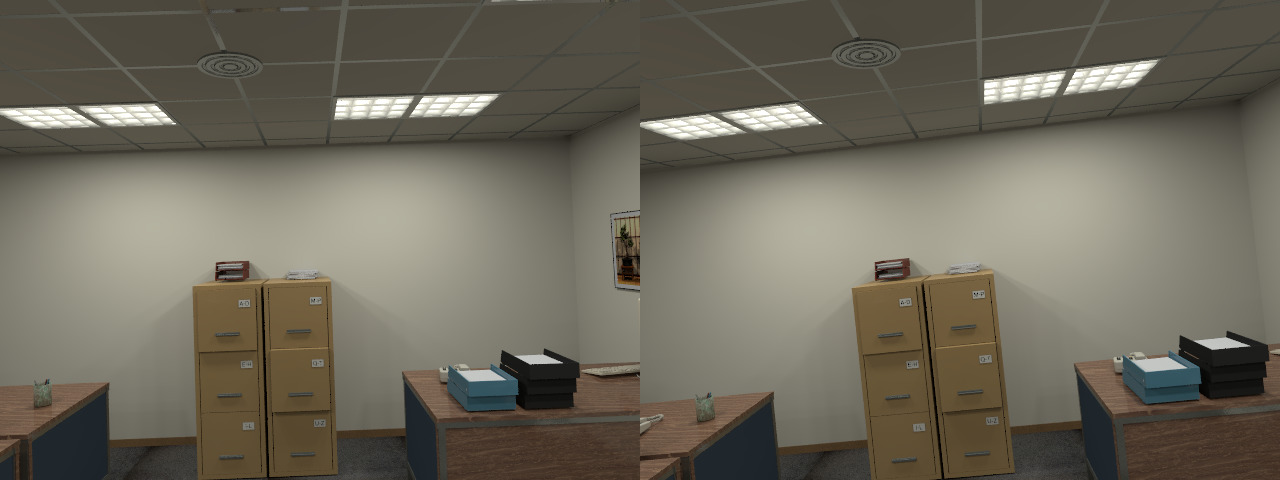}
\includegraphics[width=1\textwidth]{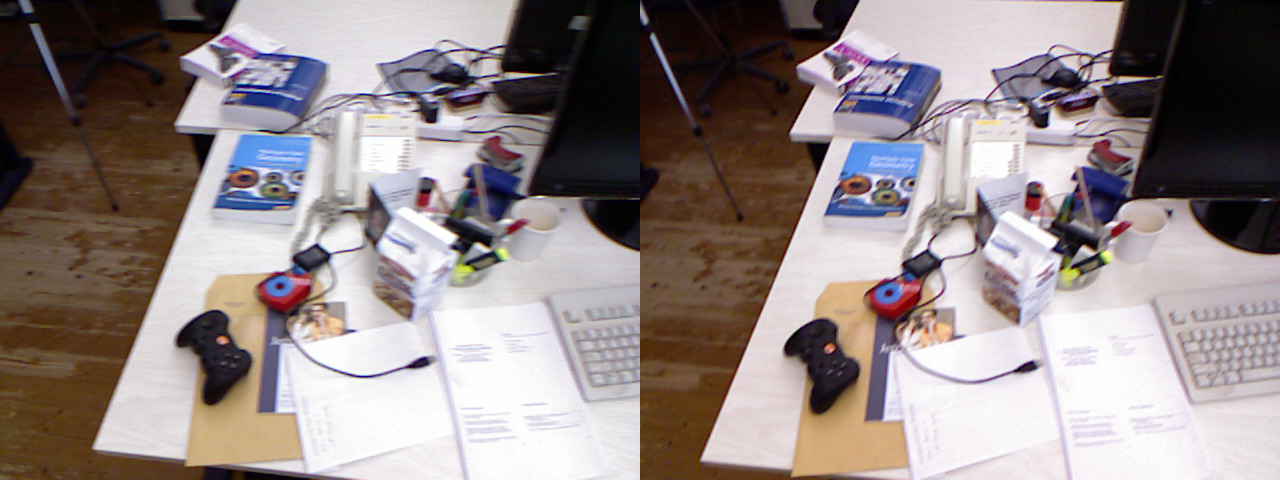}
\includegraphics[width=1\textwidth]{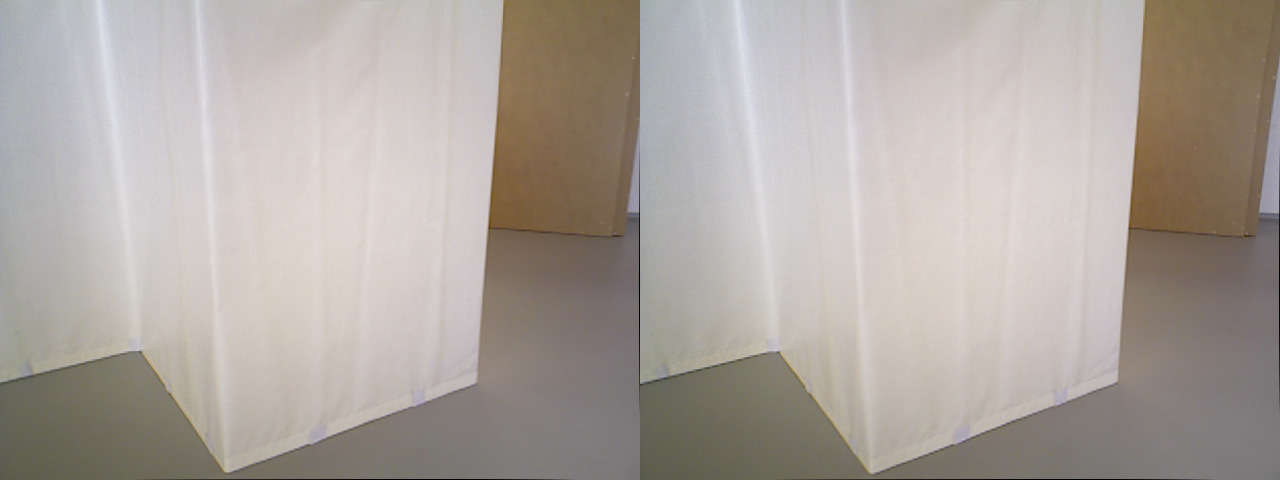}
\includegraphics[width=1\textwidth]{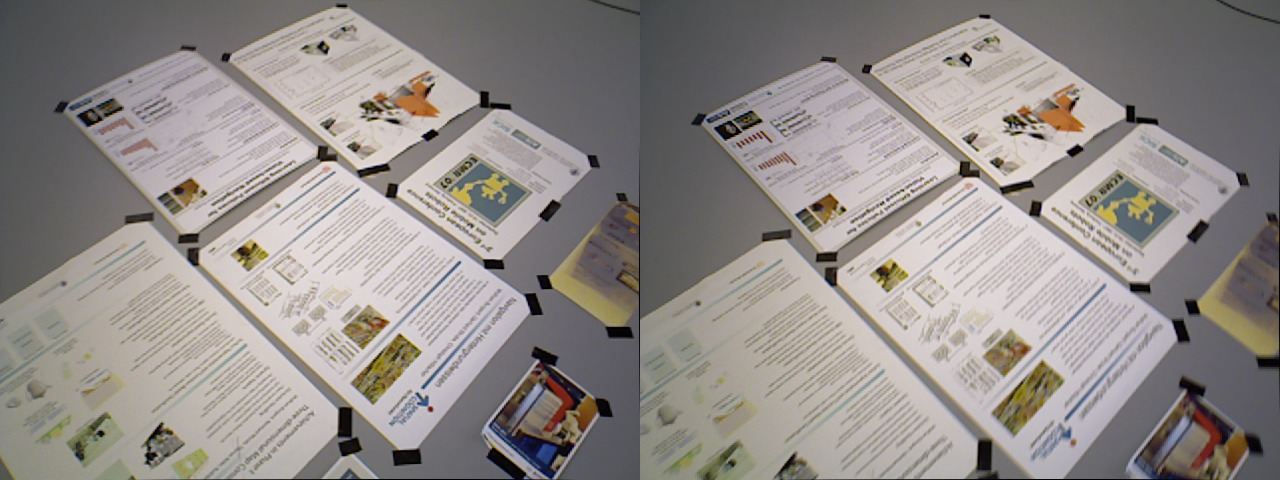}
\includegraphics[width=1\textwidth]{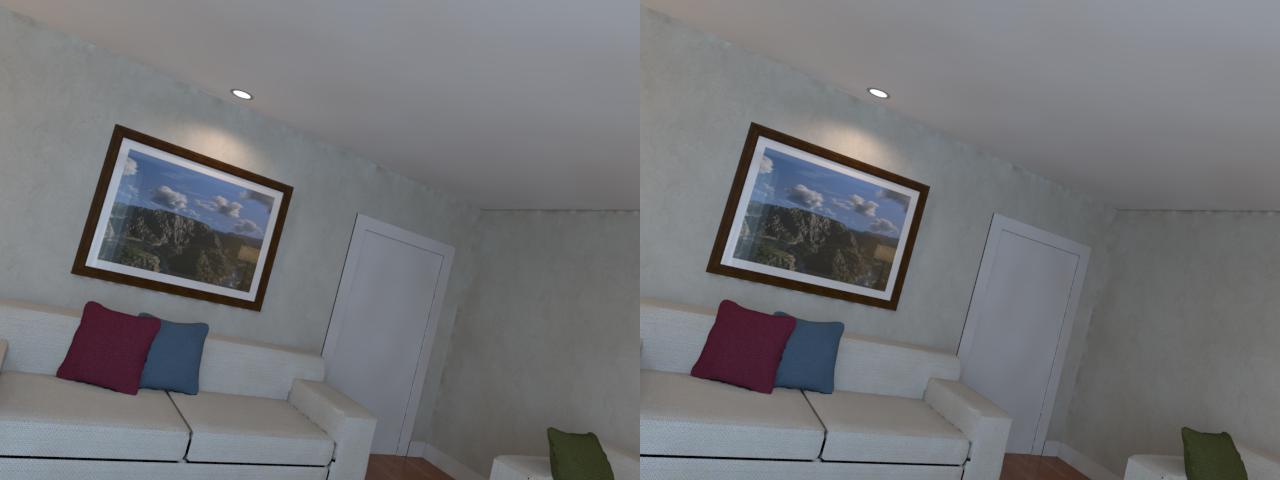}
\vspace{-2mm}  
\end{minipage}}\hspace{-5mm}
\quad
\subfigure[KLT (w/o UAM)]{
\begin{minipage}{0.24\textwidth}
\includegraphics[width=1\textwidth]{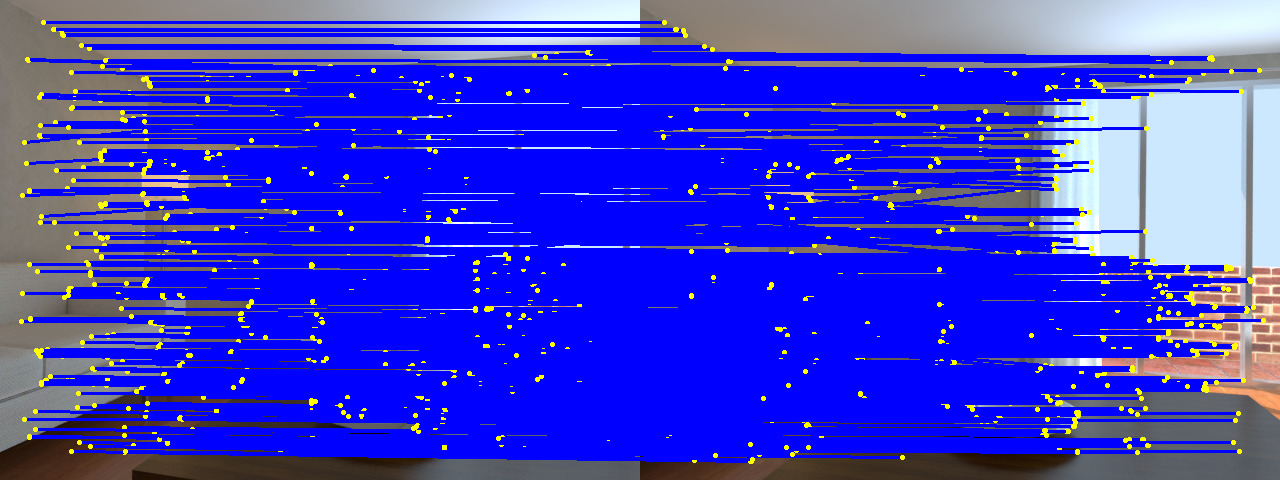}
\includegraphics[width=1\textwidth]{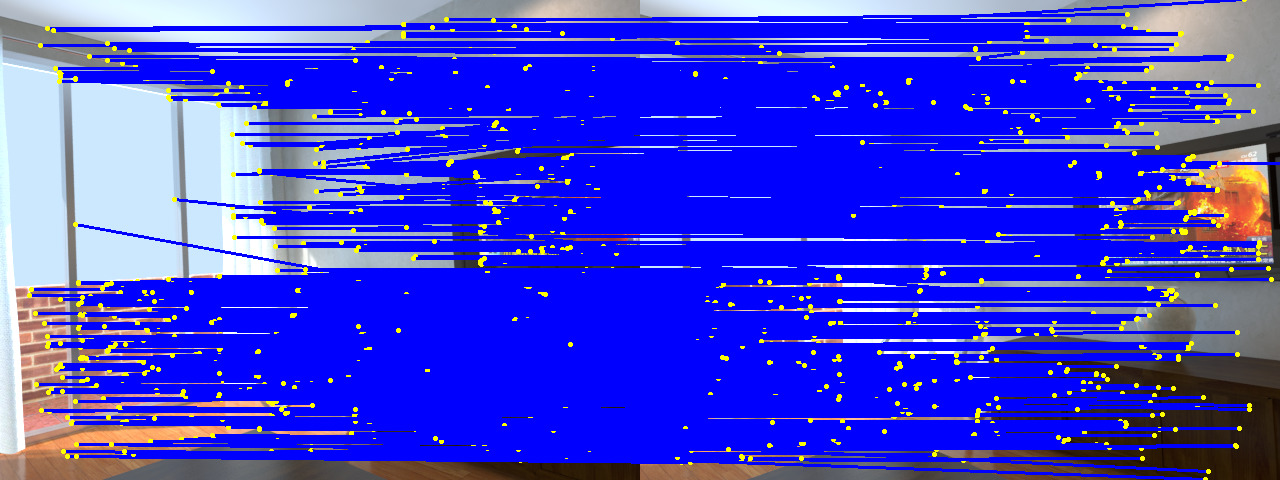}
\includegraphics[width=1\textwidth]{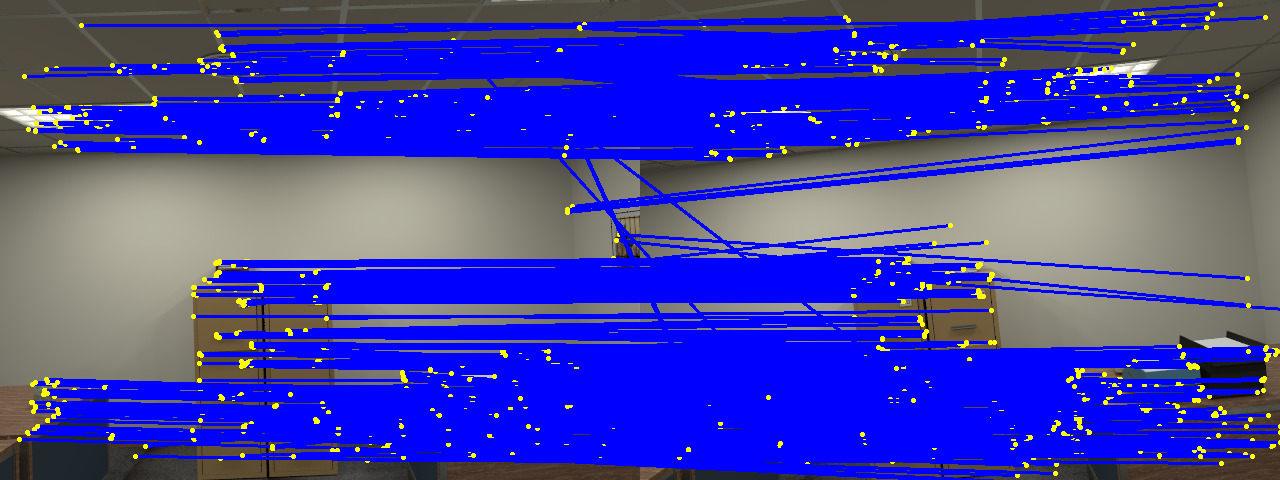}
\includegraphics[width=1\textwidth]{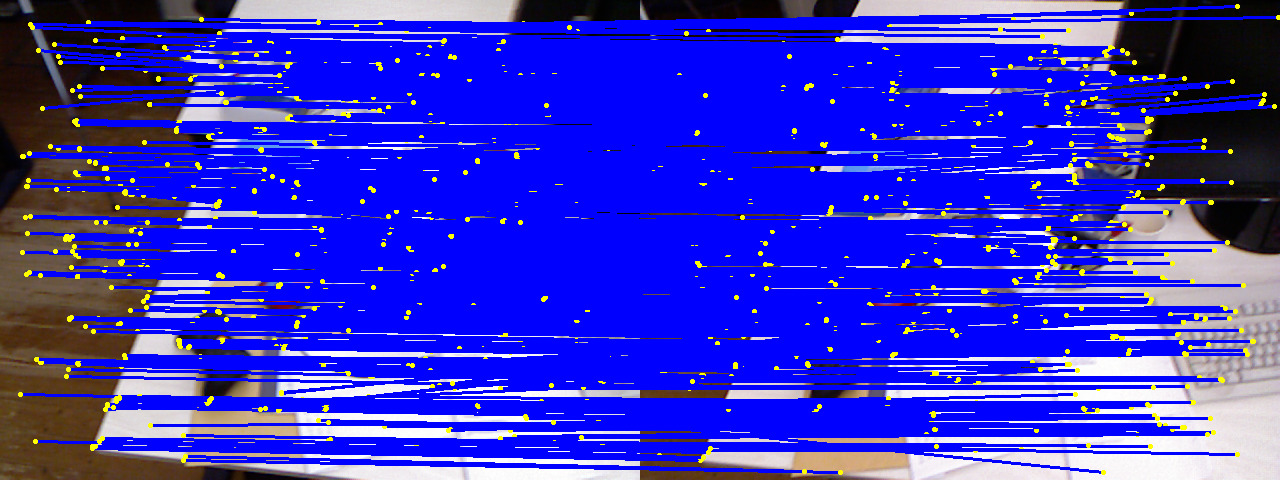}
\includegraphics[width=1\textwidth]{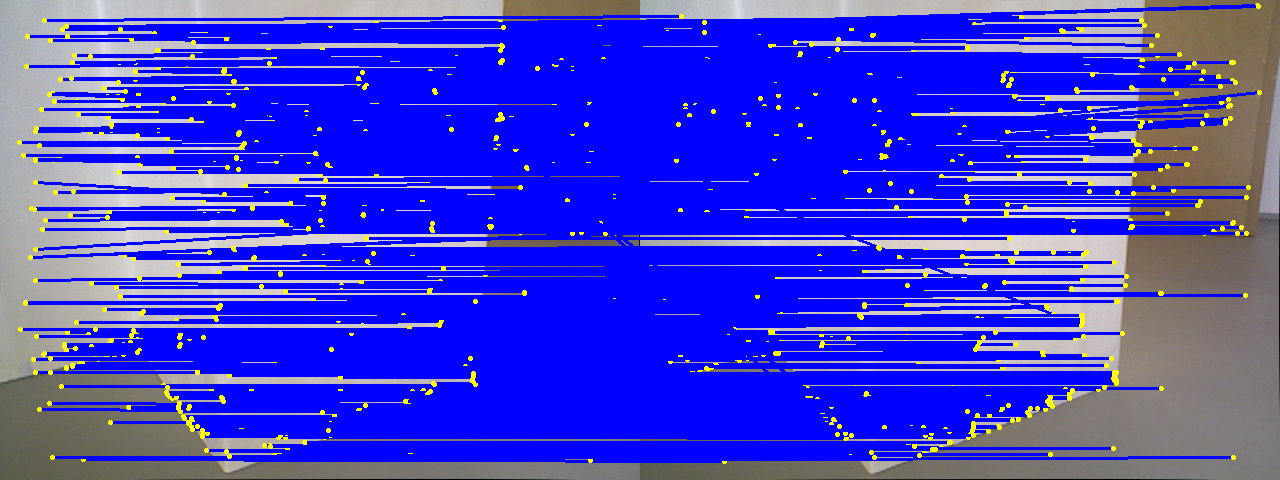}
\includegraphics[width=1\textwidth]{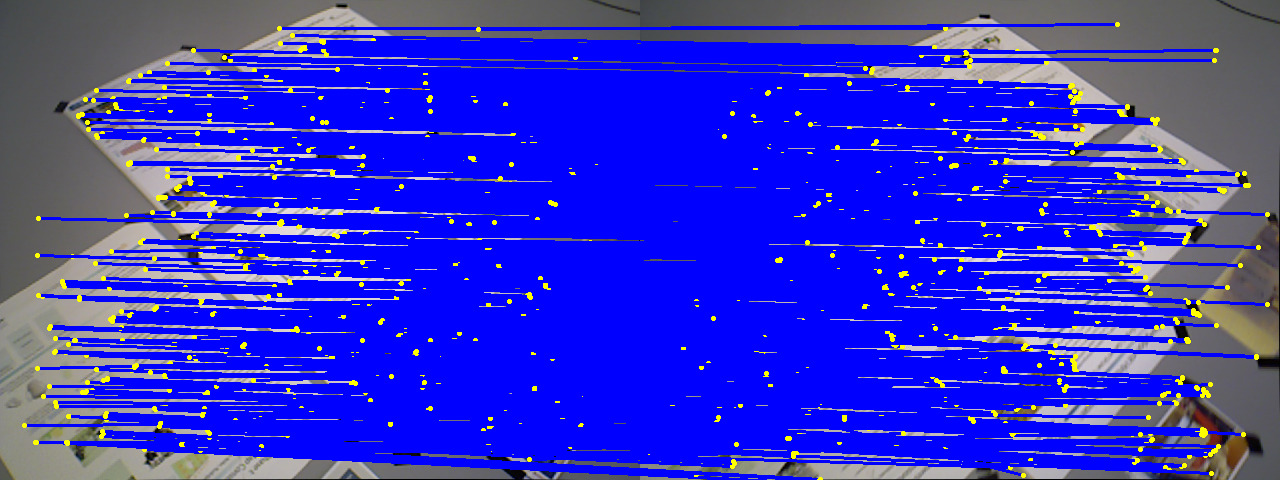}
\includegraphics[width=1\textwidth]{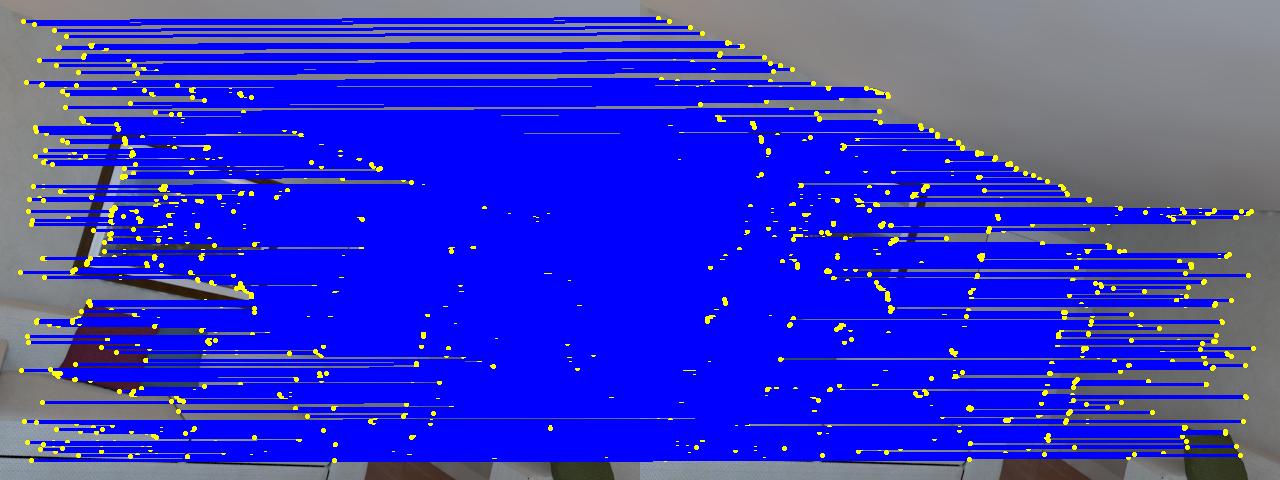} 
\vspace{-2mm}
\end{minipage}}\hspace{-5mm}
\quad
\subfigure[KLT (w/ UAM) (Ours)]{
\begin{minipage}{0.24\textwidth}
\includegraphics[width=1\textwidth]{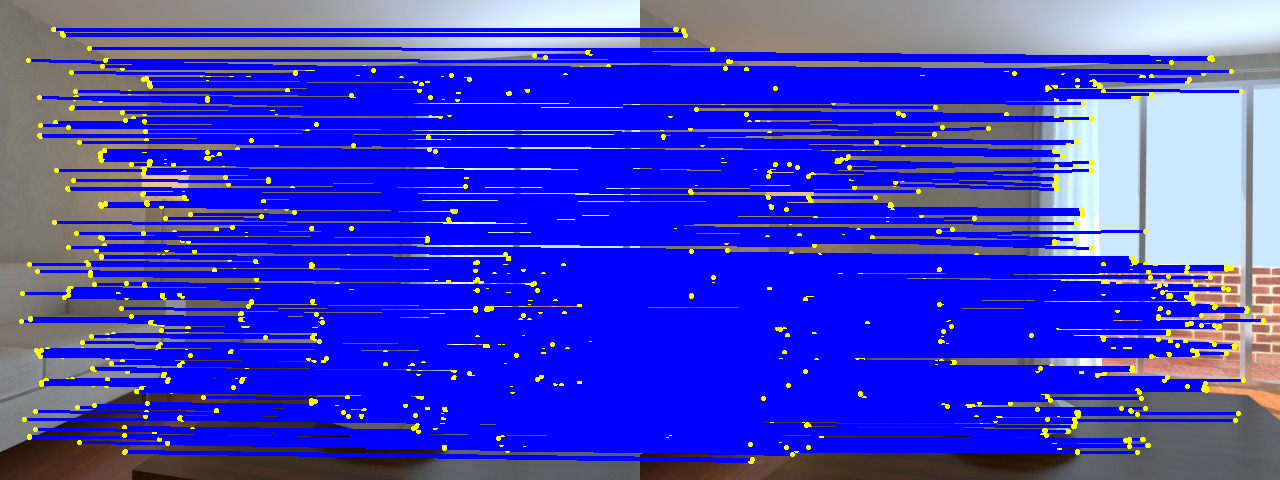}
\includegraphics[width=1\textwidth]{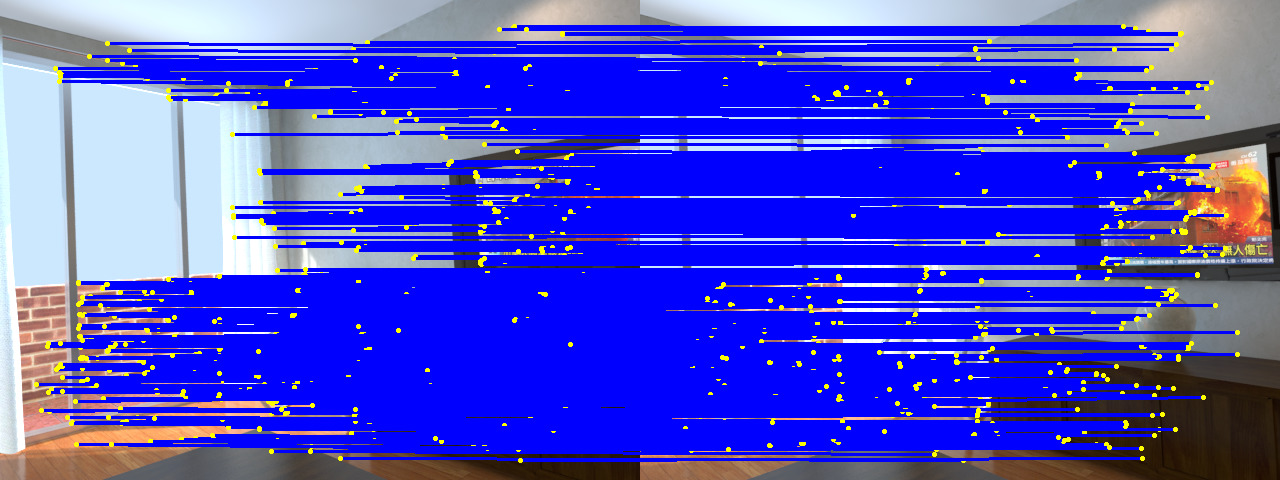}
\includegraphics[width=1\textwidth]{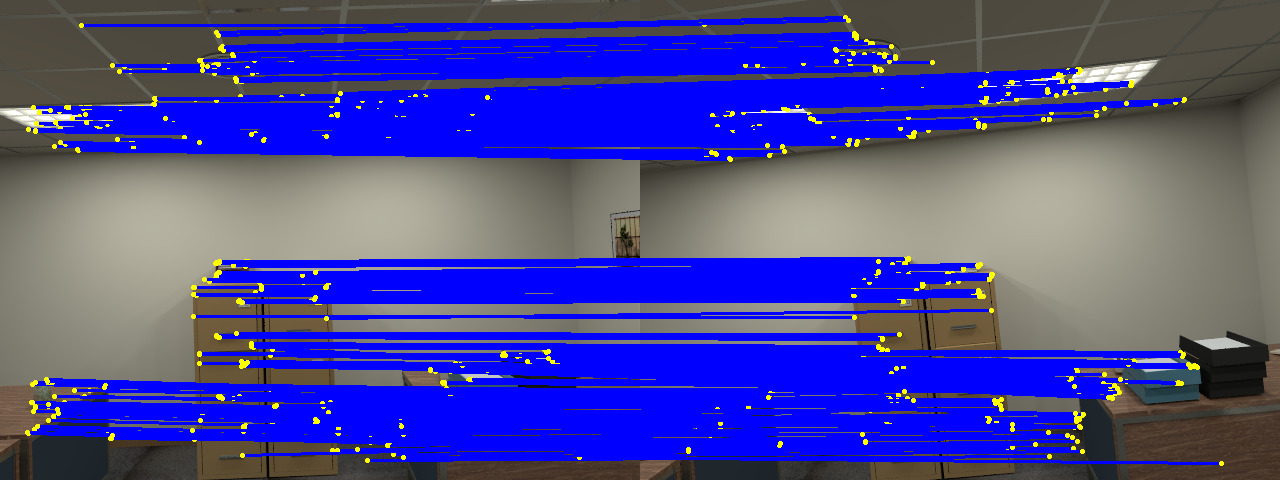}
\includegraphics[width=1\textwidth]{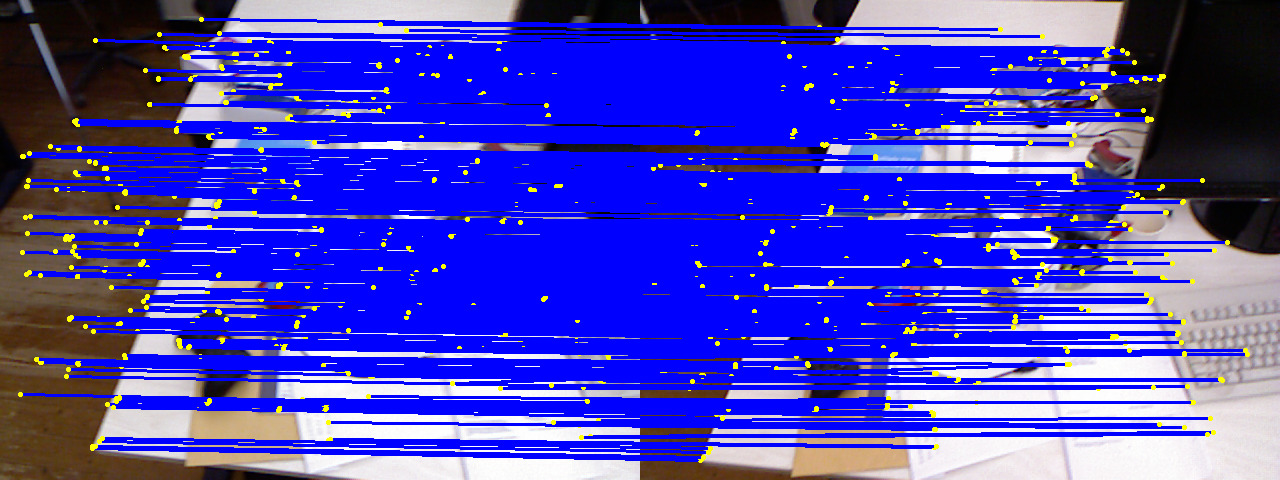}
\includegraphics[width=1\textwidth]{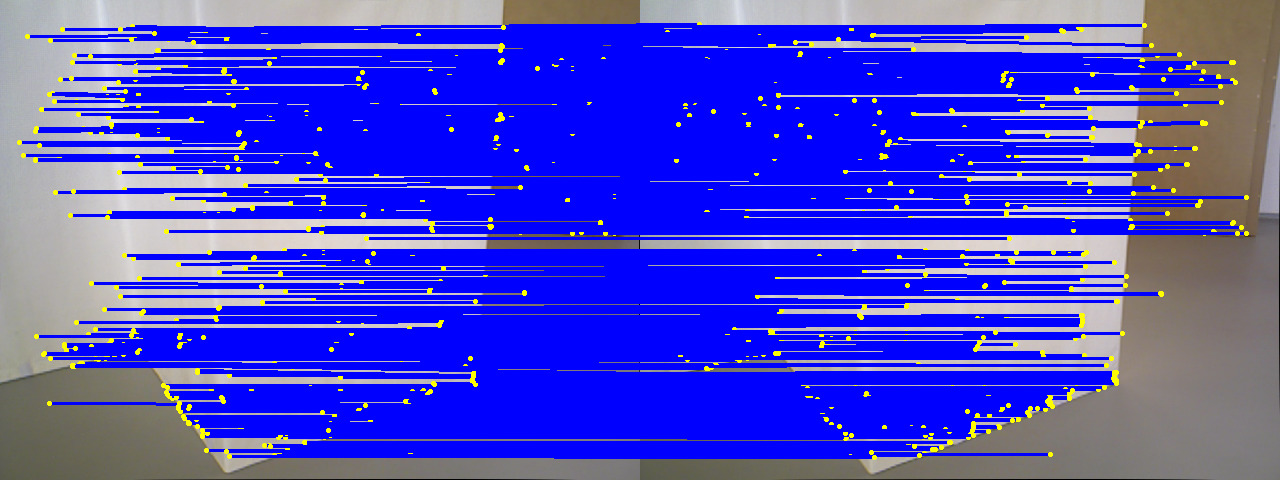}
\includegraphics[width=1\textwidth]{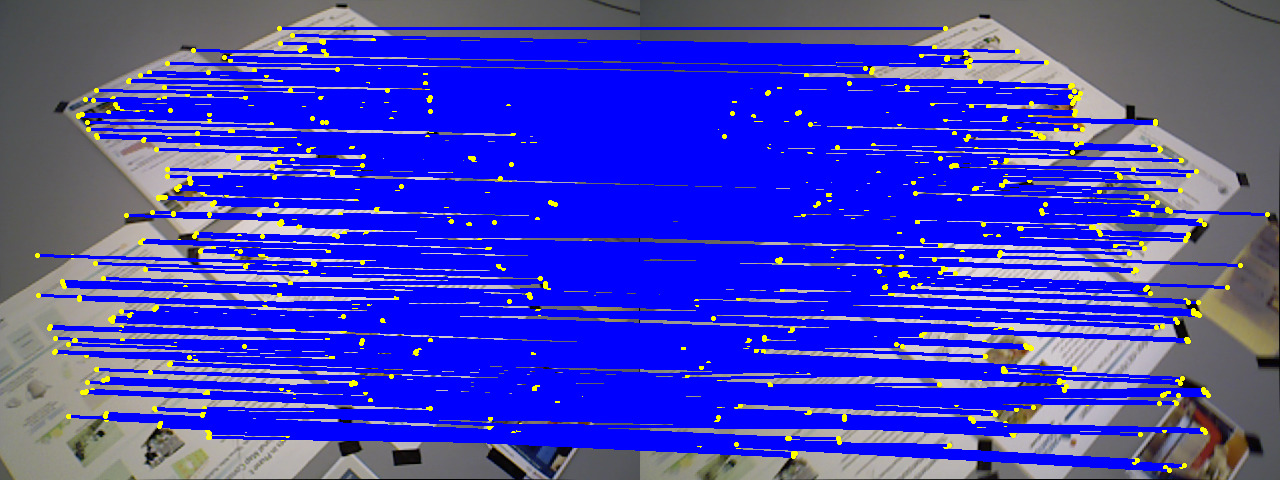}
\includegraphics[width=1\textwidth]{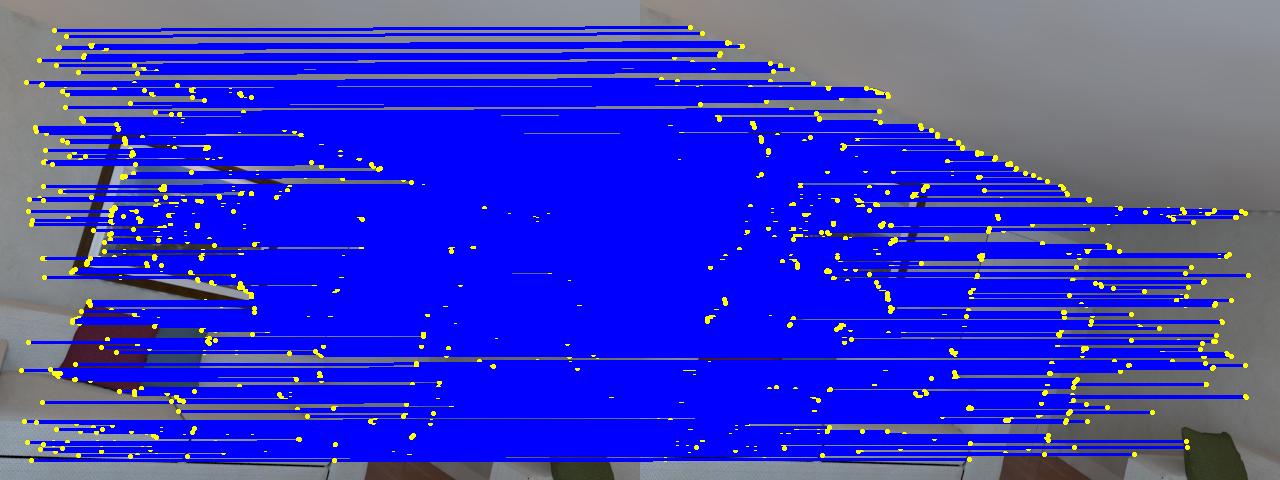}  
\vspace{-2mm}
\end{minipage}}\hspace{-5mm}
\quad
\subfigure[Inlier Refinement (Ours)]{
\begin{minipage}{0.24\textwidth}
\includegraphics[width=1\textwidth]{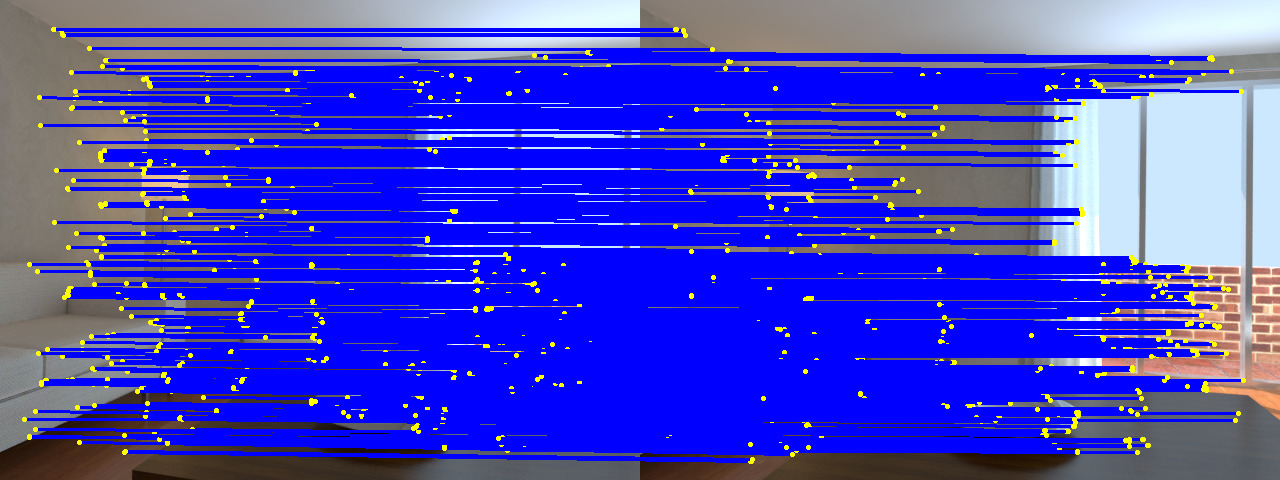}
\includegraphics[width=1\textwidth]{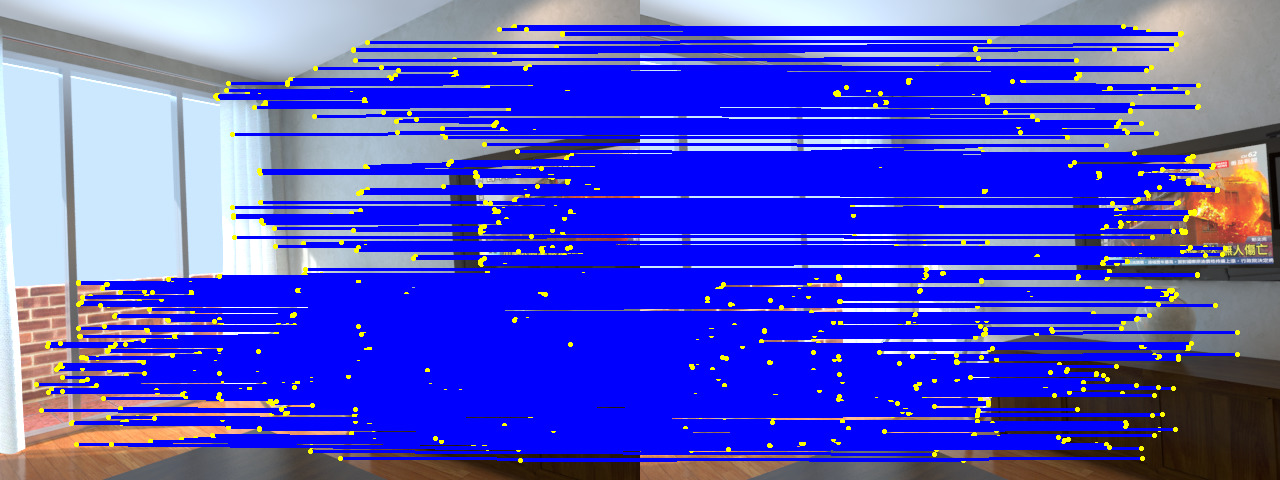}
\includegraphics[width=1\textwidth]{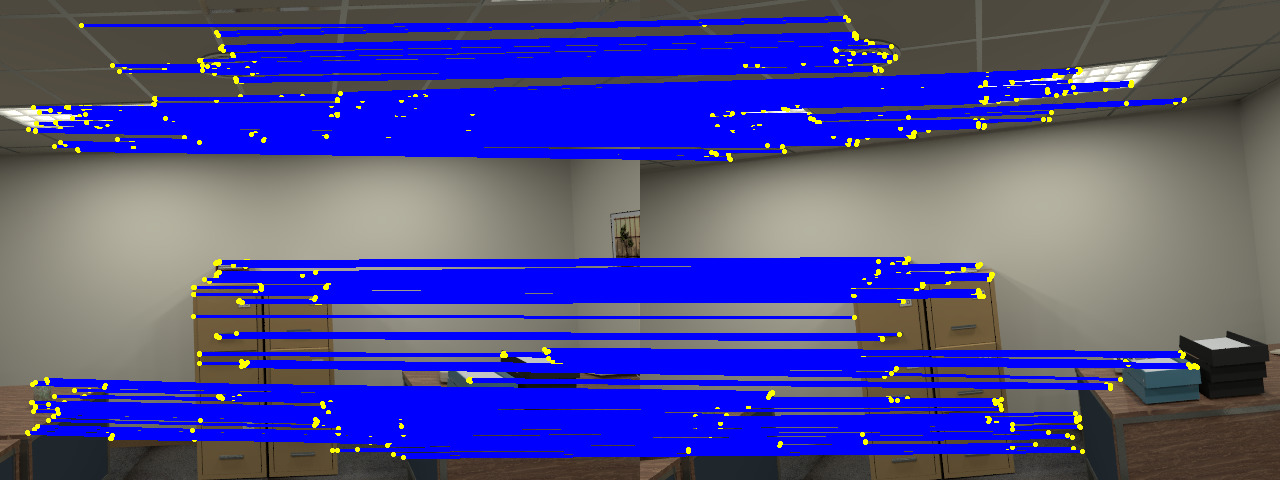}
\includegraphics[width=1\textwidth]{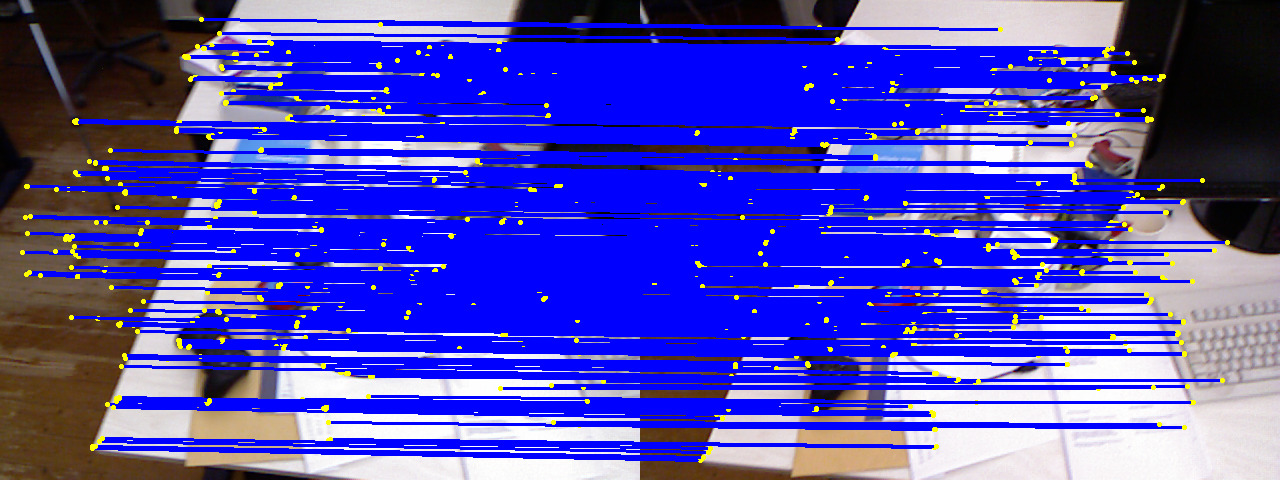}
\includegraphics[width=1\textwidth]{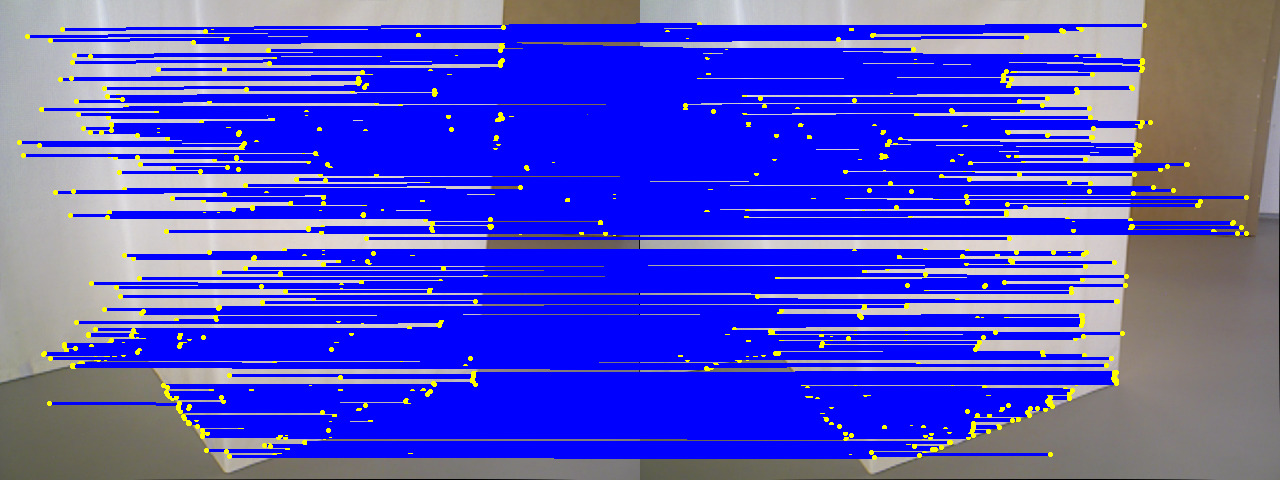}
\includegraphics[width=1\textwidth]{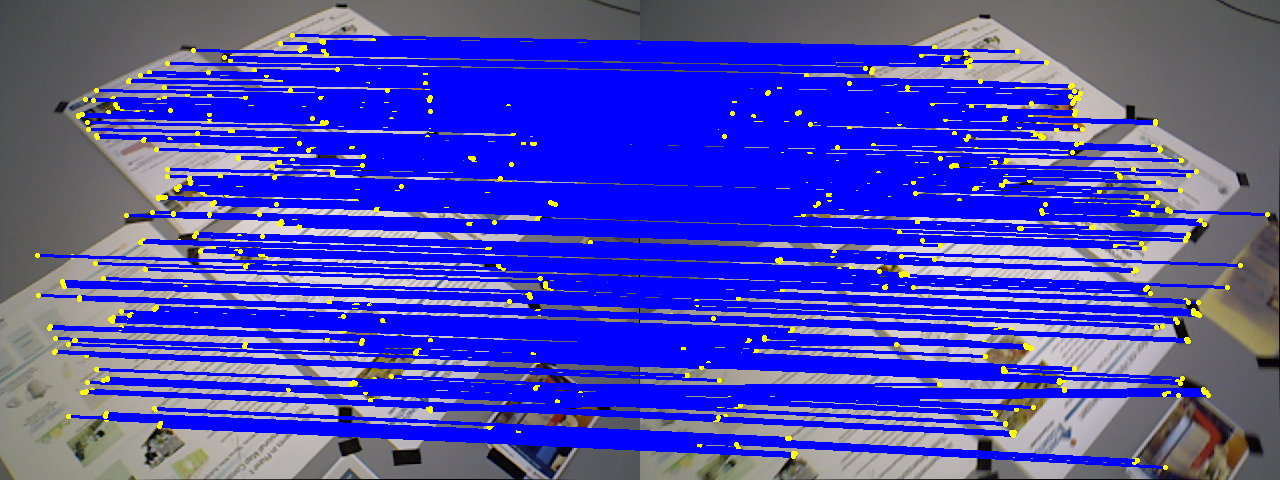}
\includegraphics[width=1\textwidth]{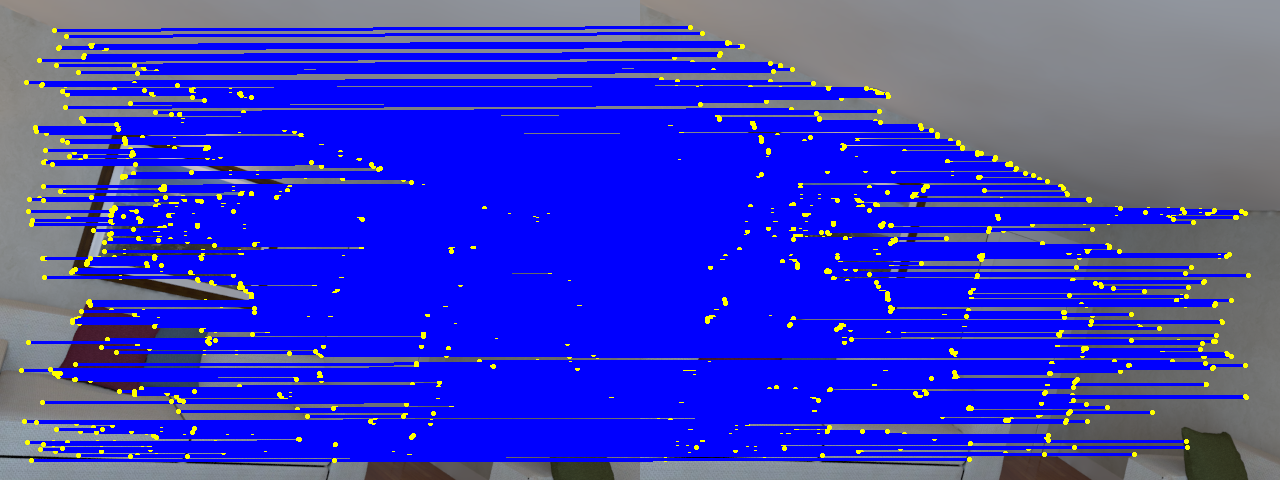} 
\vspace{-2mm}
\end{minipage}}
\vspace{-1mm}
\caption{Keypoint tracking examples of our method. We detect 1000 keypoints for every image. The left and right images in (a) represent the reference and current frame, respectively. (b) and (c) represent the results produced by KTL (w/o UAM) and ours (w/ UAM), respectively. (d) represents our results after inlier refinement including motion smoothness and epipolar constraint.}
\vspace{-4mm}
\label{matches}
\end{figure*}


\noindent \textbf{Solving for the Movement Vector}: We solve the two variances $dx$ and $dy$ in a fixed-size integration window. Specifically, we set the size as $\omega_{x}=\omega_{y} = 2$. For such a window, the problem boils down to solving $dx$ and $dy$ with a system of $25$ equations (points). Thus, the objective function is modeled as:
\vspace{-2mm}
\begin{equation}
\begin{aligned}
   \mathop{\arg\min}_{dx,dy} \sum_i^k \left|\left|I_{r}(x_{i},y_{i})-I_{c}(x_{i}^{\star}+dx,y_{i}^{\star}+dy)\right|\right|^{2},
\end{aligned}
\label{eee}
\end{equation}
where $k$ denotes the pixels in integration window (we use $k= 25$), $i \in {1,2,...,k}$. This equation can be solved with an iterative Lucas-Kanade (KLT) method \cite{KLT}.
\par

An essential observation in KLT is that image derivatives $I_{x}$ and $I_{y}$ can be computed directly from the reference image $I_{r}$ in the neighborhood of the keypoint independently from the second image $I_{c}$. Hence, the gradient expression can be defined as:
\begin{equation}
\frac{\partial \epsilon(\textbf{m})}{\partial \textbf{m}}=\sum_{i=1}^k\begin{bmatrix}I_{x}(x,y)I_{x}(x,y) & I_{x}(x,y) I_{y}(x,y)\\I_{x}(x,y) I_{y}(x,y) & I_{y}(x,y)I_{y}(x,y) \end{bmatrix}
\label{e:gradiant}
\end{equation}
where $I_{x}(x,y)$ and $I_{y}(x,y)$ denote the image derivatives of the position $(x_{i},y_{i})$ in $x$ and $y$ axes, respectively. 
Finally, we define a accuracy evaluation function $\epsilon(w)$ to determine termination criterion for the iterations. We have
\begin{equation}
\epsilon(w)=\frac{\sum_{i=1}^k{\left|\left| I_{r}(x_{i},y_{i})-I_{c}(x_{i}+dx,y_{i}+dy)\right|\right|}}{k}, 
\end{equation}
which denotes the average grayscale residual between the windows (patches). Let $N_{iter}$ be the iteration number. The termination condition is defined as
 \begin{equation}
\epsilon(w) < w_{errormin} \|  N_{iter} > N_{itermax},
\label{e:termination}
\end{equation}
where $w_{errormin}$ denotes the minimum value of the window error and $N_{itermax}$ denotes the maximum number of iterations allowed. We set $w_{errormin} = 0.02$  and $ N_{itermax} = 10$.
\par

\begin{table*}[htb]\caption{Keypoint Tracking Comparison of Ratio [\%] and Time [ms]. Row means Row Number in Fig \ref{matches}}
\vspace{-2mm}
\centering
\setlength{\tabcolsep}{2.0mm}\begin{tabular}{@{}l|cccccccccccccccc@{}}
\toprule
\multicolumn{1}{c|}{} & \multicolumn{2}{c|}{Row 1}      & \multicolumn{2}{c|}{Row 2}        & \multicolumn{2}{c|}{Row 3}        & \multicolumn{2}{c|}{Row 4}        & \multicolumn{2}{c|}{Row 5}        & \multicolumn{2}{c|}{Row 6} & \multicolumn{2}{c|}{Row 7} & \multicolumn{2}{c}{Average} \\ \cmidrule(l){2-17} 
\multicolumn{1}{l|}{} & Ratio & \multicolumn{1}{c|}{Time} & Ratio & \multicolumn{1}{c|}{Time} & Ratio & \multicolumn{1}{c|}{Time} & Ratio & \multicolumn{1}{c|}{Time} & Ratio & \multicolumn{1}{c|}{Time} & Ratio     & \multicolumn{1}{c|}{Time} & Ratio   & \multicolumn{1}{c|}{Time} & Ratio  & Time       \\ \midrule
KLT (w/o UAM)                  & 0.86  & 0.44                      & 0.80  & 0.71                      & 0.62  & 4.74                      & 0.77  & 1.26                      & 0.82  & 0.66                      & 0.70         & 2.53    & 0.99       & 0.21   & \textbf{0.79}       & \textbf{1.50}\\
Ours (w/ UAM)                 & 0.90  & 0.26                      & 0.88  & 0.25                      & 0.83  & 0.27                      & 0.84  & 0.32                      & 0.84  & 0.34 & 0.83         & 0.40       & 0.99       & 0.21   & \textbf{0.87}     & \textbf{0.29} \\ \bottomrule
\end{tabular}
\begin{flushleft}
``Ratio'' represents inlier ratio where inliers are the number of keypoints on the current frame after verification of the epipolar constraint. ``Time'' represents the time taken for the verification. Both algorithms spent 6-7 ms on keypoint matching which is not counted in this Table. Row $n$ represents the $n$-th row in Fig \ref{matches}. Our method is faster and achieves higher accuracy.
\end{flushleft}
\label{tmatch}
\vspace{-4mm}
\end{table*}

\vspace{-2mm}
\subsection{Inlier Refinement}
The previous steps establish robust keypoint correspondences between the reference and current frame, however, incorrect matches are still possible. In this stage, we adopt two efficient strategies to refine inliers: \par
\begin{enumerate}
	\item motion smoothness constraint proposed in \cite{gms1, gms2} to filter out outliers;
	\item epipolar constraint to further refine the inliers.
\end{enumerate}
To process 1000 keypoints, the motion smoothness constrain takes $\backsim$0.15ms, and the epipolar constraint, implemented via RANSAC-based fundamental matrix, takes $\backsim$0.25ms.

\begin{table*}[htb]\caption{Camera Localization RMSE [m] Error and Average Time [s] Comparison on the \textit{ICL-NUIM} dataset}
\vspace{-2mm}
\centering
\setlength{\tabcolsep}{1.6mm}{
\begin{tabular}{@{}l|cccccccccccccccc@{}}
\toprule
\multicolumn{1}{c|}{\multirow{2}{*}{}} & \multicolumn{2}{c|}{Office 0}              & \multicolumn{2}{c|}{Office 1}              & \multicolumn{2}{c|}{Office 2}              & \multicolumn{2}{c|}{Office 3}              & \multicolumn{2}{c|}{Living 0}              & \multicolumn{2}{c|}{Living 1}              & \multicolumn{2}{c|}{Living 2}              & \multicolumn{2}{c}{Living 3}    \\ \cmidrule(l){2-17} 
\multicolumn{1}{c|}{}                  & RMSE           & \multicolumn{1}{c|}{Time} & RMSE           & \multicolumn{1}{c|}{Time} & RMSE           & \multicolumn{1}{c|}{Time} & RMSE           & \multicolumn{1}{c|}{Time} & RMSE           & \multicolumn{1}{c|}{Time} & RMSE           & \multicolumn{1}{c|}{Time} & RMSE           & \multicolumn{1}{c|}{Time} & RMSE           & Time           \\ \midrule
ORB-SLAM2                              & 0.038          & 0.021                     & \textbf{0.070} & 0.024                     & \textbf{0.011} & 0.020                     & 0.066          & 0.021                     & \textbf{0.006} & 0.021                     & 0.101          & 0.024                     & \textbf{0.015} & 0.021                     & 0.013          & 0.022          \\
Ours                                   & \textbf{0.034} & \textbf{0.013}             & 0.080          & \textbf{0.013}            & 0.015          & \textbf{0.014}            & \textbf{0.037} & \textbf{0.012}            & 0.010          & \textbf{0.014}            & \textbf{0.026} & \textbf{0.015}            & 0.016          & \textbf{0.013}            & \textbf{0.009} & \textbf{0.012} \\ \bottomrule
\end{tabular}}
\begin{flushleft}
All statistics are collected via real reproduction test i.e. median over 5 executions for each sequence. RMSE represents translation in meters. Time represents average time per frame in seconds. On the \textbf{Office 3} and \textbf{Living 1} sequences, our method achieves much higher localization accuracy than ORB-SLAM2 with less computation time. Our method is also highly competitive on the other sequences.
\end{flushleft} 
\label{ticl} 
\vspace{-5mm}
\end{table*}

\section{Experiments}\label{experiment}

We evaluate the performance of FastORB-SLAM using an RGB-D camera for localization accuracy and efficiency. We first test the proposed keypoint matching in Section \ref{kpcp}) to demonstrate that our method can establish reliable keypoint correspondences and build a foundation for high-accuracy pose estimation. In Section \ref{locationexperiment}, we compare the proposed system to almost all (nine) open-source RGB-D SLAM systems to demonstrate the performance of FastORB-SLAM. 



We used two well-known public RGB-D datasets, \textit{TUM} \cite{TUM} and \textit{ICL-NUIM} \cite{ICL}. All experiments were performed on a laptop computer with Intel Core i7-10710U CPU @1.10 GHz without GPU parallelization. FastORB-SLAM was implemented in C++ on Ubuntu 18.04 LTS and based on OpenCV 3.4, Sophus, Eigen, and the g2o \cite{g2o} libraries.


\vspace{-3mm}

\subsection{Keypoint Matching}
\label{kpcp}

\begin{figure*}[htbp]
\centering  
	\includegraphics[width=0.38\textwidth]{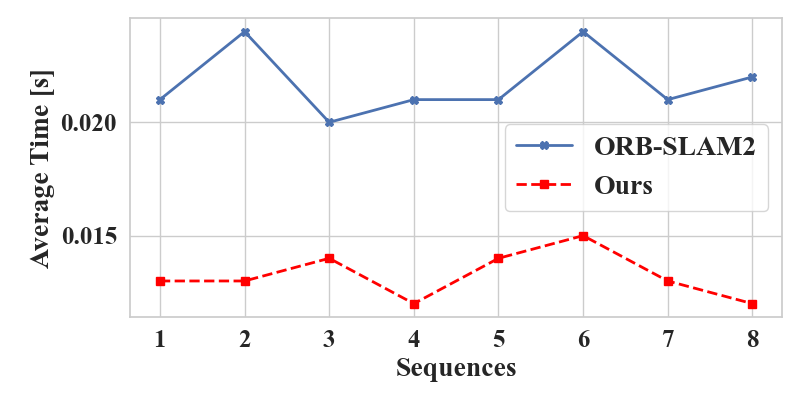} \hspace{8mm}
	\includegraphics[width=0.38\textwidth]{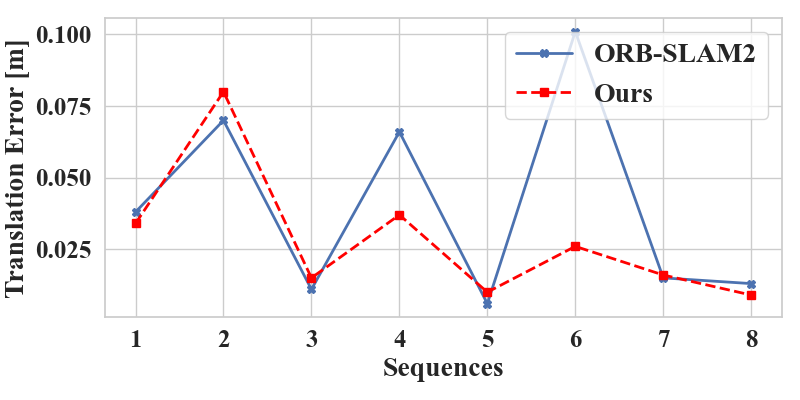}
	\vspace{-4mm}
\caption{Average Time and Translation Error comparison in all 8 sequences (Office 0-3 and Living 0-3)  of \textit{ICL-NUIM} dateset. More details are in Table \ref{ticl}. Our method produces highly competitive localization accuracy with significantly lower processing time. In Sequence 4 (Office 3) and Sequence 6 (Living 1), ORB-SLAM2 has a large drift error whereas our method maintains lower errors. 
Fig \ref{iclfigure} provides a more detailed comparison of these two sequences.
}
\label{icltime}
\vspace{-5mm}
\end{figure*}

%


\begin{figure*}[htb]
\centering  
	\subfigure[Comparison in \textit{ICL-NUIM-Office 3}]{
	\begin{minipage}{0.48\textwidth}
	\includegraphics[width=0.95\textwidth]{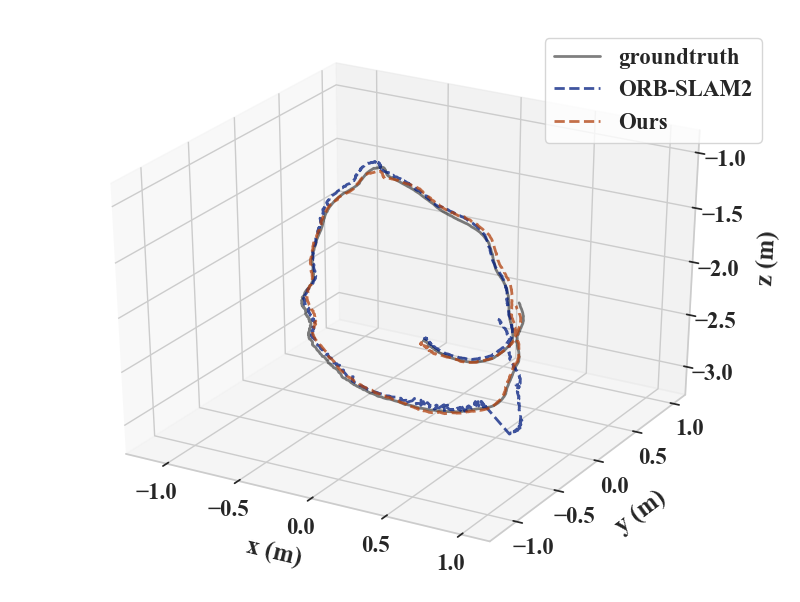}
	\includegraphics[width=0.48\textwidth]{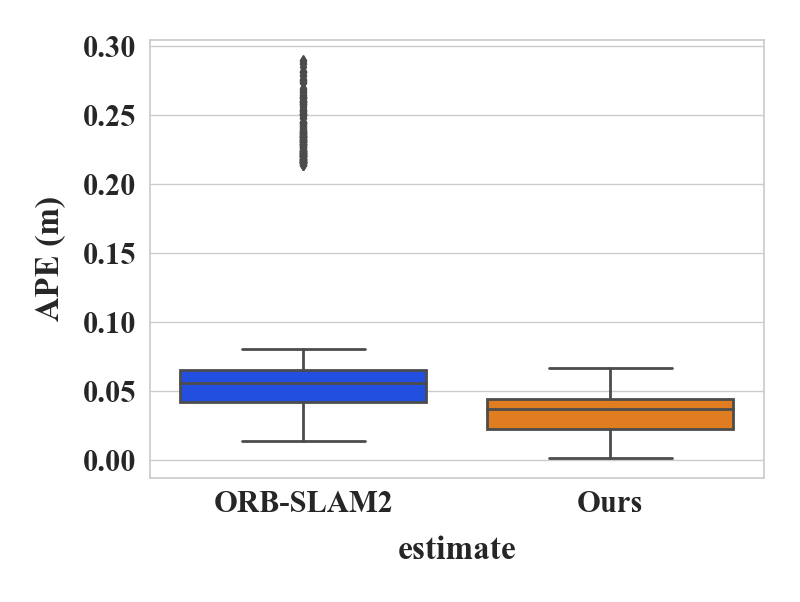}
	\includegraphics[width=0.48\textwidth]{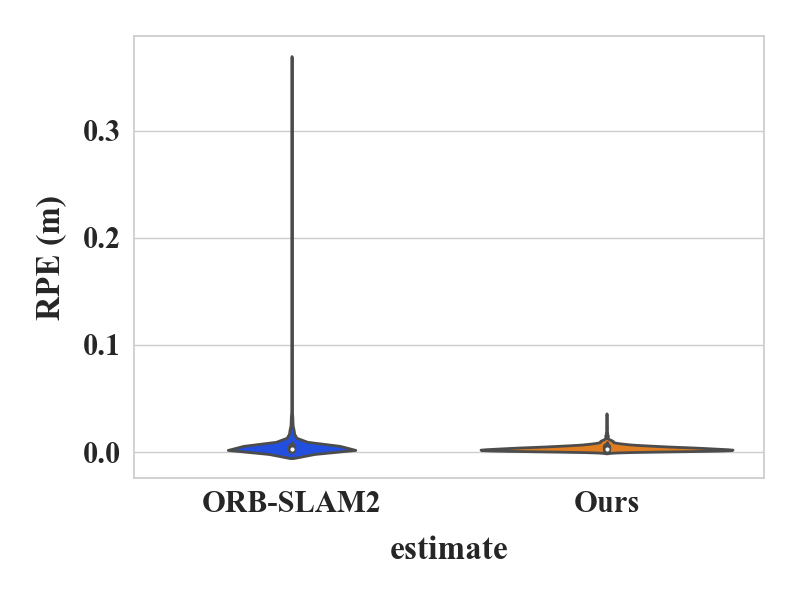}	
	\end{minipage}}\hspace{-5mm}
	\quad
	\subfigure[Comparison in \textit{ICL-NUIM-Living 1}]{
	\begin{minipage}{0.48\textwidth}
	\includegraphics[width=0.95\textwidth]{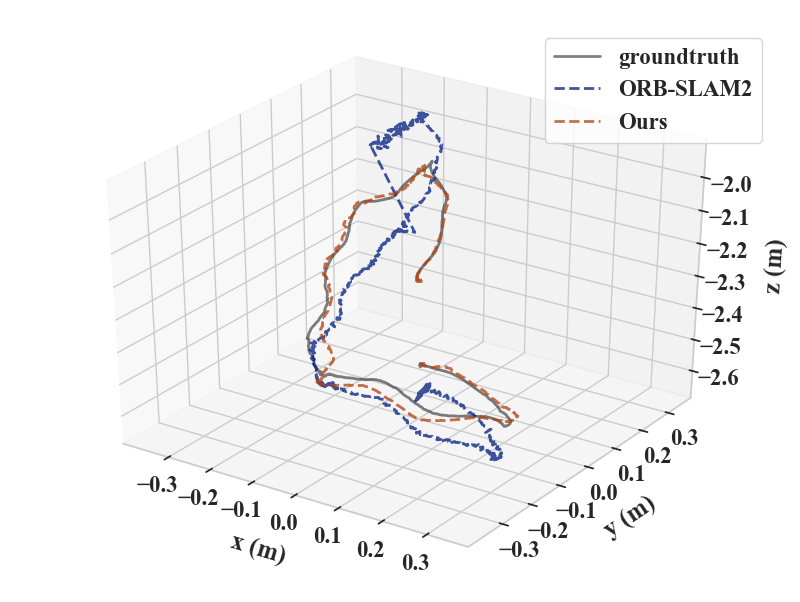}
	\includegraphics[width=0.48\textwidth]{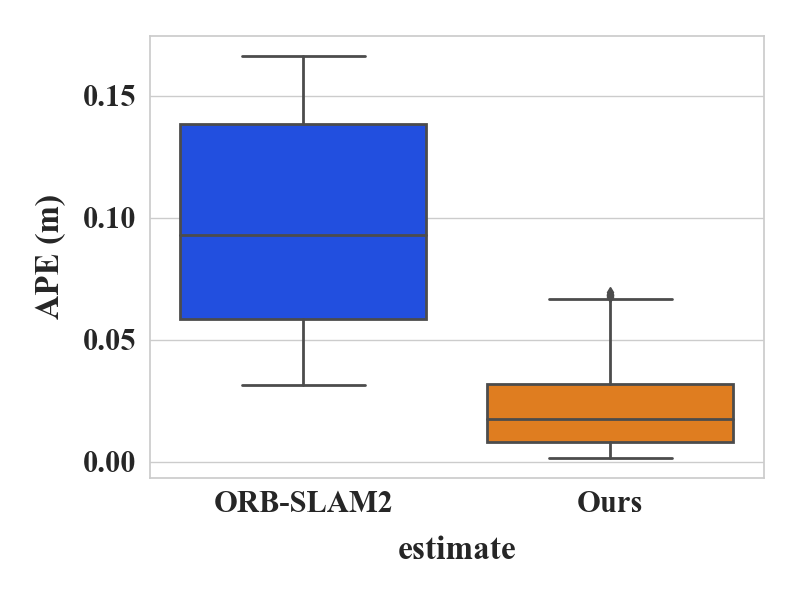}
	\includegraphics[width=0.48\textwidth]{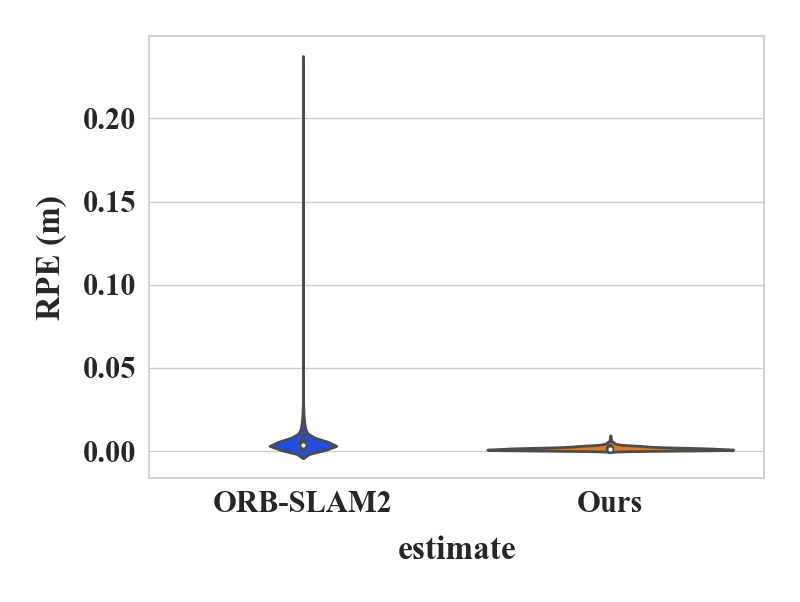}
	\end{minipage}}
	\vspace{-2mm}
\caption{Localization accuracy comparison of our method with ORB-SLAM2 on (a) \textit{ICL-NUIM-Office 3} and (b) \textit{ICL-NUIM-Living 1}. Top row contains 3D motion trajectories and second row compares the APE and RPE errors. 
Our method particularly stands out in the sequence (b) with low-texture episode.
}\label{iclfigure}
\vspace{-5mm}
\end{figure*}

We evaluate the performance of the proposed descriptor-independent keypoint matching method in terms of inlier ratio and computation time. The method includes two stages: robust keypoint matching and inlier refinement. Since the former stage is based on KLT \cite{KLT}, we adopt that as a benchmark for comparison. The major difference is that we employ the UAM model to predict keypoint correspondences as an initial guess (see Section \ref{sparsematching}). Therefore, we test its effect in the following experiments. Considering the keypoints (ORB) are extracted in an image pyramid with $L_{m}=8$ with scale ratio $1.2$, KLT and our method are implemented using the same pyramid. The condition for terminating the iterations is set to $w_{errormin} = 0.02$ or $ N_{itermax} = 10$.  \par

Figure \ref{matches} shows qualitative results where the images are from ICL-NUIM (row 1-3 and 7) and TUM (row 4-6) datasets. In Fig. \ref{matches}(a), the left and right represent reference and current frame, respectively. Fig. \ref{matches}(b) and Fig. \ref{matches}(c) represent results produced by KTL and our method, respectively. Fig. \ref{matches}(d) represent results after inlier refinement (ours). The baseline distance is relatively longer in row 1-6 and smaller in row 7. The threshold for number of keypoints is set to 1000. Notice that our method visually presents higher accuracy than KLT. 

Quantitative results are presented in Table \ref{tmatch} where the ``Ratio'' represents the inlier ratio, following \cite{gms1, gms2} computed as $N_{inlier}/N_{Total}$, where $N_{Total} = 1000$ and $N_{inlier}$ is the number of keypoints after the epipolar constraint verification. ``Time'' is the verification time. Since the both algorithms consume the same time (6-7 ms) for 1000 keypoint extraction, we exclude it from the Table. Row $n$ represents the $n$-th row of Fig. \ref{matches}. From this table, we can conclude that:
\begin{itemize} 
	\item Lower inlier ratio results in higher computation time when using the epipolar constraint to filter outliers e.g., KLT (inlier ratio = 0.80) takes 0.71 ms whereas our method (inlier ratio = 0.88) only takes 0.25 ms.
	\item Good initial guess increases the accuracy of keypoint matching. For instance, our method with UAM produces higher average inlier ratio = 0.87 whereas KLT = 0.79.
	\item Overall, our method achieves higher accuracy than KTL with lower computation time.
\end{itemize}\par

Overall, these experiments show that our method can establish reliable keypoint correspondences without descriptors, building a strong foundation for highly accuracy pose estimation. Next, we evaluate the performance of our method in terms of the localization accuracy and consumption time. 

\begin{table}[tbp]\caption{ Average Time [s] of Each Thread in the Office 3 Sequence.}
\vspace{-2mm}
\centering
\setlength{\tabcolsep}{1.6mm}{
\begin{tabular}{@{}ccccc@{}}
\toprule
             & Tracking & Local Mapping & Loop Closure & Total \\ \midrule
ORB-SLAM2    & 0.019    & 0.052         & 0.421        & 0.021 \\
FastORB-SLAM & 0.009     & 0.063         & 0.532        & 0.012 \\ \bottomrule
\end{tabular}}
\label{timeOffice3}
\vspace{-5mm}
\end{table}

\begin{table*}[ht]\caption{Camera Localization RMSE Error (m) Comparison on the \textit{TUM} dataset.}
\vspace{-2mm}
\centering
\begin{threeparttable}
\setlength{\tabcolsep}{1.7mm}{
\begin{tabular}{l|cccccccccc}
\toprule
Sequence   & Ours & ORB-SLAM2  & ElasticFusion  &Kintinuous &DVO-SLAM  &RGBD-SLAM2 &BundleFusion &BAD-SLAM & Lei \textit{et al.} & Fu \textit{et al.}\\
\midrule
fr1-xyz    & \textbf{0.010}  &0.012         &  0.016       &0.018   	& 0.023  		&0.012      &0.012					&--				&-- 					&0.011\\
fr1-desk   &\textbf{0.014}  & 0.015   		& 0.020   & 0.037  			& 0.021 	& 0.026 	& 0.016 	& 0.017			& 0.021				&0.020\\
fr1-desk2  &0.025   & \textbf{0.022}   	& 0.048   & 0.071 	& 0.046  	& 0.025 &	--	& --		&--		&0.009\\
fr1-room   &0.050 	& 0.047  	& 0.068   & 0.075 	& \textbf{0.043}    	& 0.087 &	--	& 	--	&--		&--\\
fr2-desk   &0.009   & \textbf{0.008}    	& 0.071   & 0.034 	& 0.017   	& 0.057	&	--	&	--	&--		&0.009\\
fr2-xyz    &\textbf{0.006}   & \textbf{0.006}    	& 0.011   & 0.029 	& 0.018   	& 0.026 & 0.011	& 0.011	& 0.013	&0.007\\
fr2-large  & 0.181  	& 0.140   	& --   & -- 	& --   	& X & --	& --	& --	& \textbf{0.102} \\
fr3-office & 0.011   & \textbf{0.008}    	& 0.017   & 0.030 	& 0.035   	& --	& 0.022	& 0.017	& 0.027	&0.018\\
fr3-nst    & \textbf{0.018}  	& 0.019   	& \textbf{0.018}    & 0.031 	& 0.018   	& X		& X  	&--		& 0.018	&0.021\\
\bottomrule
\end{tabular}}
    \end{threeparttable} 
\begin{flushleft}
Our method achieves SOTA performance when compared with ORB-SLAM2 \cite{orbslam2}, ElasticFusion \cite{elastic}, Kintinuous \cite{kintinuous}, DVO-SLAM \cite{dvo}, RGBD-SLAM2 \cite{rgbdslam}, BundleFusion \cite{bundle}, BAD-SLAM \cite{badslam}, Lei et al. \cite{lei}, Fu et. al \cite{fu2}. ``X'' means the system failed or lost its position at some point in the sequence. ``--'' means that we cannot obtain the value from literature. 
\end{flushleft} 
\label{ttum}  
\vspace{-5mm}
\end{table*}

\begin{figure*}[ht]
\centering  
	\subfigure[Motion Trajectory Comparison]{
	\includegraphics[width=0.32\textwidth]{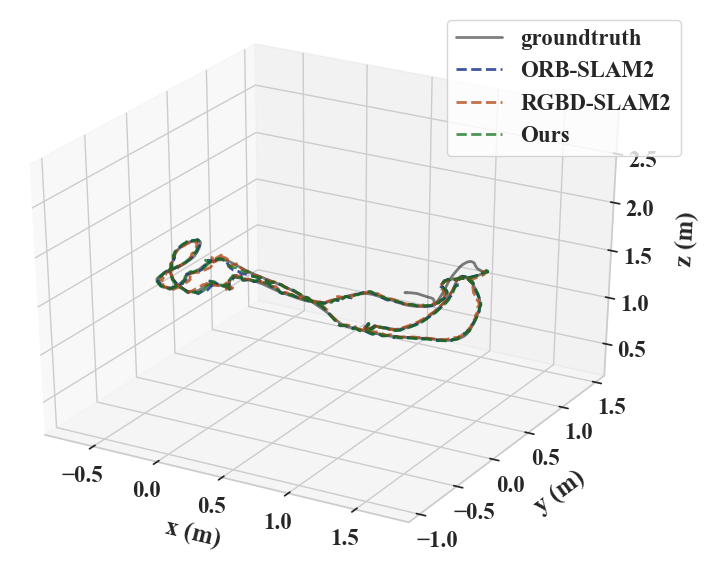}} \hspace{-1mm}
	\subfigure[Absolute Pose Error]{
	\includegraphics[width=0.32\textwidth]{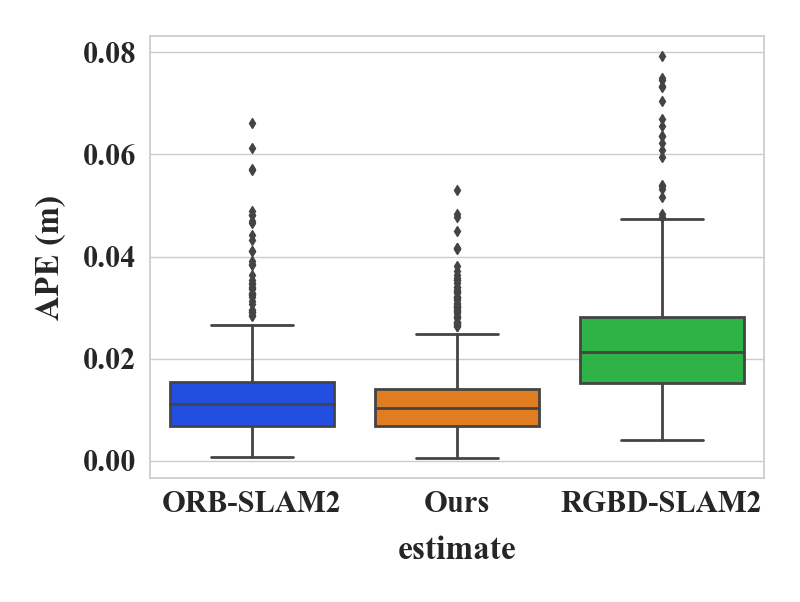}}	\hspace{-1mm}
	\subfigure[Relative Pose Error]{	
	\includegraphics[width=0.32\textwidth]{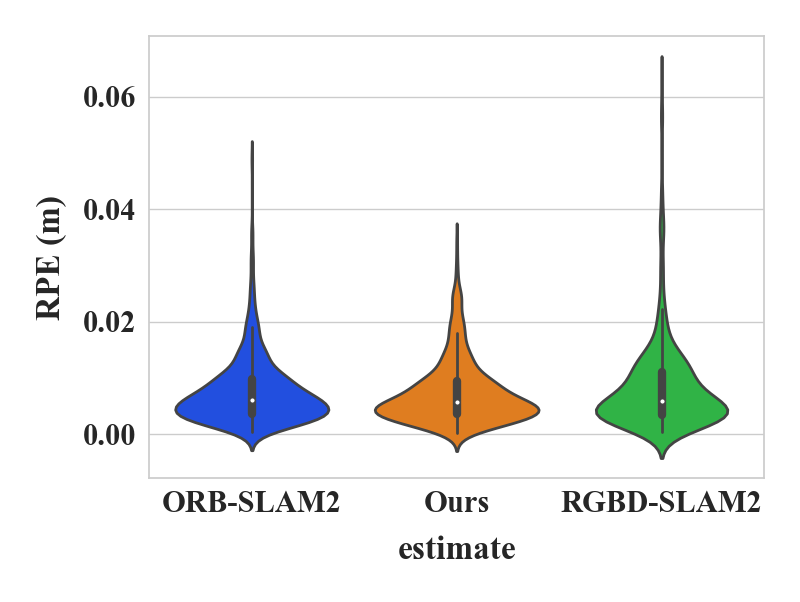}}	
	\vspace{-2mm}
\caption{Localization accuracy comparison of ORB-SLAM2, RGBD-SLAM2 and Ours in \textit{TUM-fr1-desk} sequence. 
Our method achieves the best accuracy.}\label{desk}
\vspace{-4mm}
\end{figure*}


 \vspace{-3mm}
\subsection{Localization Accuracy and Efficiency Experiments} \label{locationexperiment}


We evaluate the performance of FastORB-SLAM in terms of the localization accuracy and computation time.  We use two popular accuracy metrics, APE (absolute pose error) and RPE (relative pose error), to measure the error between the estimated trajectory and the ground-truth. \par 
APE measures the global accuracy of the estimated trajectory. For a estimated trajectory $\overline{X}=\left\{\overline{x}_{1}, \overline{x}_{2},..., \overline{x}_{n}\right\}$ and its corresponding ground-truth $X=\left\{x_{1}, x_{2},...,x_{n}\right\},$, the root mean square error (RMSE) of APE can be computed by
\begin{equation}
ATE_{RMSE}=\left(\frac{1}{n}\sum_{i=1}^{n}||trans(\overline{x}_{i})-trans(x_{i})||^{2} \right)^{\frac{1}{2}}.
\end{equation}
Similarly, RPE measures the local accuracy over a fixed time interval. More details about APE and RPE is available at\footnote{https://vision.in.tum.de/data/datasets/rgbd-dataset/tools}.
\par

\noindent \textbf{ICL-NUIM dataset}: Since our method is developed based on the benchmark ORB-SLAM2, we use it as a baseline for comparison. Experiments are conducted on all 8 sequences of the \textit{ICL-NUIM} dataset.
Quantitative and qualitative results are presented in Table \ref{ticl} and Fig. \ref{icltime}, respectively. All values are computed in a real reproduction test. RMSE represents the root mean square translation error in meters and ``Time'' represents average computation time per frame in seconds. A values are medians of 5 executions for each sequence. 
Our method and ORB-SLAM2 both show high robustness, as they successfully run on all test sequences. However, our method has an obvious advantage over ORB-SLAM2 in terms of run time (see Fig. \ref{icltime} left and Time statistics in Table \ref{ticl}). In Living 3, our method takes 0.12s compared to 0.22s by ORB-SLAM2. The frame rate (FPS) of our method is nearly 84 Hz because we do not extract keypoint descriptors during \textit{Tracking}. In the \textbf{Living 1} and \textbf{Office 3} sequences, our method obtains much higher localization accuracy than ORB-SLAM2. For visual comparison, we plot the corresponding trajectories, APE, and RPE in Fig. \ref{iclfigure}. We also provide demo video to show that ORB-SLAM2 does not track effective keypoints in the low-texture part of the two sequences leading to a big drift error. Our method can track reliable keypoints at all times because the two-stage keypoint matching establishes enough correspondences and every frame is preprocessed by the adaptive histogram equalization algorithm \cite{histogram}.

In Table \ref{timeOffice3}, we further compare the elapsed time of each thread of ORB-SLAM2 and FastORB-SLAM on the Office 3 sequence as it includes a loop. Note that \textit{Tracking} runs for every frame, \textit{Local Mapping} runs only when the current frame is selected as a keyframe, and \textit{Loop Closure} runs when the current keyframe is identified as a loop. ``Total'' means the average elapsed time per frame of the complete system. We can see that FastORB-SLAM takes more time in \textit{Local Mapping} and \textit{Loop Closure} as we extract keypoint descriptors for keyframes and much less time in \textit{Tracking} as we do not extract keypoint descriptors. FastORB-SLAM has a significant advantage in run time as shown in ``Total'' column.

\noindent \textbf{TUM dataset}: We use the \textit{TUM} dataset to compare our FastORB-SLAM to those methods which report results on this dataset. These methods include RGB-D SLAM systems such as ORB-SLAM2, DVO-SLAM \cite{dvo}, RGBD-SLAM2 \cite{rgbdslam}, BundleFusion \cite{bundle}, BAD-SLAM \cite{badslam}, Lei \textit{et al.} \cite{lei}, Fu \textit{et al.} \cite{fu2}, ElasticFusion \cite{elastic}, and Kintinuous \cite{kintinuous}. \par
Quantitative results are presented in Table \ref{ttum} where the localization accuracy is evaluated as the translation RMSE. All statistics are from \cite{orbslam2, rgbdslam, badslam, bundle, lei, fu2} and real reproduction experiments. ``X'' means that the system failed or lost its position in the sequence. ``--'' means that we cannot obtain the corresponding value in the literature. 
Table \ref{ttum} shows that ORB-SLAM2 and our method both achieve better localization performance than other methods by achieving the best accuracy on 4 sequences while others achieve only one best performance at most, such as DVO-SLAM, ElasticFusion, and Fu \textit{et al.}. For fr1-desk sequence, we plot the 3D motion trajectory of ORB-SLAM2, RGBD-SLAM2, and our method in Fig. \ref{desk}(a).
Our method achieves the best accuracy in terms of APE (Fig. \ref{desk}(b)) and RPE (Fig. \ref{desk}(c)).

\par
\noindent \textbf{Discussion}: We compared FastORB-SLAM with nine RGB-D SLAM methods on 17 sequences (8 from \textit{ICL-NUIM}, 9 from \textit{TUM}). Quantitative results (Table \ref{ttum})  show that our method and ORB-SLAM2 are both better than the remaining 8 methods due to the effective three-thread structure. ORB-SLAM2 has a relatively large drift error on the Living 1 and Office 3 sequences from \textit{ICL-NUIM}, but maintains a high-standard localization accuracy on the remaining 15 sequences. The proposed FastORB-SLAM maintains low error on all sequences. Hence, we can claim that our method is highly competitive in terms of accuracy and robustness. More importantly, our method clearly stands out in terms of computation time as shown in Fig. \ref{icltime}(a). We also demonstrated that the brightness invariance and the neighborhood motion consistency are two reasonable assumptions in time-varying sequences, and can be used to speed up the process of keypoint matching between adjacent frames without descriptor extraction.
\par

\vspace{-1mm}
\section{Conclusion}
We presented FastORB-SLAM, a fast and light-weight visual SLAM method based on ORB-SLAM2 and sparse optical flow. FastORB-SLAM exploits a new structure that matches keypoints between adjacent frames based on minimizing grayscale errors and matches keypoints between non-adjacent frames (keyframes) based  on keypoint descriptors. Such a design balances the competing needs between the  localization accuracy and computational complexity. 
To achieve this, a two-stage descriptor-independent keypoint matching method was introduced where the UAM model is used to predict the initial keypoint correspondences and an eight-level image pyramid is used in a coarse-to-fine way to solve for the movement vector in the first stage. The GMS algorithm and RANSAC-based epipolar constraints were then adopted to filter outliers. The proposed FastORB-SLAM extracts descriptors only when 
necessary saving valuable computational resources for online and embedded systems.
For future work, we plan to improve FastORB-SLAM from two aspects: (1) Further improve computational efficiency by leveraging fewer but more reliable keypoints; (2) Extend FastORB-SLAM to multiple sensors (e.g. stereo cameras, IMU) to improve robustness in challenging scenes with quick movement.\par

%

\ifCLASSOPTIONcaptionsoff
  \newpage
\fi

\bibliography{references}{}

\begin{thebibliography}{1}

\bibitem{orbslam2}R. Mur-Artal and J. D. Tardós, ``ORB-SLAM2: An open-source SLAM system for monocular, stereo, and RGB-D cameras,'' \textit{IEEE Trans. Robot.}, vol. 31, no. 5, pp. 1147–1163, Oct. 2017.

\bibitem{orbslam1}R. Mur-Artal, JMM. Montiel, JD. Tardos, ``ORB-SLAM: a versatile and accurate monocular SLAM system", \textit{IEEE Trans. Robot.}, vol. 31, no. 5, pp. 1147-1163, 2015.

\bibitem{svo1}C. Forster, M. Pizzoli, D. Scaramuzza, ``SVO: Fast semi-direct monocular visual odometry", in \textit{Proc. IEEE Int. Conf. Robot. Automat. (ICRA)}, Jun. 2014, pp. 15–22.

\bibitem{dso}J. Engel, V. Koltun, D. Cremers, ``Direct sparse odometry'', \textit{IEEE Trans. Pattern Anal. Mach. Intell.}, vol. 40, no. 3, pp. 611-625, 2017.

\bibitem{svo2}C. Forster, Z. Zhang, M. Gassner, M. Werlberger, and D. Scaramuzza, ``SVO: Semidirect visual odometry for monocular and multicamera systems,'' \textit{IEEE Trans. Robot}, vol. 33, no. 2, pp. 249-265, Apr. 2017.

\bibitem{cnn-slam}K. Tateno, F. Tombari F, L. Laina, et al., ``Cnn-slam: Real-time dense monocular slam with learned depth prediction'', in  \textit{Proc. IEEE Conf. Comput. Vis. Pattern Recognit. (CVPR)}, 2017, pp. 6243-6252.

\bibitem{cnn-svo}SY. Loo , AJ. Amiri, S. Mashohor, et al., ``CNN-SVO: Improving the mapping in semi-direct visual odometry using single-image depth prediction'', in \textit{Proc. IEEE Int. Conf. Robot. Automat. (ICRA)}, 2019, pp. 5218-5223.

\bibitem{openvslam}S. Sumikura, M.Shibuya, K. Sakurada, ``OpenVSLAM: A Versatile Visual SLAM Framework'', in \textit{Proc. ACM Int. Conf. Multimedia}, 2019, pp. 2292-2295.

\bibitem{tip1}G. Salem, J. Krynitsky, M. Hayes, T. Pohida and X. Burgos-Artizzu, ``Three-Dimensional Pose Estimation for Laboratory Mouse From Monocular Images," \textit{IEEE Trans. Image Processing}, vol. 28, no. 9, pp. 4273-4287, Sept. 2019.

\bibitem{tip2}Y. Wang, B. Zhang and C. Peng, ``SRHandNet: Real-Time 2D Hand Pose Estimation With Simultaneous Region Localization," \textit{IEEE Trans. Image Processing}, vol. 29, pp. 2977-2986, 2020.

\bibitem{posetip}H. Zhou, T. Zhang T,W. Lu, `` Vision-based pose estimation from points with unknown correspondences'', \textit{IEEE Trans. Image Processing}, vol. 23, no. 8, pp. 3468-3477, 2014. 

\bibitem{fu1}H. Yu, Q. Fu, Z. Yang, L. Tan, W. Sun, and M. Sun, ``Robust robot pose estimation for challenging scenes with an RGB-D camera,” \textit{IEEE Sens. J.}, vol. 19, no. 6, pp. 2217-2229, Mar. 2018..

\bibitem{dvo}C. Kerl , J. Sturm J, D. Cremers. ``Dense visual SLAM for RGB-D cameras'', in \textit{Proc. IEEE/RSJ Int. Conf. Intell. Robots Syst. (IROS)}, 2013, pp. 2100-2106.

\bibitem{rgbdslam}F. Endres, J. Hess J, J. Sturm, et al., ``3-D mapping with an RGB-D camera'', \textit{IEEE Trans. Robot.}, vol. 30, no. 1, pp. 177-187, 2013.

\bibitem{bundle}A. Dai , M. Nießner, M. Zollhöfer, et al., ``Bundlefusion: Real-time globally consistent 3d reconstruction using on-the-fly surface reintegration'', \textit{ACM Trans. Graphics}, vol. 36, no. 4, pp. 1-12, 2017.

\bibitem{badslam}A. Dai , M. Nießner, M. Zollhöfer, et al., ``Bad slam: Bundle adjusted direct rgb-d slam'', in \textit{Proc. IEEE Conf. Comput. Vis. Pattern Recognit. (CVPR)}, 2019, pp. 134-144.

\bibitem{lei}L. Han , L. Xu, D. Bobkov , et al., ``Real-time global registration for globally consistent rgb-d slam'', \textit{IEEE Trans. Robot.}, vol. 35, no. 2, pp. 498-508, 2019.

\bibitem{plslam}R. Gomez-Ojeda, F. Moreno, D. Zuñiga-Noël, et al., ``PL-SLAM: A stereo SLAM system through the combination of points and line segments,'' \textit{IEEE Trans. Robot.}, vol. 35, no. 3, pp. 734-746, Jun. 2019.

\bibitem{direct-imu}S. Wen, Y. Zhao , H. Zhang , et al., ``Joint optimization based on direct sparse stereo visual-inertial odometry,'' \textit{Auto. Robots}, pp. 1-19, 2020.

\bibitem{fu2}Q. Fu , H. Yu, L. Lai, et al., ``A Robust RGB-D SLAM System With Points and Lines for Low Texture Indoor Environments,'' \textit{IEEE Sens. J.}, vol. 19, no. 1, pp. 9908-9920, 2019.

\bibitem{ptam}G. Klein and D. Murray, ``Parallel tracking and mapping for small AR workspaces.'' \textit{in Proc. 6th IEEE ACM Int. Symp. Mixed Augmented Reality}, Nov. 2007, pp. 225-234.

\bibitem{bags}D. Galvez-López and J. D. Tardos, ``Bags of binary words for fast place recognition in image sequences,'' \textit{IEEE Trans. Robot.}, vol. 28, no. 5, pp. 1188-1197, Oct. 2012.

\bibitem{covision}H, Strasdat, AJ. Davison, JMM. Montiel, et al., ``Double window optimisation for constant time visual SLAM,'' in \textit{Proc. Int. Conf. Comput. Vis. (ICCV)}, 2011, pp. 2352-2359.

\bibitem{KLT}S. Birchfield , ``KLT: An implementation of the Kanade-Lucas-Tomasi feature tracker,'' http://www. ces. clemson.edu/stb/klt/, 2007.

\bibitem{ddso}N. Yang, R. Wang, J. Stuckler, et al., ``Deep virtual stereo odometry: Leveraging deep depth prediction for monocular direct sparse odometry,'' \textit{in Proc. Eur. Conf. Comput. Vis. (ECCV)}, 2018, pp: 817-833.

\bibitem{d3vo}N. Yang , L. Stumberg, R. Wang R, et al., ``D3VO: Deep Depth, Deep Pose and Deep Uncertainty for Monocular Visual Odometry,'' in \textit{Proc. IEEE Conf. Comput. Vis. Pattern Recognit. (CVPR)}, 2020, pp: 1281-1292.

\bibitem{sdso}R. Wang, M. Schworer, D. Cremers, ``Stereo DSO: Large-scale direct sparse visual odometry with stereo cameras,'' in \textit{Proc. Int. Conf. Comput. Vis. (ICCV)}, 2017, pp. 3903-3911.

\bibitem{rolldso}D. Schubert, N. Demmel, V. Usenko V, et al., ``Direct sparse odometry with rolling shutter,'' in Proc. Eur. Conf. Comput. Vis. (ECCV), 2018, pp 682-697.

\bibitem{imudso}L. Von Stumberg, V. Usenko, D. Cremers, ``Direct sparse visual-inertial odometry using dynamic marginalization,'' in \textit{Proc. IEEE Int. Conf. Robot. Autom. (ICRA)}, 2018, pp. 2510-2517.

\bibitem{vins1}T. Qin, P. Li, S. Shen, ``Vins-mono: A robust and versatile monocular visual-inertial state estimator,'' \textit{IEEE Trans. Robot.}, vol. 34, no. 4, pp. 1004-1020, 2018.

\bibitem{vins2}T. Qin, J. Pan, S. Cao, et al., ``A general optimization-based framework for local odometry estimation with multiple sensors,'' arXiv preprint arXiv:1901.03638, 2019.

\bibitem{vio}V. Usenko, N. Demmel N, D. Schubert, et al., ``Visual-inertial mapping with non-linear factor recovery,'' \textit{IEEE Robot. Autom. Lett.}, vol. 5, no. 2, pp. 422-429, 2019.

\bibitem{ldso}X. Gao, R. Wang, N. Demmel N, et al., ``LDSO: Direct sparse odometry with loop closure,'' in \textit{Proc. IEEE/RSJ Int. Conf. Intell. Robots Syst. (IROS)}, 2018: 2198-2204.

\bibitem{zhkp}Z. Dai, X. Huang, W. Chen W, et al., ``A Comparison of CNN-Based and Hand-Crafted Keypoint Descriptors,'' in \textit{Proc. IEEE Int. Conf. Robot. Autom. (ICRA)}, 2019, pp: 2399-2404.

\bibitem{zhBoCNF}Y. Hou, H. Zhang, S. Zhou, ``BoCNF: efficient image matching with Bag of ConvNet features for scalable and robust visual place recognition,'' \textit{Autonomous Robots}, vol. 42, no. 6, pp. 1169-1185.2018, 

\bibitem{xl}X. Wang, G. Peng, H. Zhang, `` Combining multiple image descriptions for loop closure detection,'' \textit{J. of Intell. Robot. Syst.}, vol. 92, no. 3, pp. 565-585, 2018.

\bibitem{ORB}E. Rublee, V. Rabaud, K. Konolige, et al., ``ORB: An efficient alternative to SIFT or SURF, '' in \textit{Proc. IEEE Int. Conf. Comput.
Vis. (ICCV)}, 2011, pp. 2564-2571.

\bibitem{dynaslam}B. Bescos, JM. Fácil, J. Civera, et al., ``DynaSLAM: Tracking, mapping, and inpainting in dynamic scenes,'' \textit{IEEE Robot. Autom. Lett.}, vol. 3., no. 4, pp. 4076-4083, 2018.

\bibitem{locationorb}A. Sujiwo, T. Ando, E. Takeuchi, et al., ``Monocular vision-based localization using ORB-SLAM with LiDAR-aided mapping in real-world robot challenge,'' \textit{J. robot. mechatronics}, vol. 28, no. 4, pp. 479-490, 2016.

\bibitem{Buyval}A. Buyval, I. Afanasyev, E. Magid, ``Comparative analysis of ROS-based monocular SLAM methods for indoor navigation,'' in \textit{Prof. Ninth Int. Conf. Machine Vision (ICMV)}, pp. 10341- 10349, 2016.

\bibitem{Zhao}Z. Zhao, Y. Mao, Y. Ding, et al., ``Visual Semantic SLAM with Landmarks for Large-Scale Outdoor Environment,'' arXiv preprint arXiv:2001.01028, 2020.

\bibitem{Webb}AM. Webb, G. Brown, M. Luján, ``ORB-SLAM-CNN: Lessons in Adding Semantic Map Construction to Feature-Based SLAM, '' in \textit{Prof. Annual Conf. Towards Auto. Robot. Syst.}, 2019, pp. 221-235.

\bibitem{orbslam3}C. Campos, R. Elvira, JJ. Rodríguez JJ, et al., ``ORB-SLAM3: An Accurate Open-Source Library for Visual, Visual-Inertial and Multi-Map SLAM," arXiv preprint arXiv:2007.11898, 2020.

\bibitem{histogram}JA. Stark, ``Adaptive image contrast enhancement using generalizations of histogram equalization,'' \textit{IEEE Trans. image processing}, vol. 9, no. 5, pp. 889-896, 2000.

\bibitem{gms1}JW. Bian, WY. Lin, Y. Matsushita, et al., ``Gms: Grid-based motion statistics for fast, ultra-robust feature correspondence,'' in \textit{Proc. IEEE Conf. Comput. Vis. Pattern Recognit. (CVPR)}, 2017, pp. 4181-4190.

\bibitem{gms2}JW. Bian, WY. Lin, Y, et al., ``GMS: Grid-Based Motion Statistics for Fast, Ultra-robust Feature Correspondence,'' {Int. J. Comput. Vision}, vol. 128, pp. 1580–1593, 2020. 

\bibitem{epipolar} G. Xu, Z. Zhang, ``Epipolar geometry in stereo, motion and object recognition: a unified approach,'' Springer Science Business Media, 2013.

\bibitem{elastic}T. Whelan, RF. Salas-Moreno, B. Glocker, et al., `` ElasticFusion: Real-time dense SLAM and light source estimation,'' \textit{Int. J. Robot. Research}, vol. 35, no. 4, pp. 1697-1716, 2016.

\bibitem{kintinuous}T. Whelan, M. Kaess, M. Fallon, et al., ``Kintinuous: Spatially extended kinectfusion,'' 2012.

\bibitem{TUM} J. Sturm, N. Engelhard, F. Endres, et al., ``A benchmark for the evaluation of RGB-D SLAM systems,'' in \textit{Proc. IEEE/RSJ Int. Conf. Intell. Robots Syst. (IROS)}, 2012, pp. 573-580.

\bibitem{ICL}A. Handa, T. Whelan, J. McDonald, et al., ``A benchmark for RGB-D visual odometry, 3D reconstruction and SLAM,'' in \textit{Proc. IEEE Int. Conf. Robot. Automat. (ICRA)}, 2014, pp. 1524-1531.

\bibitem{g2o}R. Kümmerle, G. Grisetti, H. Strasdat, K. Konolige, and W. Burgard, ``G2o: A general framework for graph optimization,'' in \textit{Proc. IEEE Int. Conf. Robot. Autom. (ICRA)}, May 2011, pp. 3607–3613.

\bibitem{dxslam}D. Li, X. Shi, Q. Long, S. Liu S, et al., ``DXSLAM: A robust and efficient visual SLAM system with deep features, arXiv preprint arXiv:2008.05416, 2020.

\bibitem{zhao}Z. Yipu, W. Ye, PA. Vela, ``Low-latency visual SLAM with appearance-enhanced local map building," in \textit{Proc. IEEE Int. Conf. Robot. Autom. (ICRA)}, 2019, pp. 8213-8219.

\bibitem{mask}K. He, G. Gkioxari, P. Dollár, ``Mask r-cnn,'' in \textit{Proc. Int. Conf. Comput. Vis. (ICCV)}, 2017, pp. 2961-2969.

\bibitem{nid} G. Pascoe, W. Maddern W, M. Tanner M, et al., ``NID-SLAM: Robust monocular SLAM using normalised information distance'', in \textit{Proc. IEEE Conf. Comput. Vis. Pattern Recognit. (CVPR)}, 2017 pp. 1435-1444.

\bibitem{fusion} J. McCormac, R. Clark R, M. Bloesch, et al., ``Fusion++: Volumetric object-level slam'', in \textit{Proc. IEEE Int. Conf. 3D vision (3DV)}, 2018, pp. 32-41.

\bibitem{camnet} M. Ding , Z. Wang, J. Sun, et al., `` Camnet: Coarse-to-fine retrieval for camera re-localization'', in \textit{Proc. IEEE Conf. Comput. Vis. Pattern Recognit. (CVPR)}, 2019, pp. 2871-2880.

\bibitem{coarse} Y. Mori, T. Hirakawa, T. Yamashita, et al., ``Coarse-to-Fine Deep Orientation Estimator for Local Image Matching'', in \textit{Proc. Asian Conf. Pattern Recognit}, 2019, pp. 378-390.

\bibitem{feng} D. Feng, C. Wang, C. He, et al., ``Kalman-filter-based integration of IMU and UWB for high-accuracy indoor positioning and navigation'', \textit{IEEE Internet of Things Journal}, vol. 7, no. 4, pp. 3133-2146, 2019.

\bibitem{combining} J. Martínez-Carranza, LO. Rojas-Perez, AA. Cabrera-Ponce, et al., ``Combining deep learning and RGBD SLAM for monocular indoor autonomous flight'', in \textit{Mexican Int. Conf. Artificial Intelligence}, 2018, pp. 356-367.

\bibitem{fan3}B. Fan, Q. Kong, X. Wang, et al.. ``A performance evaluation of local features for image-based 3D reconstruction,'' \textit{IEEE Trans. Image Processing}, vol. 28, no. 10, pp. 4774-4789, 2019.

\bibitem{fan1}B. Fan, H. Liu, H. Zeng, et al. ``Deep unsupervised binary descriptor learning through locality consistency and self distinctiveness,'' \textit{IEEE Trans. Multimedia}, 2020.

\bibitem{duan}Y. Duan, J. Lu, Z. Wang, et al., ``Learning deep binary descriptor with multi-quantization,'' in \textit{Proc. IEEE Conf. Comput. Vis. Pattern Recognit. (CVPR)}, 2017, pp. 1183-1192.

\bibitem{lin}K. Lin, J Lu, CS. Chen, et al., ``Unsupervised deep learning of compact binary descriptors,'' \textit{IEEE Trans. Pattern Anal. Mach. Intell.}, vol. 41, no. 6, pp. 1501-1514, 2018.

\bibitem{fan2}H. Liu, Q. Zhang, B. Fan, et al. ``Features combined binary descriptor based on voted ring-sampling pattern,'' \textit{IEEE Trans. Circuits and Syst. Video Technology}, vol. 35, no. 10, pp. 3675-3687, 2019.

\bibitem{fan4}B. Fan, Q. Kong, B. Zhang, et al., ``Efficient nearest neighbor search in high dimensional hamming space,'' \textit{Pattern Recognition}, 2020.

\bibitem{selflow}P. Liu, M. Lyu, I. King, et al., ``Selflow: Self-supervised learning of optical flow'' in \textit{Proc. IEEE Conf. Comput. Vis. Pattern Recognit. (CVPR)}, 2019, pp. 4571-4580.

\bibitem{flownet}E. Ilg, N. Mayer, T. Saikia, et al., ``Flownet 2.0: Evolution of optical flow estimation with deep networks,'' in \textit{Proc. IEEE Conf. Comput. Vis. Pattern Recognit. (CVPR)}, 2017, pp. 2462-2470.

\bibitem{liteflownet}TW. Hui, X. Tang X, CC. Loy, ``Liteflownet: A lightweight convolutional neural network for optical flow estimation,'' in \textit{Proc. IEEE Conf. Comput. Vis. Pattern Recognit. (CVPR)}, 2018, pp. 8981-8989.

\bibitem{zhang20} T. Zhang, H. Zhang, Y. Li, et al., ``Flowfusion: Dynamic dense rgb-d slam based on optical flow,'' in \textit{Proc. IEEE Int. Conf. Robot. Autom. (ICRA)}, 2020, pp. 7322-7328.

\bibitem{cheng19}J. Cheng, Y. Sun Y, MQ. Meng, ``Improving monocular visual SLAM in dynamic environments: an optical-flow-based approach,'' \textit{Advanced Robotics}, vol. 33, no.12, pp. 576-589, 2019.

\bibitem{yu18}C. Yu, Z. Liu, XJ. Liu, et al., ``DS-SLAM: A semantic visual SLAM towards dynamic environments,'' in \textit{Proc. IEEE Int. Conf. Robot. Autom. (ICRA)}, 2018, pp. 1168-1174.

\bibitem{lie}J. Sola, J. Deray, D. Atchuthan, ``A micro Lie theory for state estimation in robotics,'' arXiv preprint arXiv:1812.01537, 2018.

\bibitem{orb}E. Rublee, V. Rabaud, K. Konolige, et al., ORB: An efficient alternative to SIFT or SURF. in \textit{Proc. Int. Conf. Comput. Vis. (ICCV)}, 2011, pp. 2564-2571.

\end{thebibliography}
\bibliographystyle{unsrt}


\begin{IEEEbiography}[{\includegraphics[width=1in,height=1.25in,clip,keepaspectratio]{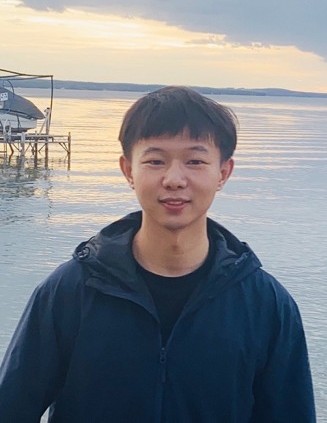}}]{Qiang Fu}
was born in China. He is currently working toward a Ph.D. degree with the National Engineering Laboratory for Robot Visual Perception and Control, College of Electrical And Information Engineering, Hunan University, China, under the supervision of Prof. Hongshan Yu. \par 
He is also currently a Visiting Scholar with the University of Alberta, Robotic and Vision Group, Edmonton, AB, Canada, under the supervision of Prof. Hong Zhang (IEEE Fellow). His research interests include mobile robot, visual SLAM, computer vision (email: cn.fq@qq.com).
\end{IEEEbiography}

\begin{IEEEbiography}[{\includegraphics[width=1in,height=1.25in,clip,keepaspectratio]{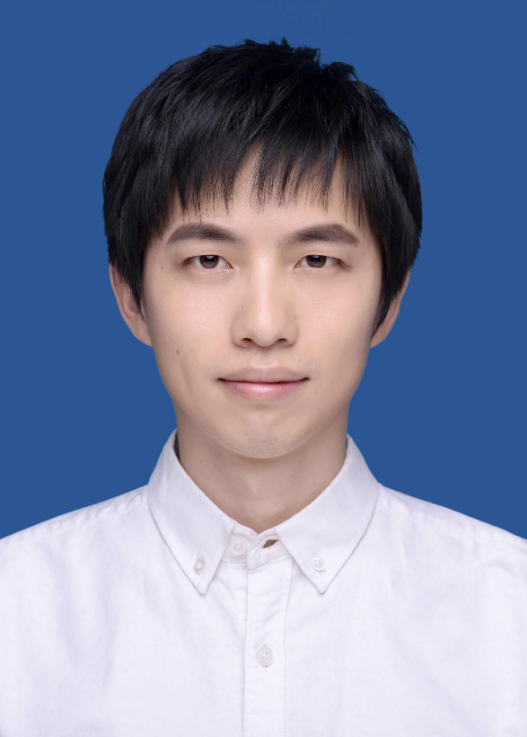}}]{Xiaolong Wang}
received his B.Sc., M.Sc. and Ph.D. degree in Applied Mathematics from Northwestern Polytechnical University (NWPU), China. He was a visiting doctoral student in the department of computing science at University of Alberta, Canada. From 2019 to 2020, he was a Post-Doctoral Researcher at the University of Alberta, Robotic and Vision Group, under the supervision of Prof. Hong Zhang (IEEE Fellow). He is currently a lecturer in School of Mathematics and Information Science, Shaanxi Normal University, China. His research is in the area of visual SLAM system and collaborative perception.
\end{IEEEbiography}


\begin{IEEEbiography}[{\includegraphics[width=1in,height=1.25in,clip,keepaspectratio]{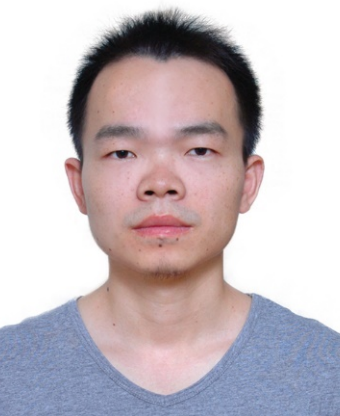}}]{Zhengeng Yang}
received the B.S. and M.S. degrees from Central South University, Changsha, China, in 2009 and 2012, respectively. He is currently working toward his Ph.D. degree with the National Engineering Laboratory for Robot Visual Perception and Control, Hunan University, China, under the supervision of Prof. Hongshan Yu. He is also currently a Visiting Scholar with the University of Pittsburgh, Pittsburgh, PA, USA. 
His research interests include computer vision, image analysis, and machine learning.
\end{IEEEbiography}


\begin{IEEEbiography}[{\includegraphics[width=1in,height=1.25in,clip,keepaspectratio]{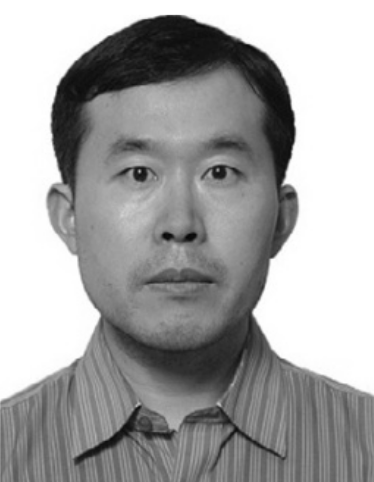}}]{Hongshan Yu}
received the B.S., M.S., and Ph.D. degrees in control science and technology in electrical and information engineering from Hunan University, Changsha, China, in 2001, 2004, and 2007, respectively. 
From 2011 to 2012, he was a Post-Doctoral Researcher with the Laboratory for Computational Neuroscience, University of Pittsburgh, USA. He is currently a Professor with Hunan University and Associate Dean of the National Engineering Laboratory for Robot Visual Perception and Control. His research interests include autonomous mobile robot and machine vision, with over 30 publications in these areas.
\end{IEEEbiography}

\begin{IEEEbiography}[{\includegraphics[width=1in,height=1.25in,clip,keepaspectratio]{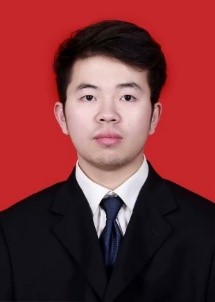}}]{Yong He}
received the B.S. degree from Anhui University of Science \& Technology, Huainan, Anhui, China, in 2015. He received the M.S. degree from China University of Mining and Technology, Xuzhou, Jiangsu, China, in 2018. He is currently pursing the Ph.D. degree with Hunan University, Changsha, China, and he is also currently a Visiting Scholar with University of Western Australia, Perth, Australia. His research interests include computer vision, point clouds analysis, and deep learning..
\end{IEEEbiography}

\begin{IEEEbiography}[{\includegraphics[width=1in,height=1.25in,clip,keepaspectratio]{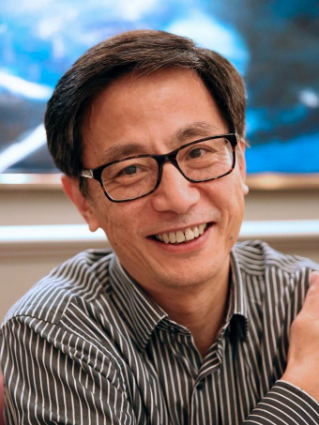}}]{Hong Zhang}
(Fellow, IEEE) received the B.S. degree from Northeastern University, Boston, MA, USA, in 1982, and the Ph.D. degticlree from Purdue University, West Lafayette, IN, USA, in 1986, both in electrical engineering. 
He conducted research at the University of Pennsylvania, Philadelphia, PA, USA, as a PostDoctoral Fellow. He is a currently chair professor at  SUSTech, China and also a professor emeritus at University of Alberta, Canada. His research interests include robotics, computer vision, and image processing, with over 200 publications in these areas. For the past 15 years, he has expended considerable effort in the study of mobile robot navigation with visual sensing. 
Dr. Zhang is a fellow of the Canadian Academy of Engineering in recognition of his accomplishments. He held a prestigious NSERC Industrial Research Chair from 2003 to 2017. He was the General Chair of the 2017 IEEE/RSJ International Conference on Intelligent Robots and Systems (IROS), in Vancouver, Canada, and is on the Senior Program Committee (SPC) of IROS 2018, IROS 2019, and ICRA 2019. He is serving as the Secretary of IEEE Robotics and Automation Society (2018–2019), and has been appointed as the Editor-in-Chief of the IROS Conference Editorial Board for a term of three years from 2020 to 2022. He has served on the editorial boards of several international journals including the IEEE TRANSACTIONS ON CYBERNETICS and the MDPI journal of Robotics. He is a Principal Investigator in the NSERC Canadian Robotics Networks, the NCFRN from 2012 to 2017, and the NCRN from 2018 to 2023, whose mandate is to develop the science and technologies to allow mobile robots to work in challenging environments and to generate and communicate critical information to humans.
\end{IEEEbiography}

\begin{IEEEbiography}[{\includegraphics[width=1in,height=1.25in,clip,keepaspectratio]{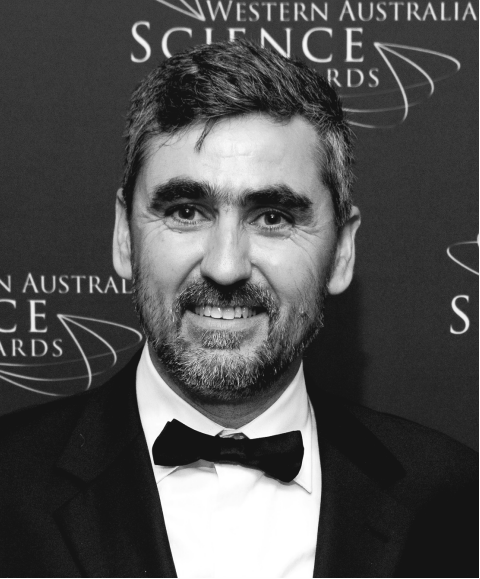}}]{Ajmal Mian}
is a Professor of Computer Science at The University of Western Australia. He is the recipient of three prestigious fellowships from the Australian Research Council (ARC). He has also received several research grants from the ARC, US Department of Defense, and the National Health and Medical Research Council of Australia with a combined funding of over $\$40$ million. He was the West Australian Early Career Scientist of the Year 2012 and has received several awards including the Excellence in Research Supervision Award, EH Thompson Award, ASPIRE Professional 
Development Award, Vice-chancellors Mid-career Award, Outstanding Young Investigator Award, the Australasian Distinguished Dissertation Award and various best paper awards. He is an ACM Distinguished Speaker and an Associate Editor for IEEE Transactions in Neural Networks \& Learning Systems, IEEE Transactions on Image Processing and the Pattern Recognition journal. He was the General Co-Chair of DICTA 2019 and ACCV 2018. His research interests are in computer vision, 3D deep learning, shape analysis, face recognition, and video analysis.
\end{IEEEbiography}

\end{document}